\documentclass[twoside, 11pt]{article}
\usepackage{jair, rawfonts}
\usepackage[round]{natbib}

\usepackage{amsthm}

\usepackage{blindtext}

\usepackage{microtype}
\usepackage{booktabs} 
\usepackage[shortlabels]{enumitem}
\usepackage[most]{tcolorbox}

\newtcolorbox{HighlightBox}[1][]{%
   enhanced,
   before    = \begin{center},
   after     = \end{center},
   width     = 14cm,
   colback   = blue!5,
   colframe  = blue!40!white, 
   arc       = 4mm, 
   outer arc = 3.5mm, 
   #1}

\newtcolorbox[auto counter]{OpenQuestionBox}[2][]{%
   enhanced,
   detach title,
   sharp corners=northwest,
   title={#2 \thetcbcounter},
   width     = 14cm,
   fonttitle=\bfseries,
   before    = \begin{center},
   after     = \end{center},
   colback   = green!5,
   colframe  = green!50!black, 
   coltitle  = green!5!white, 
   colbacktitle = green!50!black,
   attach boxed title to top left={yshift=-4mm,yshifttext=-2mm},
   boxed title style={rounded corners, arc = 2.5mm, outer arc = 2.5mm},
   arc       = 4mm, 
   outer arc = 3.5mm, 
   #1}
   
\newtcolorbox{HighlightBoxFit}[1][]{%
   enhanced,
   hbox,
   before    = \begin{center},
   after     = \end{center},
   width     = 14cm,
   colback   = blue!5,
   colframe  = blue!40!white, 
   arc       = 4mm, 
   outer arc = 3.5mm, 
   #1}

\newtcolorbox{OpenQuestionBoxFit}[2][]{%
   enhanced,
   hbox,
   detach title,
   sharp corners=northwest,
   title={#2},
   fonttitle=\bfseries,
   before    = \begin{center},
   after     = \end{center},
   colback   = green!5,
   colframe  = green!50!black, 
   coltitle  = green!5!white, 
   colbacktitle = green!50!black,
   attach boxed title to top left={yshift=-4mm,yshifttext=-2mm},
   boxed title style={rounded corners, arc = 2.5mm, outer arc = 2.5mm},
   arc       = 4mm, 
   outer arc = 3.5mm, 
   #1}

\definecolor{royalblue}{RGB}{65, 105, 225} 

\usepackage{hyperref}
\hypersetup{
    colorlinks,
    linkcolor={royalblue},
    citecolor={royalblue},
    urlcolor={royalblue}
}

\newif\ifjmlr
\jmlrtrue 

\usepackage{braket}
\usepackage{pgf,tikz}
\usetikzlibrary{arrows.meta}
\usetikzlibrary{calc}
\usetikzlibrary{decorations.pathreplacing}

\usepackage{float}
\usepackage{physics}
\usepackage{xifthen}
\usepackage{xparse}
\usepackage{float}
\usepackage{dsfont}
\usepackage{xspace}
\usepackage{wrapfig}
\usepackage{subcaption}
\usepackage{cancel}
\usepackage{thm-restate}
\usepackage{tabularx} 
\usepackage{chngcntr}
\usepackage{apptools}

\newtheorem{theorem}{Theorem}

\newtheorem{definition}[theorem]{Definition}

\newtheorem{assumption}{Assumption}
\newtheorem*{theorem*}{Theorem}
\newtheorem*{definition*}{Definition}
\newtheorem*{proposition*}{Proposition}

\newtheorem{thm}{Theorem}[section]

\newtheorem{prop}[thm]{Proposition}

\DeclareMathOperator\supp{supp}
\DeclareMathOperator*{\essinf}{ess\,inf}
\newcommand{\lr}[1]{\left (#1\right)}

\let\L\undefined
\newcommand{\L}{L(\bmtheta)}
\newcommand{\Lprime}{L(\bmtheta')}
\newcommand{\Lhat}{\hat{L}(D,\bmtheta)}
\newcommand{\Lhatprime}{\hat{L}(D,\bmtheta')}

\newcommand{\J}{J_{\bmtheta}(\lambda)}
\newcommand{\Jprime}{J_{\bmtheta'}(\lambda)}

\newcommand{\pn}{\tfrac{1}{n}\ln\tfrac{k^p}{\delta}}

\newcommand{\I}[1]{{\cal I}_\bmtheta(#1)}
\newcommand{\Iprime}[1]{{\cal I}_{\bmtheta'}(#1)}
\newcommand{\Iinv}[1]{{\cal I}^{-1}_{\bmtheta}\left(#1\right)}
\newcommand{\Iinvprime}[1]{{\cal I}^{-1}_{\bmtheta'}\left(#1\right)}

\newcommand{\V}{\mathbb{V}}

\NewDocumentCommand{\1}{o}{\mathds 1{\IfValueT{#1}{\lr{#1}}}}
\let\P\undefined
\NewDocumentCommand{\P}{o}{\mathbb P{\IfValueT{#1}{\lr{#1}}}}

\DeclareMathOperator*{\argmin}{arg\,min}
\DeclareMathOperator*{\esssup}{ess\,sup}

\newcommand{\bmtheta}{{\ensuremath{\boldsymbol{\theta}}}}
\newcommand{\bmTheta}{{\ensuremath{\boldsymbol{\Theta}}}}

 \newcommand{\bmx}{{\ensuremath{\boldsymbol{x}}}}
\newcommand{\bmy}{{\ensuremath{\boldsymbol{y}}}} \newcommand{\E}{{\ensuremath{\mathbb{E}}}}
\newcommand{\Lim}[1]{\raisebox{0.5ex}{\scalebox{0.8}{$\displaystyle \lim_{#1}\;$}}}

\usepackage[framemethod=TikZ]{mdframed}

\usepackage{amsmath}
\usepackage{xstring}
\usepackage{regexpatch}

\makeatletter
\xpatchcmd\thmt@restatable{%
\csname #2\@xa\endcsname\ifx\@nx#1\@nx\else[{#1}]\fi
}{%
\ifthmt@thisistheone
\csname #2\@xa\endcsname[{\hyperref[proof:#3]{Proof}}]\ifx\@nx#1\@nx\else{\textup{\textbf{(#1).}}}\fi
\else
\csname #2\@xa\endcsname[{\hyperref[#2:#3]{Return}}]
\fi}{}{}
\makeatother

\usepackage[utf8]{inputenc} 
\usepackage[T1]{fontenc}    
\usepackage{hyperref}       
\usepackage{url}            
\usepackage{booktabs}       
\usepackage{amsfonts}       
\usepackage{nicefrac}       
\usepackage{microtype}      
\usepackage{xcolor}         

\usepackage{lastpage}

\jairheading{82}{2025}{503-562}{08/2024}{02/2025}

\ShortHeadings{PAC-Chernoff Bounds}{Masegosa \& Ortega}
\firstpageno{503}

\begin{document}

    \defcitealias{bubeck2023universal}{Bubeck and Sellke \color{black}{(}\color{royalblue}{2023}\color{black}{)}}
    \defcitealias{nagarajan2019uniform}{Nagarajan and Kolter \color{black}{(}\color{royalblue}{2019b}\color{black}{)}}

\title{PAC-Chernoff Bounds: Understanding Generalization \\ in the Interpolation Regime}

\author{\name Andr{\'e}s R. Masegosa \email                 arma@cs.aau.dk \\
       \addr Department of Computer Science\\
       University of Aalborg\\
       \AND
       \name Luis A. Ortega \email luis.ortega@uam.es \\
       \addr Machine Learning Group \\
       Department of Computer Science \\
       Escuela Politécnica Superior\\
       Universidad Autónoma de Madrid}

\maketitle

\begin{abstract}%
This paper introduces a distribution-dependent PAC-Chernoff bound that exhibits perfect tightness for interpolators, even within over-parameterized model classes. This bound, which relies on basic principles of Large Deviation Theory, defines a natural measure of the smoothness of a model, characterized by simple real-valued functions. Building upon this bound and the new concept of smoothness, we present an unified theoretical framework revealing why certain interpolators show an exceptional generalization, while others falter. We theoretically show how a wide spectrum of modern learning methodologies, encompassing techniques such as $\ell_2$-norm, distance-from-initialization and input-gradient regularization, in combination with data augmentation, invariant architectures, and over-parameterization, collectively guide the optimizer toward smoother interpolators, which, according to our theoretical framework, are the ones exhibiting superior generalization performance. This study shows that distribution-dependent bounds serve as a powerful tool to understand the complex dynamics behind the generalization capabilities of over-parameterized interpolators.
\end{abstract}

\section{Introduction}\label{sec:intro}
    
    In modern machine learning, model classes have such large capacity that optimizers virtually always retrieve a model \textit{interpolating} the training data; that is, with null training error \citep{zhang2017understanding}. These models are called \textit{interpolators} and it is well known that there are many within a large model class \citep{livni2014computational} and that some of them have a remarkably small generalization error (difference between ``training error'' and ``expected or test error'') while others do not \citep{feldman2020neural}. 
    
    The machine learning community has made a great effort during the last years to understand why modern learning algorithms retrieve, most of the time, interpolators with a small generalization error \citep{nagarajan2019uniform,bartlett2020benign}. Most of these attempts often focus on providing generalization bounds. These bounds provide an upper limit on the expected error, tying it to variables related to the training dataset and the model produced by a learning algorithm. Such bounds typically resemble the following form, 
    \begin{equation}\label{eq:generalbounds}
        L({\cal A}(D)) \leq \hat{L}({\cal A}(D),D) + {\cal C}({\cal A}(D),D,\delta)\,,
    \end{equation}
    \noindent 
    where \(D\) represents the training dataset, independently and identically distributed (i.i.d.) from a data-generating distribution \(\nu\); \(\mathcal{A}\) denotes the learning algorithm, which generates a hypothesis (a model) from a given data set (for example, via the gradient descent method); \(L\) and \(\hat{L}\) indicate the expected and empirical errors of a hypothesis, respectively; \(\mathcal{C}\) denotes a complexity measure; and the inequality holds with high probability, at least, $1-\delta$ over draws of datasets from the distribution \(\nu\). 
    
    Examples of bounds falling inside this scheme include Vapnik–Chervonenkis (VC) bounds \citep{vapnik2015uniform,bartlett2019nearly}, Rademacher bounds \citep{mohri2018foundations}, (Norm and margin)-based bounds \citep{nagarajan2019generalization, neyshabur2015path} and Sharpness-based measures \citep{nagarajan2019deterministic, keskar2016large}. Even PAC-Bayes with a fixed prior \citep{mcallester1999pac} can be included here if we consider algorithms retrieving a distribution over the models.

    However, many recent works are increasingly suggesting that these bounds are probably vacuous in the over-parameterized regime and, in consequence, unable to explain generalization in this setting. \cite{zhang2017understanding} was the first one to provide quite convincing empirical evidence of this in the context of deep neural networks. \citetalias{nagarajan2019uniform} later provided a series of experiments and theoretical results in the same direction. These works also show that even in the case where the complexity measure \(\mathcal{C}\) depends directly on the algorithm \(\mathcal{A}\) rather than simply in its output \(\mathcal{A}(D)\) (for example, assuming the algorithm always retrieve models whose parameters' norm is lower than a given constant) are also vacuous. Recently, \cite{wang2024near} showed how any near-interpolator exhibits rapid parameter norm growth, which implies that existing data-dependent parameter-norm-based bounds are necessarily loose, concluding that \textit{``explaining the generalization capability of near-interpolators will require new tools''}. More generally, \cite{gastpar2023fantastic} managed to show that, under some conditions resembling over-parameterization, there is no generalization bound like Equation~\eqref{eq:generalbounds} tight for all data-generating distributions, and that algorithm-dependent bounds are also provably vacuous. For all these reasons, the emerging conclusion is that:
    
    \begin{HighlightBox}
    \begin{center}
        Bounds that solely depend on the training data are provably vacuous for over-parameterized model classes and are unable to explain generalization. 
    \end{center}
    \end{HighlightBox}
    
    The aforementioned works \citep{nagarajan2019uniform,nagarajan2021explaining,gastpar2023fantastic} directly or indirectly advocate for the need of exploring alternative bounds that not solely depend on the training data but also use information of the data-generating distribution. These bounds are usually referred to as \emph{distribution-dependent bounds}; where the complexity term ${\cal C}$ explicitly depends on the data-generating distribution $\nu$. Distribution-dependent bounds have been used before in different contexts \citep{zhang2006information, catoni2007pac}, however, to the best of our knowledge, none of these bounds is known to be tight or, at least, non-vacuous for over-parameterized interpolators. Even more importantly, these kind of bounds have not been used before to analyze the generalization of over-parametrized interpolators. 
    
    \subsection{Our Contribution}
    
    In this study, we investigate the use of distribution-dependent bounds within a finite hypothesis space. Our aim is to understand how modern learning techniques manage to produce models that both interpolate the data and have a small generalization error. More precisely, Theorem~\ref{thm:LDTinv} introduces a distribution-dependent PAC-Chernoff bound that is \textit{perfectly tight} for any algorithm retrieving models interpolating the training data, even when the model class is over-parameterized. This bound has the following standard form
    \begin{equation*}
        L({\cal A}(D)) \leq \hat{L}({\cal A}(D),D) + {\cal C}({\cal A}(D),n,\nu,\delta)\,,
    \end{equation*}
    where, now, the complexity term ${\cal C}$ also depends on the data-generating distribution $\nu$. Proposition~\ref{prop:boundtightness} shows that with high-probability \(1 - \delta\), that
    \begin{equation*}
        {\cal C}({\cal A}(D),n,\nu,\delta) \leq L({\cal A}(D)) \leq \hat{L}({\cal A}(D),D) + {\cal C}({\cal A}(D),n,\nu,\delta)\,,
    \end{equation*}
    which ensures that the proposed bound is \emph{perfectly tight for interpolators}, that is to say, to models whose empirical loss is null or small enough to be considered negligible or within an acceptable error margin, denoted as \(\hat{L}({\cal A}(D),D)\leq \epsilon\). 
    
    As we show in this work, these bounds are directly connected to Large Deviation Theory (LDT) \citep{ellis2006entropy} because their complexity measure  \({\cal C}({\cal A}(D),n,\nu,\delta)\) directly depends on the so-called \textit{rate function} (also known as the Cramér-Chernoff function), which is the central element of LDT. We employ the \textit{rate function} to present a new characterization of the \textit{smoothness} of a model using distribution-dependent measures. According to Theorem~\ref{thm:smallerL}, this approach enables a precise characterization of \textit{which interpolators better generalize}, addressing an outstanding open question in machine learning.
    
    Despite being based on oracle quantities, we show that the theoretical framework built around this complexity measure based on the rate function allows to create an unified explanation of a wide range of learning techniques used in modern machine learning. Namely, \(\ell_2\)-norm, distance from initialization and input-gradient regularization (Section~\ref{sec:explicit}), invariant architectures, data augmentation (Section~\ref{sec:invariances}) and over-parameterization (Section~\ref{sec:Over-parameterization}). Under this framework, we show why each of these learning techniques produce interpolators that generalize well and why and when they complement each other (for example, why \(\ell_2\)-norm regularization is also effective in combination with the use of invariant architectures). 
    
    We \textit{do not claim} that the proposed theoretical analyses of each of these learning techniques are better than existing ones at individual level. For example, we do not claim that our explanation about why \(\ell_2\)-norm promotes generalization is better than existing ones, for example, in the context of linear regression \citep{bartlett2020benign}. However, we \textit{do claim} that using the rate function and PAC-Chernoff bounds, which are distribution-dependent quantities, it is possible to jointly explain, up to some degree, the generalization of interpolators, the role of over-parameterization and why many widely used modern learning techniques find interpolators that generalize. To the best of our knowledge, this is the first theoretical framework capable of achieving insights that covers such a wide range of learning techniques. At the same time, this work shows that distribution-dependent bounds are a promising research direction for understanding the generalization of modern machine learning methods.
    

\section{Preliminaries}\label{sec:Preliminaries}
    
    First of all, let us introduce the notation and the assumptions of this work. Let $D$ denote a given \emph{training data set} with size $n>0$, which we assume has been i.i.d. generated from an unknown distribution, denoted by \(\nu(\bmy, \bmx)\). We assume we have a  model class parameterized by a parameter vector $\bmtheta \in \bmTheta$. 
    We denote for any model
     $\bmtheta\in\bmTheta$, its loss as $\ell(\bmy,\bmx,\bmtheta)$, its expected loss as $L(\bmtheta)=\E_\nu[\ell(\bmy,\bmx,\bmtheta)]$, and its empirical loss as $\hat{L}(\bmtheta,D) = \frac{1}{n}\sum_i \ell(\bmy_i,\bmx_i,\bmtheta)$. Finally, the subset of models having a null variance (constant predictions) is defined as $\bmTheta_0 := \{\bmtheta\in\bmTheta: \V_\nu(\ell(\bmy,\bmx,\bmtheta))=0\}$. 
    
    We also consider that our model class \textit{lives} inside a finite-precision computing machine. In that sense, we assume our model class is represented by $p$ parameters. Then, we will have $k^{p}$ different models, under a numeric precision of \(\log_2(k)\) bits (for example, the value of \(k\) would be \(k = 2^{32}\) for single floating point precision and \(k = 2^{64}\) for double floating point precision). This model class defines a finite grid  (using a discretization approach) contained in $\mathbb{R}^{p}$, but it does not imposed any restriction in the definition of the loss, which can be defined in a continuous space. In Section \ref{sec:conclusions}, we discuss how the results presented here could be potentially extended to infinite model classes by using recently introduced PAC-Bayes-Chernoff bounds \citep{casado2024pac}. Let us now introduce the specific assumptions of this work:

    \begin{assumption}\label{assump:lowerbound}
        The loss is lower-bounded; more precisely, the essential infimum of the loss is finite and positive. That is,
        \(
        m_\bmtheta :=\essinf_{(\bmx, \bmy)\, \in\, \supp(\nu)} \ell(\bmy,\bmx,\bmtheta) \geq 0 \quad \forall \bmtheta\in\bmTheta
        \). And the expected loss is always finite, $\forall \bmtheta\in\bmTheta\;\; \L< \infty$.
    \end{assumption}
    
    This assumption assures that the loss function is lower-bounded and the generalization error is finite (not necessarily upper-bounded). The first part of the assumption is naturally satisfied in many standard problems; for example, in multiclass classification problems with softmax activation and the cross-entropy loss and regression problems with mean squared error, as it is clear that \(m_\bmtheta = 0\). The second part of the assumption ($\L< \infty$) is mainly introduced for the sake of simplicity in the mathematical exposition. In fact, this could be relaxed in many of the theoretical results of this work, under some considerations. However, under the scope of this work, that is studying the generalization error of interpolators, assuming a finite expected error, and thus, a finite generalization error, is a reasonable simplification for the exposition of the theoretical results.

    Some extra considerations regarding this assumption raise in the case of a Gaussian likelihood and log-loss, which can be considered another \emph{usual setting} in machine learning. In this setup, the density has to be lower than one for any data sample and the variance of the Gaussian distribution can not be null in order to satisfy Assumption~\ref{assump:lowerbound}. However, this is not restrictive; lower than one Gaussian densities is usual in high dimensional Gaussians and can be ultimately imposed with a restriction on the variances. On the other hand, non-zero variances are usually \emph{desirable} to ensure stability of machine learning models. In fact, variances are typically restricted to be positive or computed as the exponential of a logarithm-scaled variable, ensuring their positiveness.

    As an example of a case not considered under Assumption~\ref{assump:lowerbound}, the loss function cannot be Gaussian-distributed, as \(m_\bmtheta = - \infty\) in this case. In this regard an exponentially-distributed loss could be used verifying Assumption~\ref{assump:lowerbound}.

    
    
\section{The Rate Function}\label{sec:rate}

    \begin{figure}
        \centering
      \begin{subfigure}[t]{.45\linewidth}
        \includegraphics[width=\linewidth]{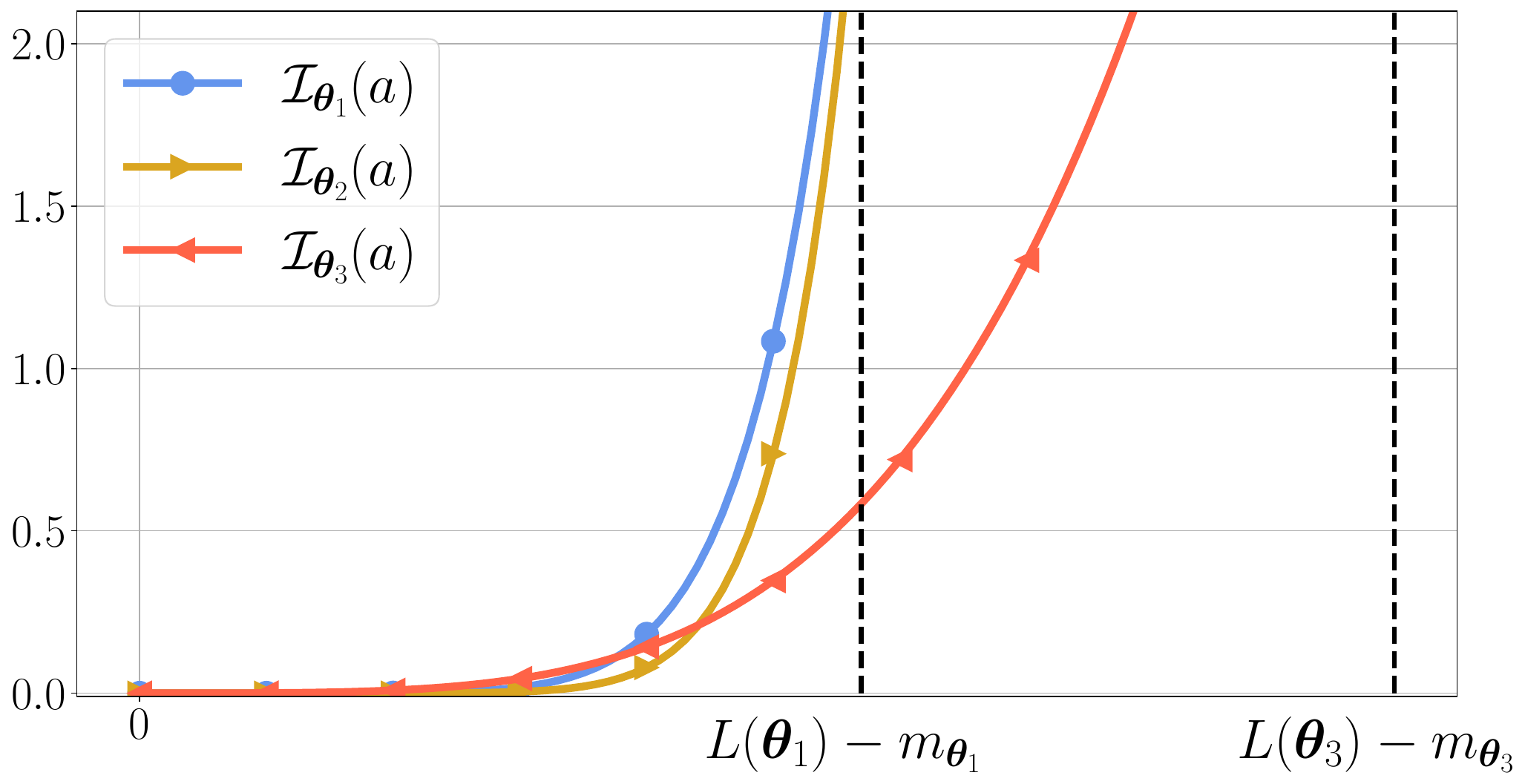}
      \end{subfigure}
      \begin{subfigure}[t]{.44\linewidth}
        \includegraphics[width=\linewidth]{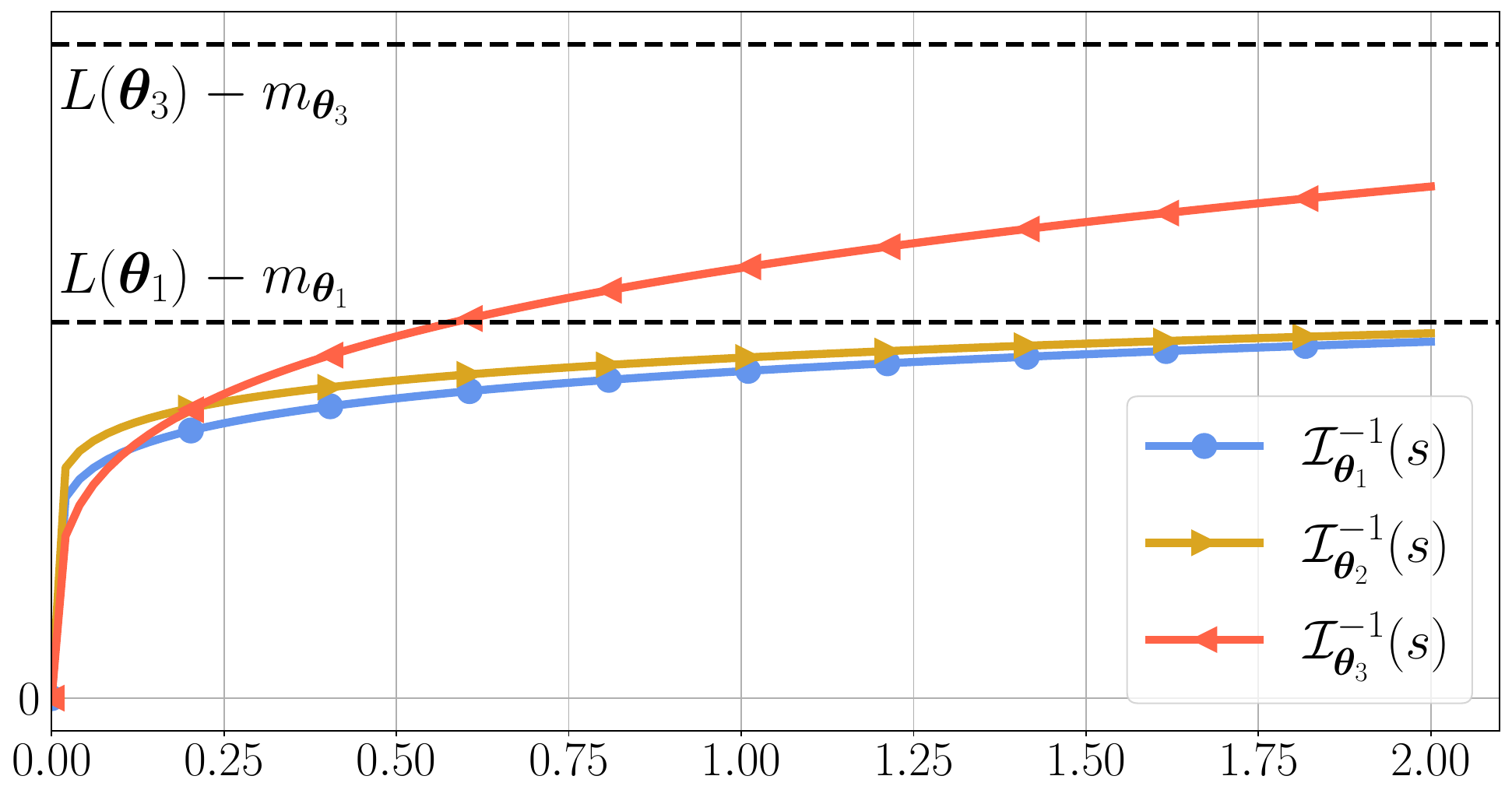}
      \end{subfigure}
    \caption{Illustration on different \emph{rate function} (left) and \emph{inverse rate function} (right) with the same or different domains of definition: three different models \(\bmtheta_1, \bmtheta_2, \bmtheta_3 \in \bmTheta\) are shown where \(\bmtheta_1\) and \(\bmtheta_2\) share the same definition interval for their rate functions. } 
    \label{fig:RateFunction}
    \end{figure}

    We start introducing the key function of this work, the so-called \textit{rate function}, denoted by $\I{a}$. The rate function places a central role in Large Deviation Theory (LDT) \citep{ellis2006entropy}, a branch of probability theory tightly connected to statistical mechanics, that deals with understanding the behavior of rare events or large fluctuations in random systems. The rate function is normally used to understand the underlying structure of these events and how their probabilities change as they move away from the typical or average behavior.
    
    Mathematically, the rate function is defined as the \textit{Legendre transform} of the \emph{cummulant-generating function}, denoted by $\J$, which is the natural logarithm of the \emph{moment-generating function} of a random variable, in this case \(\L - \ell(\bmy, \bmx, \bmtheta)\) with $\bmy,\bmx\sim\nu(\bmy,\bmx)$.
     
    \begin{definition}[Rate Function]\label{def:ratefunction}
        For any model \(\bmtheta \in \bmTheta\), its rate function is a real-valued function \(\mathcal{I}_{\bmtheta}: [0,\L-m_\bmtheta) \to \mathbb{R}^{+}_{0}\), defined as
        \begin{equation*}\label{eq:ratefunction}
             \I{a}=\sup_{\lambda>0}\ \lambda a - \J \quad \forall a \in [0,\L-m_\bmtheta)\,,
        \end{equation*}
        where the cummulant-generating function is a real-valued function \(J_{\bmtheta}: \mathbb{R}^{+}_{0} \to \mathbb{R}^{+}_{0}\), defined as
        \begin{equation*}
            \J = \ln \E_{\nu}\Big[e^{\lambda (L(\bmtheta)-\ell(\bmy,\bmx,\bmtheta))}\Big] \quad \forall \lambda \geq 0\,.
        \end{equation*}    
    \end{definition}
    
    \noindent The inverse of the rate function, denoted $\Iinv{s}$, will also place a relevant role in this work. 
    \begin{definition}[Inverse Rate Function]\label{def:inverseratefunction}
        For any model \(\bmtheta \in \bmTheta\), the inverse rate function is a real-valued function \(\mathcal{I}_{\bmtheta}^{-1}: \mathbb{R}^{+}_{0}\to [0,\L-m_\bmtheta]\), defined as
        \begin{equation*}\label{eq:inverseratefunction}
            \Iinv{s}=\inf_{\lambda>0} \frac{\J + s}{\lambda}\quad \forall s \geq 0\,.
        \end{equation*}
    \end{definition}
    Note that, when $\mathbb{P}_\nu(\ell(\bmy,\bmx,\bmtheta)=m_\bmtheta)\neq 0$, for \(\forall s \geq -\ln \mathbb{P}_\nu(\ell(\bmy,\bmx,\bmtheta)=m_\bmtheta)\), \(\Iinv{s}\) is constantly equal to \( L(\bmtheta) - m_\bmtheta\) and, in consequence, \(\Iinv{s}\) is a \emph{generalized inverse} of \(\I{a}\) \citep{Rockafellar+1970}. For the sake of simplicity, we note that we abuse notation and assume throughout the rest of work that for $a> \L-m_\bmtheta$, $\I{a}=\infty$. We also assume that binary operators apply to this value following common sense. According to the following proposition, both the rate function and its inverse are well defined for models satisfying Assumption \ref{assump:lowerbound}.

    \begin{restatable}{prop}{Iwelldefined} \label{prop:Iwelldefined}
         Under Assumption \ref{assump:lowerbound}, $\forall\bmtheta\in\bmTheta$, $\I{\cdot}$ and $\Iinv{\cdot}$, are well defined. That is, $\forall a\in[0,\L-m_\bmtheta)$, $\I{a}<\infty$ and $\forall s\in\mathbb{R}^{+}_{0}$, $\Iinv{s}<\infty$. 
    \end{restatable}
    
    Finally, the following result states some properties of the rate and inverse rate function, shedding some light on their monotony, curvature and shape. In combination with Figure~\ref{fig:RateFunction}, this result should provide some intuition on how this two functions behave in general.

\ifjmlr
\begin{prop}[\cite{Rockafellar+1970}]\label{prop:Rateproperties}
For any $\bmtheta\in\bmTheta$, the rate function $\I{\cdot}$ and the inverse rate function $\Iinv{\cdot}$ satisfy the following properties,
        \begin{enumerate}[label=(\roman*)]
            \item \label{i} $\I{\cdot}$ is convex and $\Iinv{\cdot}$ is concave; both monotonically increasing.
            \item \label{ii} \(\Lim{a\to 0} \frac{\partial}{\partial a} \I{a} = 0\) and \( \Lim{s \to 0}\frac{\partial}{\partial s} \Iinv{s} = +\infty \).
            \item \label{iii} $\Lim{a\rightarrow (\L - m_\bmtheta)^-} \I{a}= -\ln \P(\ell(\bmy,\bmx,\bmtheta)=m_\bmtheta) $ and $\Lim{s\to -\ln \P(\ell(\bmy,\bmx,\bmtheta)=m_\bmtheta)} \Iinv{s}=\L - m_\bmtheta$.
            \item \label{iv} $\I{\cdot}$ and $\Iinv{\cdot}$ are invariant to reparameterizations. 
        \end{enumerate}
\end{prop}
\else
    \begin{restatable}{prop}{Rateproperties}\label{prop:Rateproperties}
        For any $\bmtheta\in\bmTheta$, the rate function $\I{\cdot}$ and the inverse rate function $\Iinv{\cdot}$ satisfy the following properties,
        \begin{enumerate}[label=(\roman*)]
            \item $\I{\cdot}$ is convex and $\Iinv{\cdot}$ is concave.
            \item $\I{\cdot}$ and \(\Iinv{\cdot}\) are monotonically increasing. 
            \item \(\lim_{a\to 0} \frac{\partial}{\partial a} \I{a} = 0\) and \( \lim_{s \to 0}\frac{\partial}{\partial s} \Iinv{s} = +\infty \).
            \item $\lim_{a\rightarrow \L - m_\bmtheta^-} \I{a}= -\ln \P(\ell(\bmy,\bmx,\bmtheta)=m_\bmtheta) $ and $\lim_{s\to \infty} \Iinv{s}=\L - m_\bmtheta$ 
            \item $\I{\cdot}$ and $\Iinv{\cdot}$ are invariant to reparameterizations. 
        \end{enumerate}
    \end{restatable}
\fi    
    Note that the rate function of a Gaussian random variable, like many other standard random variables, does not have a vertical asymptote. However, we should then consider that if the loss follows a Gaussian distribution, it will not satisfy the first condition in Assumption \ref{assump:lowerbound}, as the essential infimum is not finite.
    
    The relevance of the rate function is consequence of the following results; firstly,  the classic Chernoff bound, defines how likely is to observe, over different i.i.d. data sets, an empirical loss $\Lhat$ that deviates from the expected loss $L(\bmtheta)$ by a positive quantity $a > 0$.

\ifjmlr
\begin{thm}[\cite{chernoff1952measure}] \label{thm:LDT}
For any fixed $\bmtheta\in\bmTheta$ and $a>0$, it satisfies
        \begin{equation*}\label{eq:LDT}
            \P_{D\sim \nu^n}\Big(L(\bmtheta) - \hat{L}(D,\bmtheta) \geq a\Big)\leq e^{-n \I{a}}\,.
        \end{equation*}
\end{thm}
\else
    \begin{restatable}[\cite{chernoff1952measure}]{thm}{LDT} \label{thm:LDT}
        For any fixed $\bmtheta\in\bmTheta$ and $a>0$, it satisfies
        \begin{equation*}\label{eq:LDT}
            \P_{D\sim \nu^n}\Big(L(\bmtheta) - \hat{L}(D,\bmtheta) \geq a\Big)\leq e^{-n \I{a}}\,.
        \end{equation*}
    \end{restatable}
\fi    
    
    The above bound is relevant in this work because of two main factors that characterize its behavior w.r.t. the data set size \(n\) and the generalization error gap \(a\): on one hand, Chernoff's bound is known to be quite loose in the mean of the variable \((\text{when }a \approx 0)\) but tight on the tail \((\text{when }a \approx \L - m_{\bmtheta})\). As a result, the bound is specially useful when talking about interpolators, where the value of \(\L - \Lhat\) is close to its maximum possible value \((a = \L - m_{\bmtheta})\), as stated in the following result. 
    
    \begin{restatable}{prop}{tightchernoff}\label{prop:tightchernoff}
        For any fixed $\bmtheta\in\bmTheta$ and $n>0$, it satisfies
        \begin{equation*}\label{eq:TightLDT}
            \lim_{a\rightarrow \L - m_{\bmtheta}}\P_{D\sim \nu^n}\Big(L(\bmtheta) - \hat{L}(D,\bmtheta) \geq a\Big) =  \lim_{a\rightarrow \L - m_{\bmtheta}} e^{-n \I{a}}\,.
        \end{equation*}
    \end{restatable}
    
    On the other hand, Cramér's Theorem \citep{cramer1938nouveau} states that Chernoff's bound is exponentially tight for \textit{large} $n$. Formally, this statement is written as follows, 
\ifjmlr
\begin{thm}[\cite{cramer1938nouveau,ellis2006entropy}]\label{thm:crammer}
    For any fixed $\bmtheta\in \bmTheta$ and any $a>0$, it satisfies
    \begin{equation*}
        \lim_{n\rightarrow \infty} -\frac{1}{n}\ln \P_{D \sim \nu^n}\Big(\L - \Lhat \geq a\Big) = \I{a}\,.
    \end{equation*}
\end{thm}
\else
    \begin{restatable}[\cite{cramer1938nouveau,ellis2006entropy}]{thm}{crammer}\label{thm:crammer}
    For any fixed $\bmtheta\in \bmTheta$ and any $a>0$, it satisfies
    \begin{equation*}
        \lim_{n\rightarrow \infty} -\frac{1}{n}\ln \P_{D \sim \nu^n}\Big(\L - \Lhat \geq a\Big) = \I{a}\,.
    \end{equation*}
    \end{restatable}\fi    
    \noindent In LDT, the above asymptotic result is intuitively interpreted using the following equality, 
    \begin{equation}\label{eq:asympoticEquality}
    \P_{D \sim \nu^n}\Big(L(\bmtheta) - \Lhat \geq a\Big)= e^{-n\I{a} + o(n, a)}\,,
    \end{equation}
    \noindent which shows that the exact expression of $\P_{D\sim \nu^n}(L(\bmtheta) - \Lhat \geq a)$ is defined by the rate function, up to a sub-exponential term, that is negligible when $n$ is \textit{large}, because $\lim_{n\rightarrow\infty} \tfrac{o(n,a)}{n}=0$ and, in consequence, it does not have any meaningful effect. Then, according to LDT, when $n$ is large, the rate function would be the key quantity describing the generalization error of a model, that is, statistical behavior of the difference between the expected $L(\bmtheta)$ and the empirical loss $\Lhat$.

    \begin{HighlightBox}
        The Chernoff bound, governed by the rate function ${\cal I}_\bmtheta(\cdot)$, is tighter for larger datasets (when $n\rightarrow\infty$) and for models interpolating the data  (\(a \approx \L - m_{\bmtheta}\)), which are the settings of modern machine learning.  
    \end{HighlightBox}

\begin{figure}[t]
      \begin{minipage}[c]{0.45\textwidth}
        \centering
        \scalebox{0.65}{
        \begin{tabular}{lcccccc}\toprule
          Inception & Crop & L2 & Train Acc. & Test Acc. & Test NLL & \(\ell_2\)-norm\\ \midrule
          Standard  & no & no & \(99.99\%\) & \(84.36 \%\) & \(0.65\) & \(304\) \\
          Crop & yes & no & \(99.94\%\) & \(86.89\%\)  & \(0.58\) & \(309\) \\
          L2 &  no & yes &\(100.0\%\) & \(86.60 \%\)  & \(0.49\) & \(200\) \\
          L2-Crop & yes & yes & \(99.98\%\) & \(88.45 \%\) & \(0.42\)  & \(130\) \\
          \midrule
          Random &  no & no & \(100.0\%\) & \(10.13 \%\) & \(5.52\)  & \(311\) \\
          \midrule
          Initial &  - & - & \(10.00\%\) & \(10.00 \%\)  & \(2.30\) & \(593\) \\
          \bottomrule
          \end{tabular}
          }
        \end{minipage}
      \hfill
      \begin{minipage}[c]{0.44\textwidth}
        \centering
            \includegraphics[width = 0.99\textwidth]{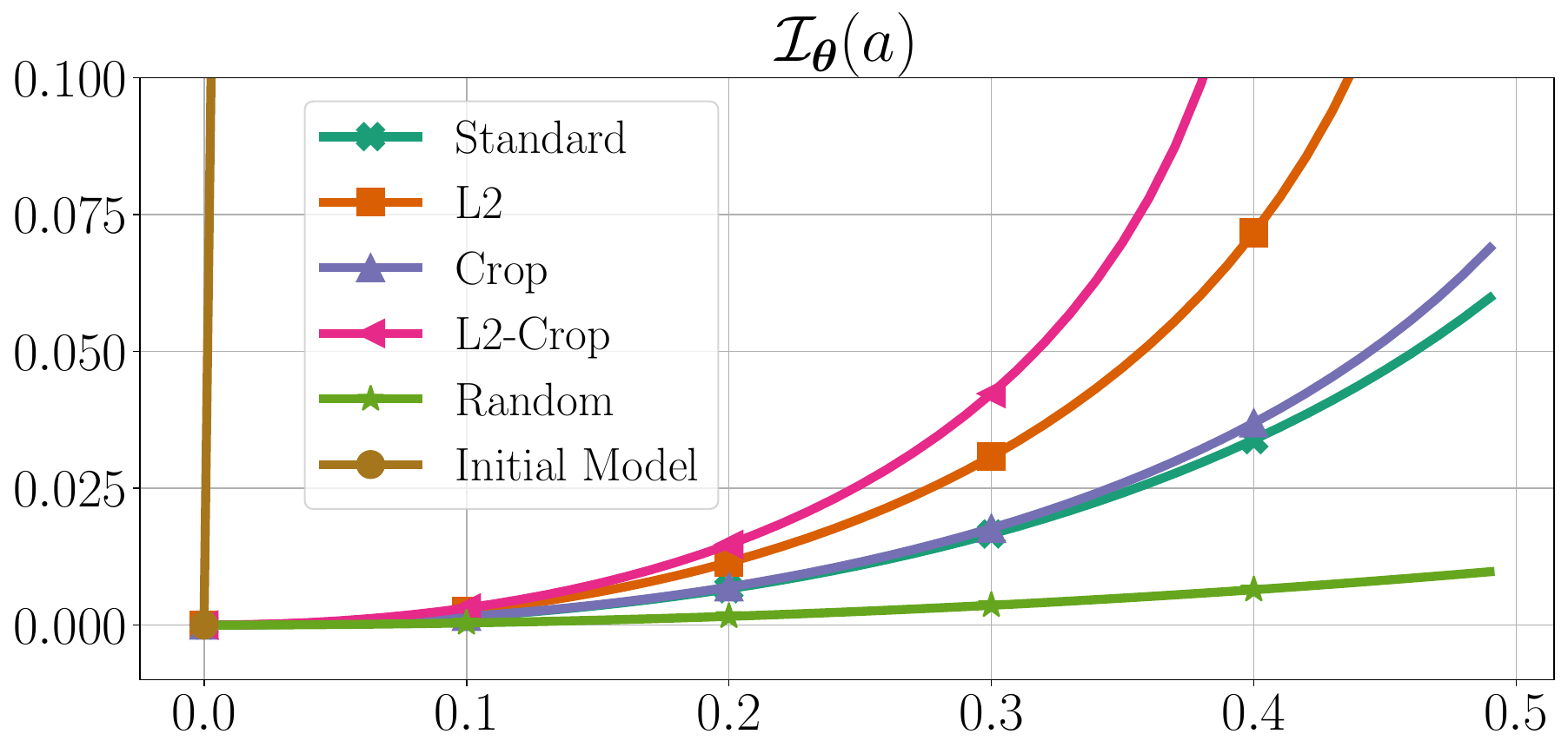}
        \end{minipage}
        \caption{Metrics of Inception models on Cifar10 using $\ell_2$ regularization and/or random cropping (Crop), and randomly sampled class labels (Random). The corresponding \emph{rate functions} are shown on the right.}
        \label{fig:1}
    \end{figure}

    The Chernoff bound states that the generalization error a model is governed by its rate function. The rate function is an \emph{oracle} and \emph{distribution-dependent} quantity because it depends on the data-generating distribution $\nu$. In machine learning, this distribution $\nu$ is always unknown,  which may be considered as a problem. However, as discussed in the introduction, we will show how the use of a distribution-dependent quantity allows to explain, for example, why data-augmentation works, why we need over-parameterized model families, and why we should consider invariant architectures, like convolutional neural networks. Furthermore, in this work, we also show how the rate function can be easily estimated from an independent data set, if it is used as a proxy for the data-generating distribution. This is a common practice in machine learning by making use of a separated \emph{validation data set} to get estimates of $\L$ without incurring in \emph{data snooping}. In Appendix \ref{app:discussion:assumption}, we show that the rate function $\I{\cdot}$ can be easily estimated by \textit{running} the model once on this validation data set and using log-sum-exp operations to estimate \(\J\); and using a simple grid search to optimize the value of \(\lambda\) in Definition~\ref{def:ratefunction}. \textit{The fact that the cumulant and rate function can be easily estimated and plotted (as in Figure~\ref{fig:1}) is, in our opinion, a relevant finding of this work.} This will allow us to empirically illustrate and validate the predictions made by our theoretical framework.

    \begin{figure}[t]
        \centering
      \begin{subfigure}[b]{.33\linewidth}\centering Standard\\[1.5mm]
        \includegraphics[width=\linewidth]{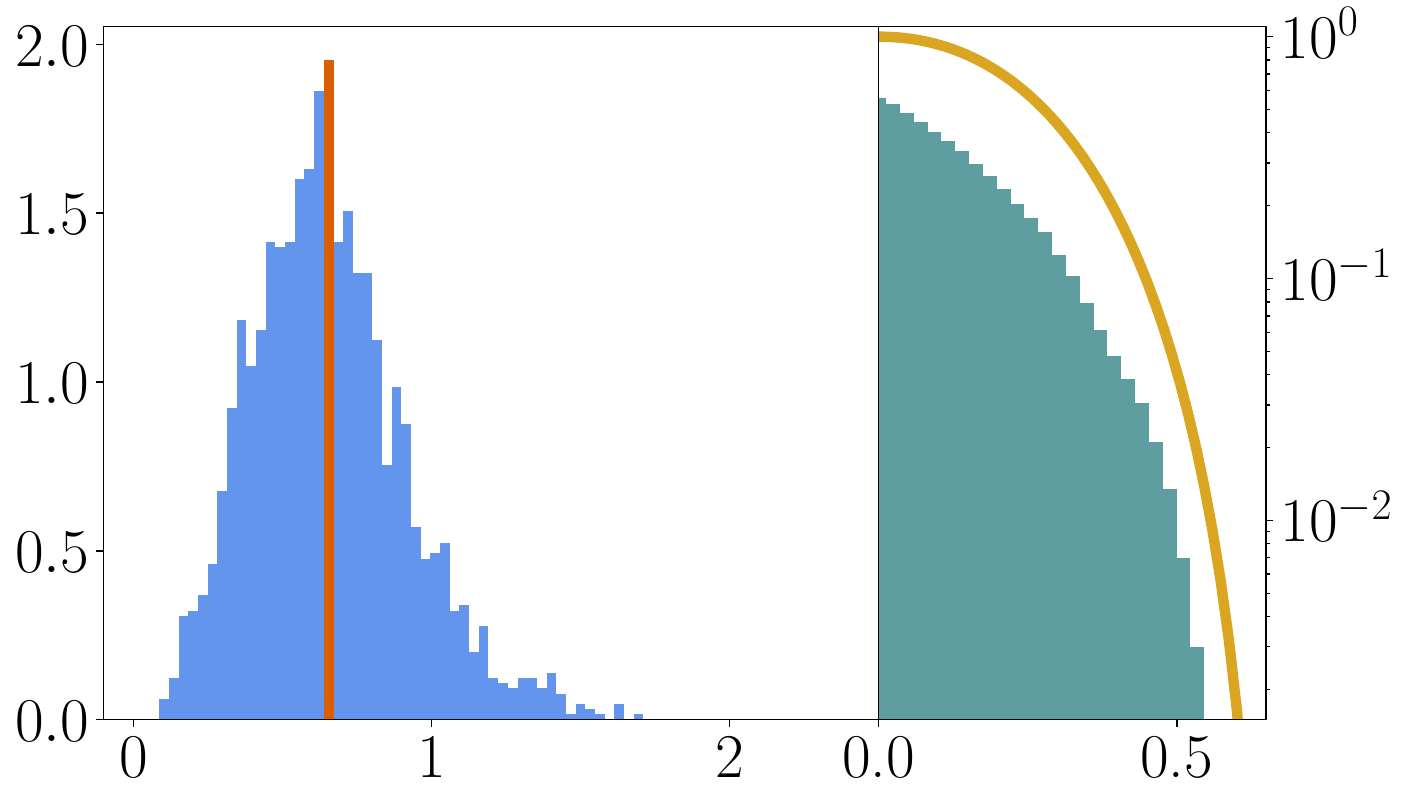}
      \end{subfigure}
      \begin{subfigure}[b]{.315\linewidth}\centering L2-Crop\\[1.5mm]
        \includegraphics[width=\linewidth]{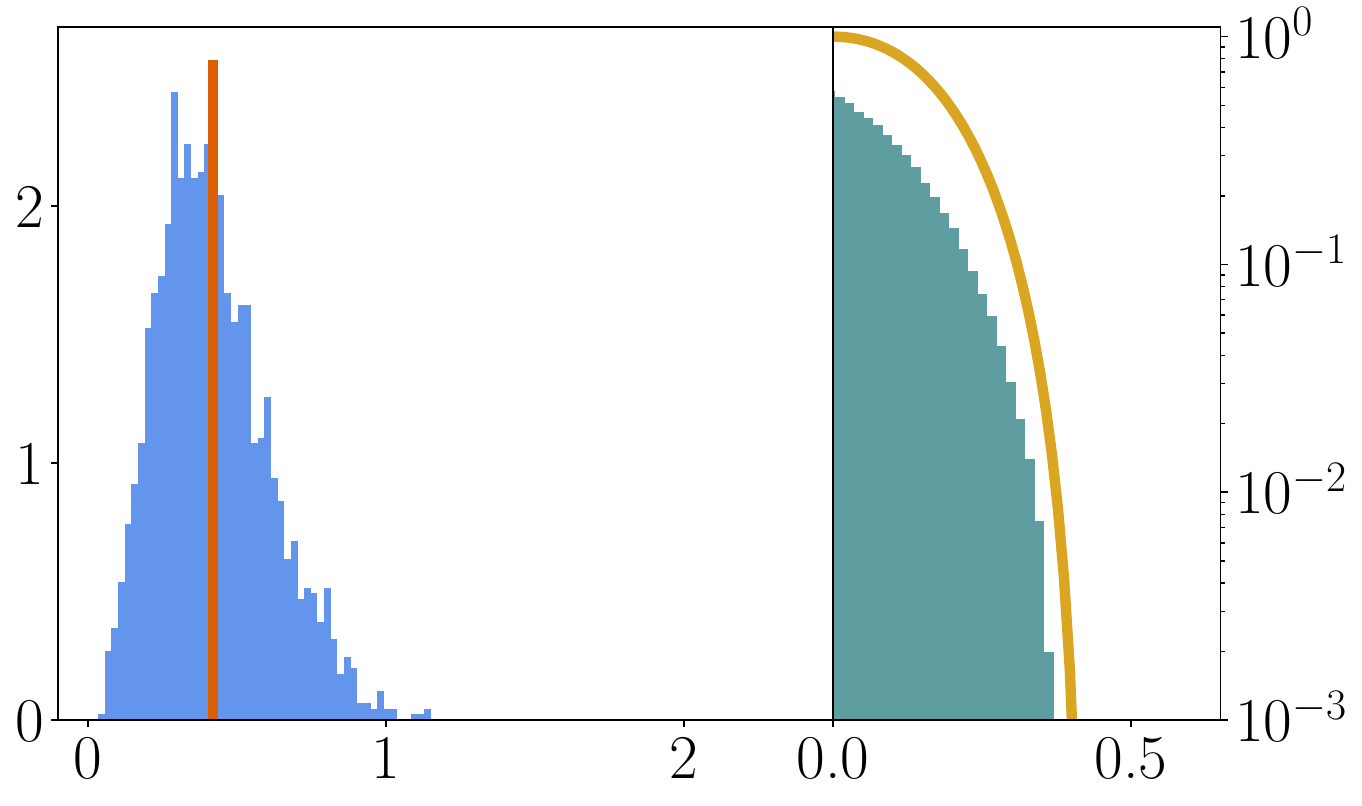}
      \end{subfigure}
      \begin{subfigure}[b]{.33\linewidth}\centering Initial Model\\[1.5mm]
        \includegraphics[width=\linewidth]{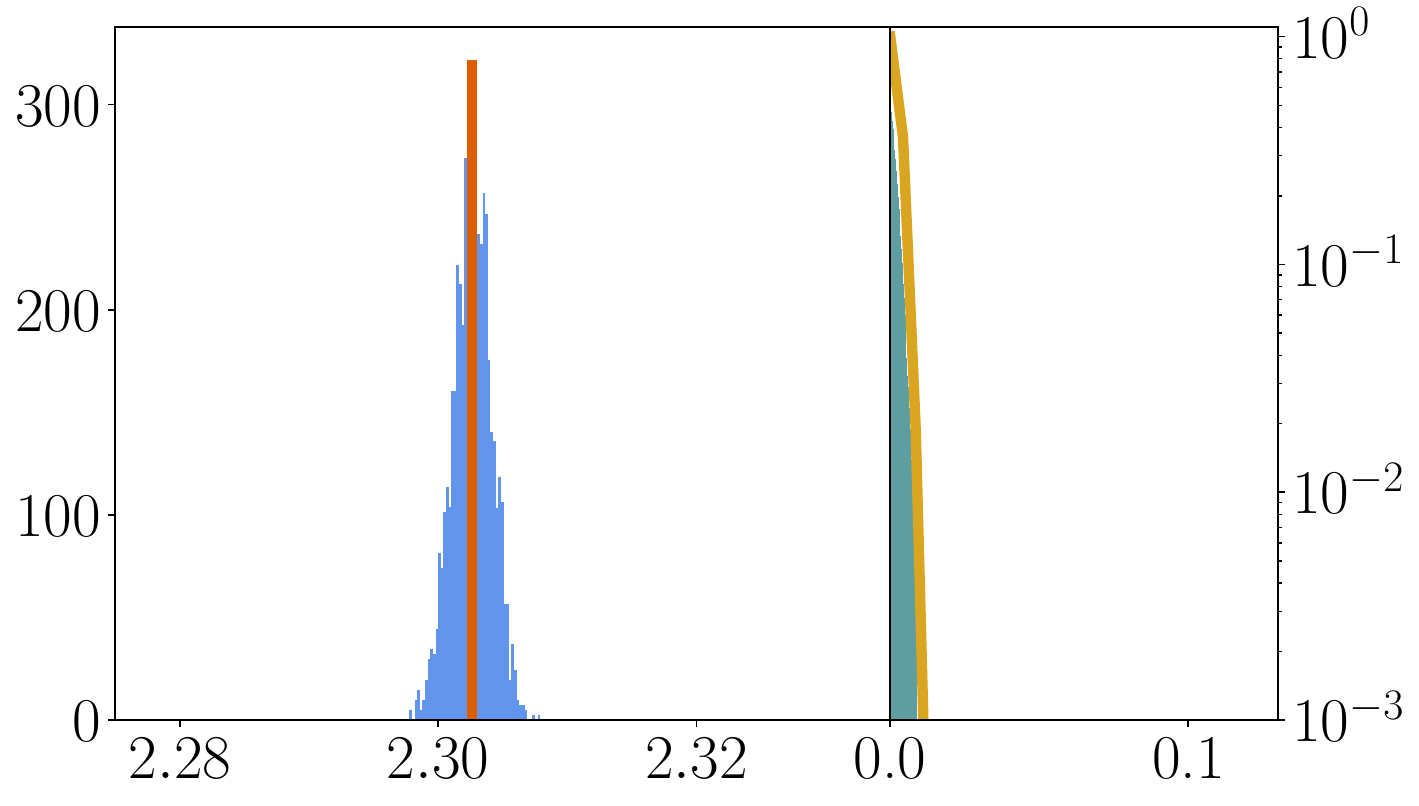}
      \end{subfigure}
        \medskip
        \begin{subfigure}[c]{.7\linewidth}
        \includegraphics[width=\linewidth]{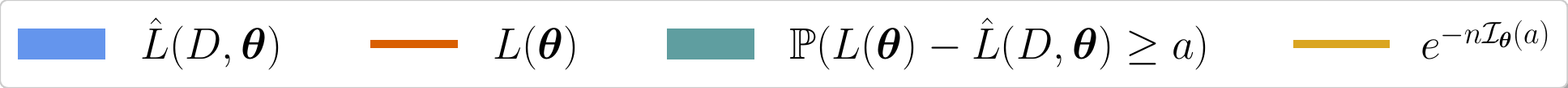}
      \end{subfigure}
    \caption{Illustrations on the distribution of \(\Lhat\) for data sets of size \(n = 50\), its concentration around the generalization error \(\L\) and Chernoff's inequality for increasing values of \(a \in [0, \L)\). All these quantities have been approximated using the test set of Cifar10 data set for 3 of the Inception models used in Figure~\ref{fig:1}.}
    \label{fig:histograms}
    \end{figure}
    
    Let us briefly illustrate why the rate function is useful to understand the generalization of interpolators. Figure \ref{fig:1} displays an estimation of the rate function $\I{\cdot}$ for some neural network models used in \cite{zhang2017understanding}. This work demonstrated that, within the same model class, interpolators that merely memorize the training data (as represented by ``Random'' in the figure, which has been learned using a random-labeled training data set) can co-exist with others that generalize exceptionally well, as the ones learned using weight-norm regularization (labeled as ``L2'') and/or data-augmentation techniques (random-cropping labeled as ``Crop''). According to Chernoff's bound, models with a larger rate function are less likely to have significant disparities between their expected and empirical losses; in other words, the empirical loss $\Lhat$ is more concentrated around its mean $\L$. Figure~\ref{fig:histograms} illustrates this fact for three of these models by plotting histograms that showcase the distribution of $\Lhat$ across various data sets of size $n=50$ (retrieved from the test set). From the histograms, it is clear that the concentration of $\Lhat$ varies among the models; the initial model, defined by \textit{Kaiming or He initialization} \citep{goodfellow2016deep}, has a prominent rate function. For this model, $\Lhat$ is tightly concentrated around its mean, $\L=\ln 10$, observable through the minute scale of its x-axis. Comparatively, the \textit{Standard} model, trained using SGD and characterized by a smaller rate function, has a more dispersed distribution around its mean $\L=0.65$. This dispersion is notably wider than that of the \textit{L2-Crop} model. The latter, also trained with SGD but incorporating both $\ell_2$ regularization and data-augmentation, has a larger rate function than the \textit{Standard} model.
    
    The main point of Figures \ref{fig:1} and \ref{fig:histograms} is to illustrate how $\Lhat$ is a random variable that could be either highly or poorly concentrated around its mean $\L$, something which is captured by the rate function of the model. When a model interpolates the training data, we must think we are observing a realization of this random variable $\Lhat$ for a particular data set $D$. Consequently, when comparing two models that interpolate the training data, we should choose the one that is more concentrated around its mean (higher rate function); because it will likely have a smaller expected loss. For example, Figure \ref{fig:histograms} (left) and (center) describe the distribution of $\Lhat$ for two models, Standard and L2-Crop, respectively. As both models interpolate the training data, we should prefer L2-Crop because $\Lhat$ is more concentrated around its expected value for that model.
    

\section{The Rate Function Characterizes the Generalization of Interpolators}\label{sec:smoothness}
    In this section, we showcase the main results of this work, Theorems~\ref{thm:LDTinv} and \ref{thm:smallerL}, which show how the (inverse) rate function can characterize the generalization performance of interpolators. Furthermore, we present a new notion of smoothness of a model which perfectly aligns with the following statement: \emph{a (sufficiently) smoother interpolator generalizes better}.

    \subsection{Tight Distribution-Dependent Bounds for Over-parameterized Interpolators}\label{sec:bounds}
    
    As discussed in the introduction, current high-probability generalization bounds in the form of Equation~\eqref{eq:generalbounds} have been unable so far to explain the generalization of over-parameterized interpolators. Different works \citep{nagarajan2019uniform,nagarajan2021explaining,gastpar2023fantastic,wang2024near} have also shown that the problem lies in the impossibility of having tight generalization bounds solely depending on the training data for over-parameterized model classes. \cite{gastpar2023fantastic} even concludes that \textit{``a bound without explicit distributional assumptions is likely to be not tight''}. Thus, an open question raises:
    \begin{OpenQuestionBox}[label=op:bounds]{Open Question}
    Are there tight distribution-dependent bounds for over-parameterized models? 
    \end{OpenQuestionBox}
    
    The answer to this question does not seem to be straightforward given that \cite{nagarajan2019uniform} also showed examples of distribution-dependent bounds (\cite{nagarajan2019uniform}'s Definition 3.3) which are not tight. However, the following result shows an uniform-convergence bound which applies simultaneously over the model class, 
    \begin{restatable}[PAC-Chernoff Bound]{thm}{LDTinv}\label{thm:LDTinv} 
        With h.p. \(1 - \delta\) over \(D \sim \nu^n\), for all \(\bmtheta\in \bmTheta\), simultaneously, 
        \begin{equation*}\label{eq:upperbound}
            \L \leq \Lhat +  \Iinv{\textstyle \pn}\,.
        \end{equation*}
    \end{restatable}
    During the rest of this work, many results will resemble the same structure used in Theorem~\ref{thm:LDTinv}: ``\emph{with h.p. \(1 - \delta\) over \(D \sim \nu^n\), for all \(\bmtheta\in \bmTheta\), simultaneously\dots}''. This kind of enunciating is common in PAC theory but might be confusing to other readers. We want to highlight that the above theorem states that \emph{for any \(\delta \in (0,1)\), the event \(\text{\textbf{Event}}(\bmtheta,D, \delta):=\L \leq \Lhat +  \Iinv{\textstyle \pn}\) will hold for every \(\bmtheta \in \bmTheta\) unless you have drawn a ``bad'' data set \(D \sim \nu^n\); that is, the event holds with probability at least \(1-\delta\).} Mathematically:
    \begin{equation*}
    \forall \delta \in (0,1), \quad \P_{D\sim  \nu^n}\Big( \textstyle \bigcap \limits_{\bmtheta \in \bmTheta} \textbf{Event}(\bmtheta,D, \delta)\Big)\geq 1-\delta\,.
    \end{equation*}
    
    Many results will resemble this structure through the text where the ``event'' might be an inequality (\(A \leq B\)) as in Theorem~\ref{thm:LDTinv}, or an implication (\(A \implies B\)) as in Proposition~\ref{prop:boundtightness}.
    
    Going back to Theorem~\ref{thm:LDTinv}, in the above bound, $\Iinv{\textstyle \pn}$ defines a complexity measure for the model $\bmtheta$ in the context of a model class defined by $p$ parameters, the training data set of $n$ samples and, according to the own definition of $\Iinv{\cdot}$ (see Definition~\ref{def:inverseratefunction}), it also depends on the data-generating distribution $\nu$. From Proposition~\ref{prop:Rateproperties}, this complexity measure $\Iinv{\textstyle \pn}$ monotonically grows with the size of the model class and monotonically decreases with the level of confidence $\delta$ and the size of the training data. 

    It is common to find models, within the same model class, that define the same loss function, \(\ell(\cdot, \cdot, \bmtheta_1) = \ell(\cdot, \cdot, \bmtheta_2)\). For instance, in a Multi-Layer Perceptron (MLP), weights can be permuted in specific ways without altering model predictions, and models with zeros in the last layer produce the same predictions regardless of the weights in other layers. Consequently, when calculating the size of the model class, we can effectively ``exclude'' models that define the same empirical loss across different datasets. Let \(\bar{\bmTheta} \subseteq \bmTheta\) be a subset of the model class where, if \(\bmtheta, \bmtheta' \in \bar{\bmTheta}\), there exists a dataset \(D \sim \nu^n\) such that \(\Lhat \neq \Lhatprime\). The following result provides a refined bound, based on the size of \(\bar{\bmTheta}\).

        \begin{restatable}{cor}{corLDTinv}\label{cor:corLDTinv} 
        With h.p. \(1 - \delta\) over \(D \sim \nu^n\), for all \(\bmtheta\in \bmTheta\), simultaneously, 
        \begin{equation*}\label{eq:cor_upperbound}
            \L \leq \Lhat +  \Iinv{\tfrac{1}{n}\ln\tfrac{|\bar{\bmTheta}|}{\delta}}\,.
        \end{equation*}
    \end{restatable}
     
    For the remainder of this work, we keep the term $k^p$ in all of our results. However, in each case, we can replace $k^p$ with the typically smaller term $|\bar{\bmTheta}|$ by applying the result above. The following result shows that, when $n$ goes to infinity, the proposed complexity measure converges with rate \(1/\sqrt{n}\) to the (scaled) standard deviation of the loss function.

    \begin{restatable}{thm}{boundlimit}\label{thm:boundlimit} 
    For any \(\delta \in (0, 1)\) and any \(\bmtheta\in \bmTheta\), it verifies that
    \begin{equation*}\label{eq:boundlimit}
    \lim_{n\rightarrow\infty} \ \sqrt{n}\, \Iinv{\pn} = \sqrt{2\mathbb{V}_\nu( \ell(\bmy, \bmx, \bmtheta))\ln \tfrac{k^p}{\delta}}\,.
    \end{equation*}
    \end{restatable}

    The relevant property of this novel complexity measure is that it is a \textit{perfectly tight proxy} of the expected loss $\L$ for models that interpolate the training data, $\Lhat\leq \epsilon$, even if the model class is over-parameterized,
    \begin{restatable}{prop}{boundtightness}\label{prop:boundtightness} 
    With h.p. \(1 - \delta\) over \(D \sim \nu^n\), for all \(\bmtheta\in \bmTheta\), simultaneously, 
    \begin{equation}\label{eq:boundtightness}
    \text{if } \quad \Lhat\leq \epsilon \quad\text{then}\quad  0 \leq \L - \Iinv{\textstyle \pn}  \leq \epsilon\,.
    \end{equation}
    \end{restatable}
    Note that in the above result , the event that holds with probability \(1-\delta\) is the implication in Equation~\ref{eq:boundtightness}. Using there results, we can derive an algorithm-distribution-dependent bound that is \textit{perfectly tight} for any data distribution and any learning algorithm; even for over-parameterized model classes. However, for this bound to be tight it is imperative that the model interpolates the training data. Formally, for any fixed \(\epsilon>0\), with \(2^{\mathcal{X} \times \mathcal{Y}}\) denoting the power set of all possible data sets of any size; let \(\mathcal{A}_\epsilon: 2^{\mathcal{X} \times \mathcal{Y}} \to \bmTheta\) be an algorithm that takes any data set and returns a model \(\bmtheta \in \bmTheta\) such that its training loss is lower than \(\epsilon\). That is, for any training data set \(D \in 2^{\mathcal{X} \times \mathcal{Y}}\), \(\mathcal{A}_\epsilon\) verifies that ${\hat L}({\cal A}_{\epsilon}(D),D)\leq \epsilon$. This kind of algorithms are quite common in machine learning where the model space \(\bmTheta\) is big enough that there exists models capable of memorizing random labels \citep[Theorem 1]{zhang2017understanding}. Then, 
    for any ${\cal A}_{\epsilon}$ algorithm, with h.p. \(1 - \delta\) over \(D \sim \nu^n\), 
    \begin{equation*}
    L({\cal A}_{\epsilon}(D)) \leq \hat{L}({\cal A}_{\epsilon}(D),D) +  {\cal I}^{-1}_{{\cal A}_{\epsilon}(D)}\big(\textstyle \pn\big)\leq L({\cal A}_{\epsilon}(D)) + \epsilon\,.
    \end{equation*}
    This provides an answer to the Open Question~\ref{op:bounds} using this newly presented bound:
   \begin{HighlightBox}
        PAC-Chernoff bounds are perfectly tight for (over-parameterized) interpolators.
   \end{HighlightBox}
    
    The question now is whether this distribution-dependent bound is useful to understand the generalization of algorithms retrieving (over-parameterized) interpolators. A first insight from the above bound is that the (inverse) rate function of the retrieved interpolator defines its generalization error. To the best of our knowledge, this is the first (distribution-dependent) complexity measure characterizing the generalization error an interpolator even in the context of a over-parameterized model class. In the next section, we explore further how the rate function can be used to precisely characterize which are the interpolates that better generalize.

    \subsection{Smoother Interpolators Generalize Better}
    
    As mentioned in the introduction, an open question in machine learning is the following,
    \begin{OpenQuestionBox}{Open Question}
    Given two models $\bmtheta \in \bmTheta$ and \(\bmtheta' \in \bmTheta'\), both successfully interpolating the training data, which of them generalizes better?
    \end{OpenQuestionBox}
    
    In this section, we introduce a formal result that provides the following answer to the above question: \emph{the smoother interpolator is the one that achieves better generalization, given that it is sufficiently smoother}. The condition of being \textit{smoother} is defined in terms of the rate function of the models. We contend that a model $\bmtheta$ is \textit{smoother} than a model $\bmtheta'$ if, for any $a>0$, observing a deviation larger than $a$ between the expected (test) loss and the empirical (train) loss is consistently smaller for $\bmtheta$ than for $\bmtheta'$. This is formalized as follows,  
    
    \begin{definition}\label{def:smoothness}
        Given a data-generating distribution $\nu$ and a loss function $\ell$, a model $\bmtheta\in\bmTheta$ is \(\beta\)-smoother than a model $\bmtheta'\in\bmTheta'$ if 
        \begin{equation*}
            \forall a\in(0,\beta]\quad \I{a} \geq \Iprime{a}\,.
        \end{equation*}
    \end{definition}
    Notice that if $\bmtheta$ is \(\beta\)-smoother than $\bmtheta'$, it is \(\beta'\)-smoother for any \(\beta' \in (0, \beta]\). Furthermore, if $\bmtheta$ is \(\beta\)-smoother than $\bmtheta'$, it verifies that $\bmtheta'$ cannot be \(\beta'\)-smoother than $\bmtheta$ for any \(\beta' > 0\). This properties allow to reasonably compare degrees of smoothness between models in the same or different model spaces. It is important to notice that the concept of smoothness is defined in the context a given data-generating distribution and given loss function. 

    According to the above definition, the higher the rate function, the smoother the model when comparing it to others. Furthermore, according to Chernoff's bound (see Theorem~\ref{thm:LDT} and Equation~\eqref{eq:asympoticEquality}),  smoother models will have an empirical loss $\Lhat$ more concentrated around its expected value $\L$. In consequence, it is more unlikely to observe higher differences between \(\Lhat\) and \(\L\) in smoother models. In fact, the following result states that there is a correspondence between the variance of the model's loss and the newly introduced notion of smoothness.

    \begin{restatable}{prop}{smootherlimit}\label{prop:smootherlimit}
     For any $\bmtheta \in \bmTheta$ and $\bmtheta'\in\bmTheta'$,
    \begin{equation*}
       \mathbb{V}_\nu( \ell(\bmy, \bmx, \bmtheta)) \leq \mathbb{V}_\nu( \ell(\bmy, \bmx, \bmtheta' )) \iff \exists \beta > 0 \text{ s.t. } \bmtheta \text{ is } \beta\text{-smoother than }\bmtheta'\,.
    \end{equation*}
    \end{restatable}
    
    This result shows the relation between the smoothness and the variance the loss function between a pair of models \(\bmtheta\) and  \(\bmtheta'\). As a result, we can \emph{intuitively} understand this notion of smoothness as a \textit{generalization of the variance of the loss of a model}. 
    
    Using this definition of smoothness and the results presented in Section \ref{sec:bounds}, we can easily derive a result providing an answer to which interpolator generalizes better,  
    
    \begin{figure}
      \begin{subfigure}{.49\linewidth}
        \includegraphics[width=\linewidth]{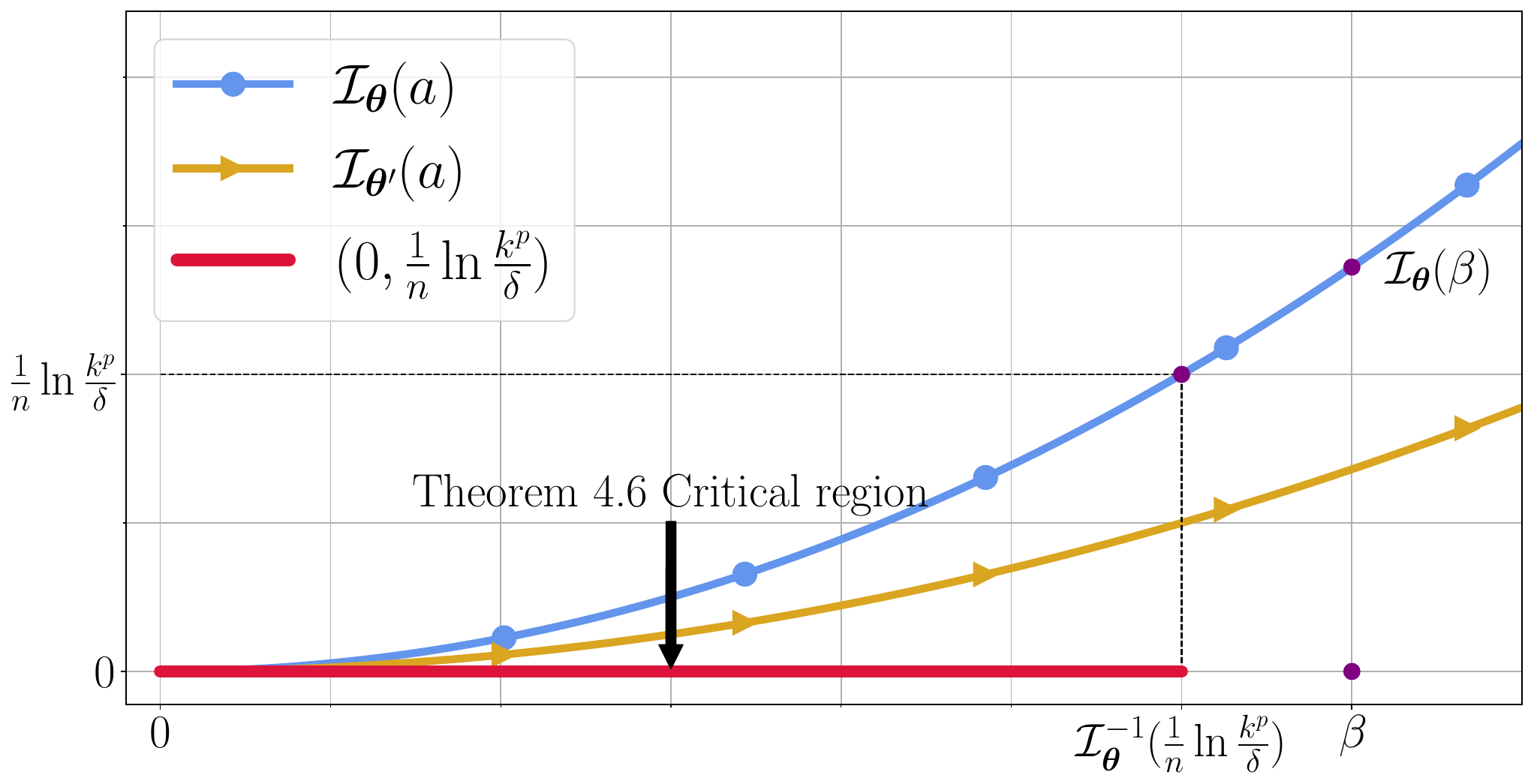}
      \end{subfigure}
      \begin{subfigure}{.465\linewidth}
          \includegraphics[width=\linewidth]{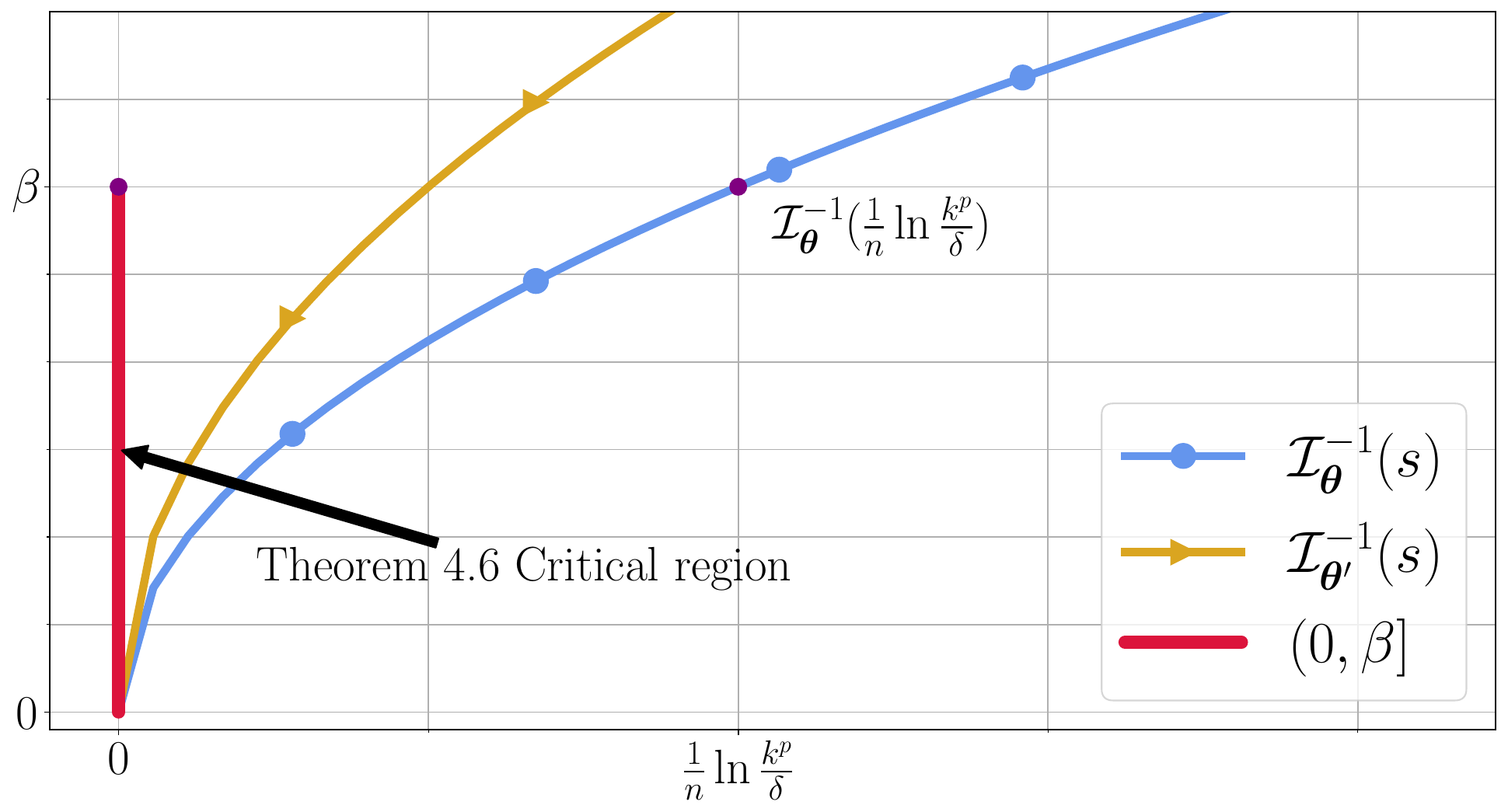}
      \end{subfigure}
        \caption{Visual illustration on Theorem~\ref{thm:smallerL}. The rate and inverse rate function of two models \(\bmtheta, \bmtheta' \in \bmTheta\) is shown with \(\bmtheta\) being \(\beta\)-smoother than \(\bmtheta'\) with \(\beta > \Iinv{\pn}\). This highlights the critical region that must fall below \(\beta\) for the theorem to hold.}
        \label{fig:thm:smoothness:smallerL}
    \end{figure}

    \begin{figure}
        \centering
      \begin{subfigure}{.95\linewidth}
          \includegraphics[width=\linewidth]{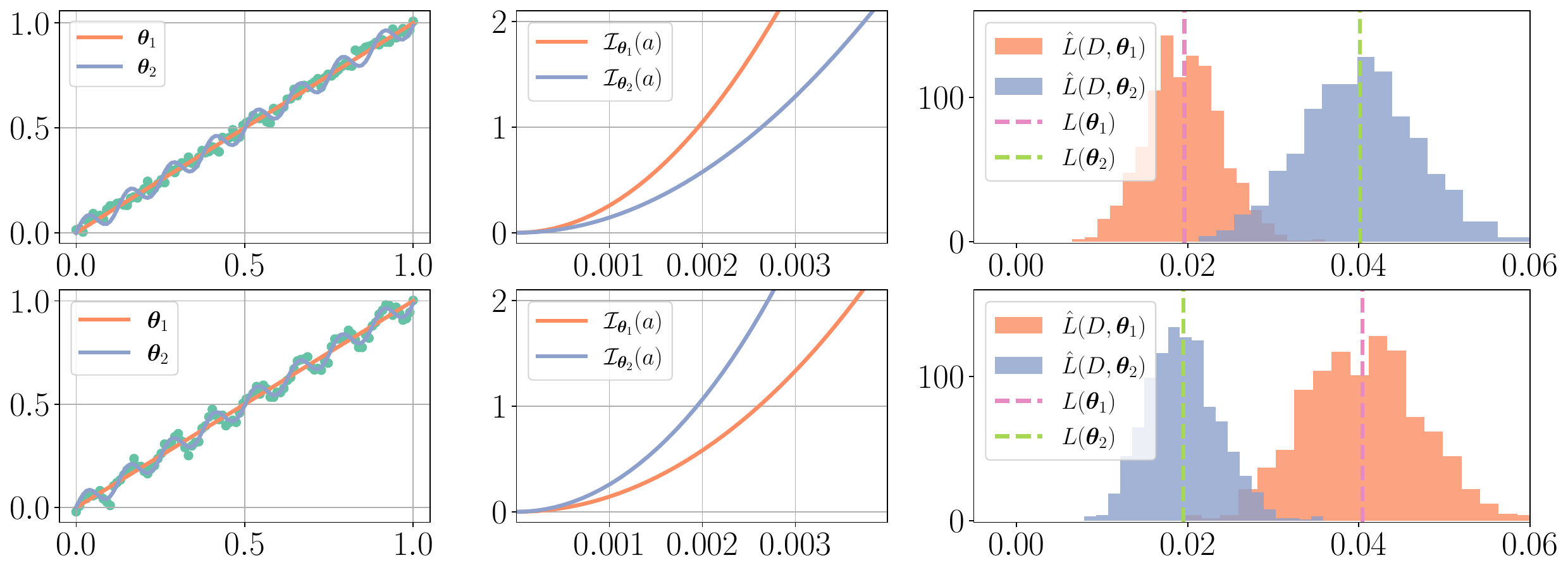}
      \end{subfigure}
        \caption{Visualization on the presented notion of smoothness. Two models, a linear model \(\bmtheta_1\) and a complex model \(\bmtheta_2\), are evaluated on two data-generating distributions \(\nu_1\) (first row) and \(\nu_2\) (second row). The first column shows the data and models, the second shows rate functions, and the third displays \(\hat{L}(D, \bmtheta)\) distributions.}
        \label{fig:toy}
        \vspace{-0.5cm}
    \end{figure}
    
    \begin{restatable}{thm}{smallerL}\label{thm:smallerL}
    For any $\epsilon\geq 0$, with h.p. $1-\delta$ over $D\sim\nu^n$, for all $\bmtheta \in \bmTheta \subset \mathbb{R}^p$ and $\bmtheta'\in\bmTheta'$, simultaneously, 
    \begin{equation*}
        \text{if $\Lhat\leq \epsilon$ and $\bmtheta$ is \ \(\mathcal{I}^{-1}_{\bmtheta} \big(\textstyle \pn\big)\)-smoother than $\bmtheta'$, then, $\L\leq L(\bmtheta')+\epsilon$}\,.
    \end{equation*}
    \end{restatable}

    Note that, due to Proposition \ref{prop:smootherlimit}, the smoothness condition given by ``\(\mathcal{I}^{-1}_{\bmtheta} \big(\textstyle \pn\big)\)-smoother'' in Theorem~\ref{thm:smallerL} can be understood as an inequality between the variances of the loss function under the hypothesis that \(n\) is sufficiently large. In fact, if  \(\mathbb{V}_\nu( \ell(\bmy, \bmx, \bmtheta)) \leq \mathbb{V}_\nu( \ell(\bmy, \bmx, \bmtheta' ))\), there exists \(\beta > 0\) such that \(\bmtheta \text{ is } \beta\text{-smoother than }\bmtheta'\). Then, if \(n\) is large enough to verify that \(\Iinv{\pn} \leq \beta\), we got that with h.p., if $\Lhat\leq \epsilon$, then, \(\L\leq L(\bmtheta')+\epsilon\). In short, Theorem~\ref{thm:smallerL} is a generalization of the following result: if \(n\) is large enough, with h.p.,
    \[
        \text{ if $\Lhat\leq \epsilon$ and } \mathbb{V}_\nu( \ell(\bmy, \bmx, \bmtheta)) \leq \mathbb{V}_\nu( \ell(\bmy, \bmx, \bmtheta' )) \text{, then, } \L\leq L(\bmtheta')+\epsilon\,.
    \]
    
    Theorem~\ref{thm:smallerL} states that an interpolator generalizes better than another (with h.p.) if it is \emph{sufficiently smoother} in terms of its rate function. Figure \ref{fig:thm:smoothness:smallerL} illustrates the premise of this theorem. We should note that the above result holds even for the \textit{log-loss}, which is the \textit{default} loss used for \textit{training}, and for over-parameterized model classes. This result is specially useful when $\epsilon$ is very small or null, as it states, with h.p., that \textit{smooth interpolators} generalize better, up to an  $\epsilon$, than other less smooth models; independently on whether these interpolate the data or not. Furthermore, the above result verifies that the higher the probability \(1- \delta\) or the number of parameters \(p\), the stronger the smoothness condition needs to be; and the opposite for larger $n$. In that sense, we could have interpolators which are $\beta$-smoother than others but have worse generalization performance, because they are not smooth enough in order to apply Theorem \ref{thm:smallerL}.

   \begin{HighlightBox}
    Interpolators with a larger rate function ${\cal I}_\bmtheta(\cdot)$ or, equivalently, smoother interpolators, are the ones that better generalize.
   \end{HighlightBox}
   
   Figure~\ref{fig:toy} shows a synthetical example to highlight how the notion of smoothness depends on the \emph{specific data-generating distribution}, and why one model is \textit{smoother} than another only relative to such distribution. The example considers two different data generating distributions \(\nu_1\) (first row, which adds random noise to a linear function) and \(\nu_2\) (second row, adds random noise to a complex sinusoidal function). The considered loss is the mean squared error. The introduced notion of smoothness using the rate function shows how \(\bmtheta_1\) (a linear model) is \textit{smoother} than \(\bmtheta_2\) (a more complex model) under \(\nu_1\). In fact, under \(\nu_1\), the distribution of \(\hat{L}(D, \bmtheta_1)\) is more concentrated around a smaller mean value. However, the second row of this example also shows how the more complex model \(\bmtheta_2\) can be smoother than a linear model \(\bmtheta_1\) under a different data-generating distribution \(\nu_2\) for exactly the same reasons. With this example we want to highlight that a good notion of smoothness (as the one presented here) should consider the data in which the models are being evaluated, rather than just the complexity of the function they induced.

    The results from this section clearly indicate that the generalization error of a interpolator is defined by its level of smoothness and its rate function. The question is whether this theoretical characterizations, which relies on distribution-dependent quantities, is useful for understanding the inner workings of current learning techniques and complex phenomenons appearing in deep learning. In Section \ref{sec:doubledescent}, we show how the PAC-Chernoff bound and our smoothness criteria are powerful enough to analyze the so-called double-descent phenomenon \citep{belkin2019reconciling}. In Section \ref{sec:explicit}, we examine the relationship between the smoothness of a model and widely used regularization techniques such as the parameter norm of a model, distance from initialization, input-gradient norm, and Lipschitz constant. Many of these quantities have also been previously used as measures of smoothness or complexity of a model \citep{neyshabur2017exploring}. In Section \ref{sec:invariances}, we show why invariant architectures and data-augmentation induce smoother interpolators. Finally, in Section \ref{sec:Over-parameterization}, we examine how over-parametrization is a necessary condition for having smoother interpolators. In Section \ref{sec:conclusions}, we also elaborate on the need to stablish connections our definition of smoothness with current existing definitions \citep{keskar2017large,shalev2014understanding}.


\section{Understanding Double-Descent with PAC-Chernoff Bounds}\label{sec:doubledescent}
    
    The existing literature consistently demonstrates that interpolators with a larger number of parameters tend to perform better. A finding that defies traditional beliefs from classical statistical learning theory. Traditionally, it was assumed that increasing the number of parameters in a model would lead to higher overfitting and, consequently, poor generalization performance. However, this perspective has been challenged by the phenomenon of the double-descent curve \citep{belkin2019reconciling}, which exemplifies a shift in dynamics as models enter the interpolation regime. In this regime, an increase in parameters paradoxically leads to improved performance. This counter-intuitive behavior suggests that once models begin to interpolate, their generalization performance improves as they grow larger. 
    
    \begin{figure}
      \begin{minipage}[c]{0.35\textwidth}
        \centering
        \scalebox{0.72}{
        \begin{tabular}{cccccc}\toprule
          Size & Train Acc. & Test Acc. & Test NLL & Bound\\ \midrule
          \(4\)k  & \(43.82\%\) & \(42.74\%\) & \(1.56\) & \(2.93\)\\
          \(32\)k & \(70.84\%\) & \(59.86\%\)  & \(1.17\)& \(2.01\)\\
          \(85\)k & \(86.69\%\) & \(62.73 \%\)  & \(1.35\)& \(1.77\)\\
          \(163\)k & \(97.65\%\) & \(63.97 \%\) & \(1.82\)& \(1.94\)\\
          \(266\)k & \(99.84\%\) & \(65.19 \%\) & \(2.15\)& \(2.19\)\\
          \(395\)k & \(100.0\%\) & \(65.74 \%\) & \(2.28\)& \(2.30\)\\
          \midrule
          \(548\)k & \(100.0\%\) & \(67.27 \%\) & \(2.10\)& \(2.11\)\\
          \(727\)k & \(100.0\%\) & \(69.88 \%\) & \(1.84\)& \(1.86\)\\
          \(931\)k & \(100.0\%\) & \(69.11 \%\) & \(1.83\)& \(1.84\)\\
          \(1161\)k & \(100.0\%\) & \(69.82 \%\) & \(1.72\)& \(1.74\)\\
          \bottomrule
          \end{tabular}
          }
        \end{minipage}
      \hfill
      \begin{minipage}[c]{0.56\textwidth}
        \centering
            \includegraphics[width = 0.99\textwidth]{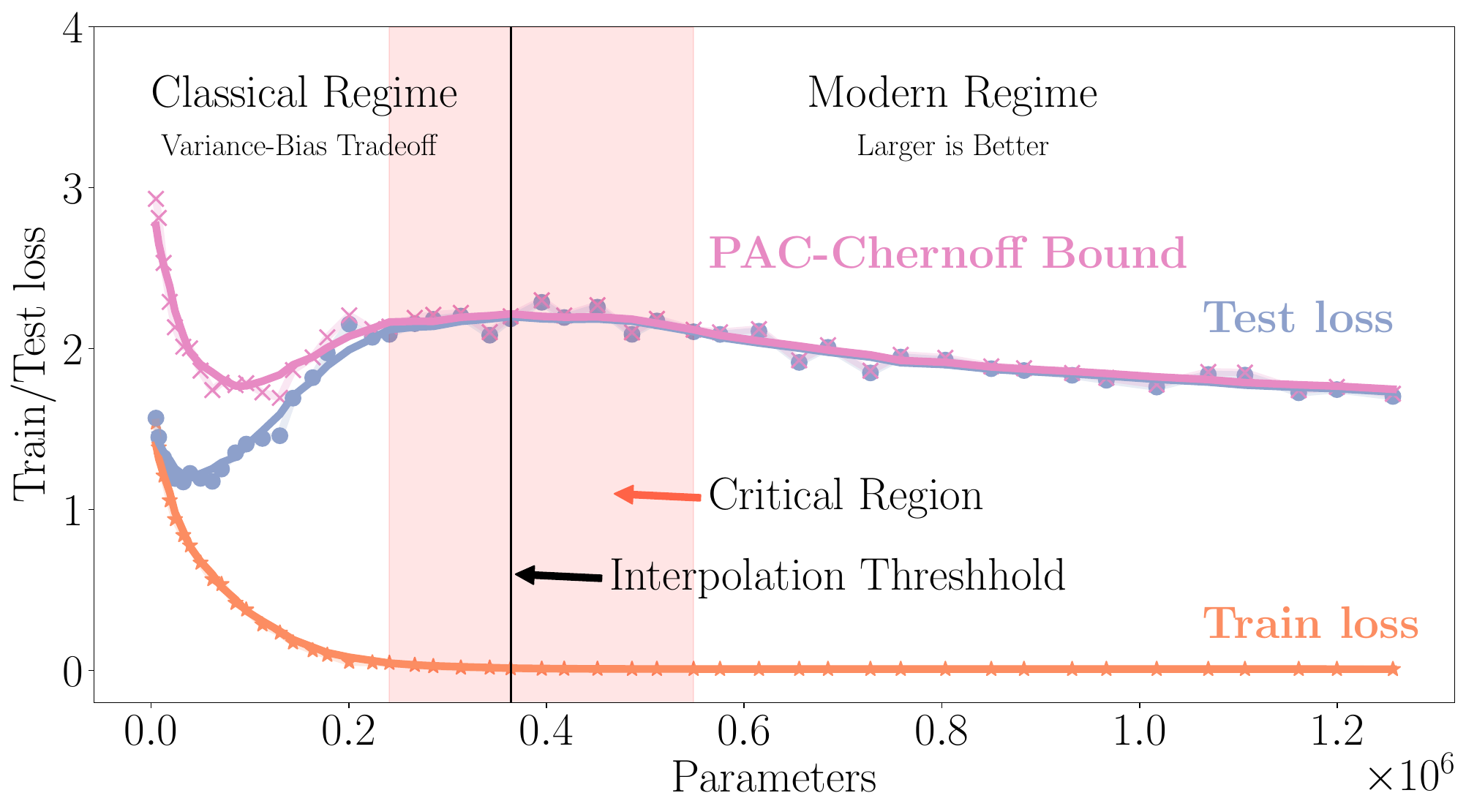}
        \end{minipage}
        \caption{Illustration of the \emph{double-descent phenomenon}: Evolution of the train loss, test loss and the proposed PAC-Chernoff Bound of nested convolutional neural networks with increasing number of parameters. Increasing the number of parameters promotes models with worse population loss until the critical region; further increasing the number of parameters decreases the population error. The proposed bound is perfectly tight for interpolators. The point where models reach the interpolation regime is marked with a black line. The training and test loss of models are marked with points and their smoothed curve is shown. Only a few of the total models are shown in the table.}
        \label{fig:double_descent}
    \end{figure}
    
    Figure \ref{fig:double_descent} illustrates this phenomenon by showing the training and the test loss of a sequence of convolutional networks with a growing number of parameters (implemented by adding more channels to the layers of a simple convolutional model). More precisely, models with a number of parameters ranging from 4k to 1161k\footnote{All models were found by running stochastic gradient descent on Cifar10's training data, until training loss reaches \(0.01\) or until it did not improve in two consecutive epochs of training.}. 
    
    Figure~\ref{fig:double_descent} also displays the evolution of our PAC-Chernoff bound for the whole range of models. As theoretically shown, this bound is perfectly tight for interpolators. In the classical regime, the bound also presents the so-called \emph{double descent phenomena}, even thought its tightness is not guaranteed. This highlights how distribution-dependent bounds are powerful enough to capture the complex dynamics emerging in the interpolation regime.
    
    \begin{OpenQuestionBox}[label=op:over1]{Open Question}
    Why does the generalization performance of interpolators improve with an increasing number of parameters?
    \end{OpenQuestionBox}
    
    \begin{figure}
        \begin{center}
      \begin{subfigure}{.475\linewidth}
        \includegraphics[width=\linewidth]{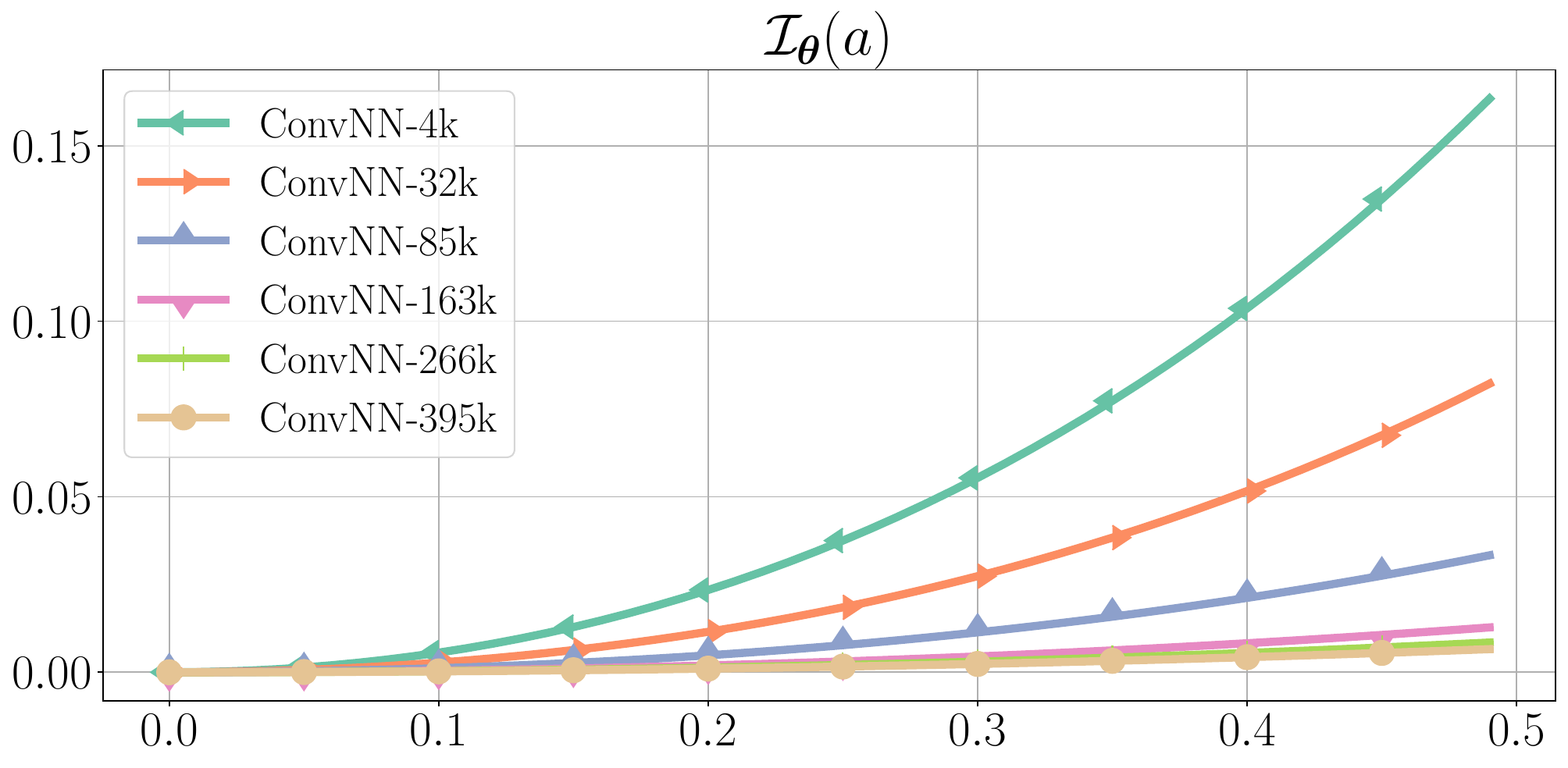}
      \end{subfigure}
      \begin{subfigure}{.49\linewidth}
        \includegraphics[width=\linewidth]{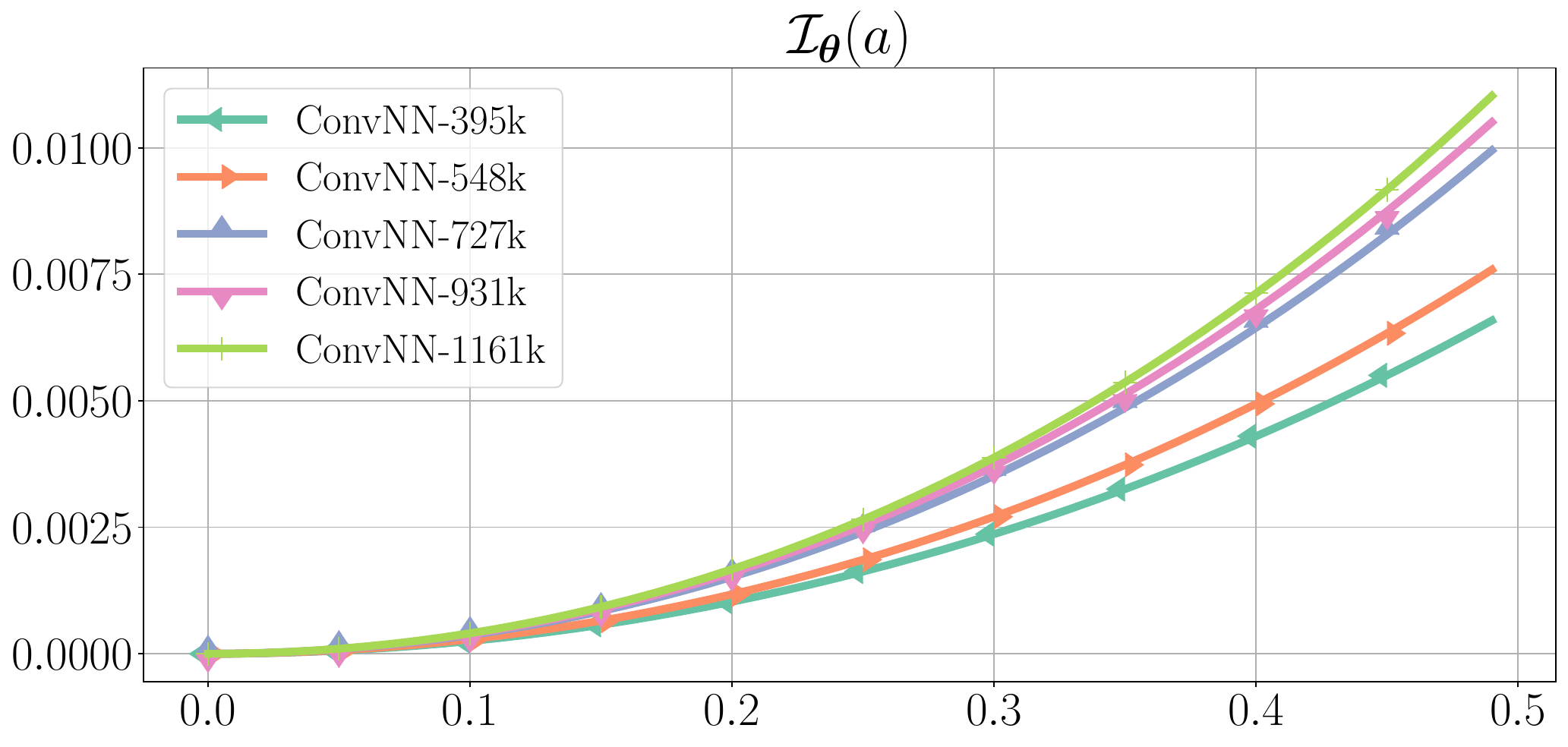}
      \end{subfigure}
    
      \end{center}
        \caption{Rate functions obtained by interpolating CNNs with 4k-1161k parameters on Cifar10. Models correspond to those in the table of Figure~\ref{fig:double_descent}. Parameters are increased by increasing the number of channels in the architecture, resulting in \emph{nested} model classes. Two images are shown where models are divided so that the rate function decrease on the left figure and increase on right one.}
        \label{fig:over-parameterization}
    \end{figure}
    The complexity term of our  own PAC-Chernoff bound \(\Iinv{\pn}\) monotonically increases with the size of the mode class, which, in principle, would contradict the \emph{double-descent phenomena} on the bound and the fact that generalization error is reduced as we increase the number of models. The following result, viewed as an extension of Theorem \ref{thm:smallerL}, offers a partial explanation for this puzzling phenomena. More precisely, it demonstrates that if the interpolators become progressively \textit{smoother} (higher rate function), their generalization error will be reduced, despite being part of a model class characterized by a greater number of parameters.

    \begin{restatable}{thm}{largermodelssmoothnesssecond}\label{thm:largermodelssmoothnesssecond}
        Let $\bmTheta\subset \bmTheta'$ be two nested model classes with $p < p'$ parameters respectively. For any $\epsilon>0$, with h.p. $1-\delta$ over $D\sim\nu^n$,  for any $\bmtheta'\in\bmTheta'$, $\bmtheta\in\bmTheta$, simultaneously
        \begin{equation*}
        \begin{aligned}
            \text{ if } \Lhatprime\leq\epsilon \,,\, \Lhat \leq \epsilon \, \text{ and }\,  \bmtheta' \, &\text{ is } \,\mathcal{I}^{-1}_{\bmtheta'}\big(\tfrac{1}{n}\ln \tfrac{k^{p'}}{\delta}\big)\text{-smoother than } \, \bmtheta\\
            &\Downarrow\\
            \Lprime\leq &\, \L+\epsilon\,.
        \end{aligned}
        \end{equation*}
    \end{restatable}
    Figure \ref{fig:over-parameterization} (right) illustrates the rate functions of the interpolators from Figure \ref{fig:double_descent}, highlighting how \textit{larger interpolators} become increasingly smoother, meaning their rate functions are progressively higher  (i.e., they are progressively smoother). This explains how it is possible to obtain larger interpolators with smaller generalization error. In fact, by reversing the implication of Theorem \ref{thm:largermodelssmoothnesssecond}, we can infer that if a larger interpolator exhibits better generalization performance, then, it cannot be \textit{less smooth} than a smaller interpolator.

    \begin{restatable}{cor}{largermodelssmoothness}\label{cor:largermodelssmoothness}
    Let $\bmTheta\subset \bmTheta'$ be two nested model classes with $p < p'$ parameters respectively. For any $\epsilon>0$, with h.p. $1-\delta$ over $D\sim\nu^n$, for any $\bmtheta'\in\bmTheta'$, $\bmtheta\in\bmTheta$, simultaneously, 
    \begin{equation*}
    \begin{aligned}
        \Lhatprime\leq\epsilon, \    \Lhat\leq&\,\epsilon \text{ and } \Lprime + \epsilon < \L\\
        &\Downarrow\\
        \bmtheta \text{ is not } \mathcal{I}_{\bmtheta}^{-1}\big(\tfrac{1}{n}\ln&\tfrac{k^{p'}}{\delta}\big)\text{-smoother than } \bmtheta'\,.
    \end{aligned}
    \end{equation*}
    \end{restatable}

    Consequently, Open Question \ref{op:over1} can be reformulated in terms of model smoothness rather than generalization performance.
    
    \begin{OpenQuestionBox}[label=op:over2]{Open Question}
    Why does the smoothness of interpolators improve with an increasing number of parameters?
    \end{OpenQuestionBox}
    
    Based on Theorem~\ref{thm:largermodelssmoothnesssecond} and Corollary~\ref{cor:largermodelssmoothness}, addressing Open Question~\ref{op:over2} is equivalent to answering Open Question~\ref{op:over1}. In other words, if we could answer why the rate function of the interpolators is increasingly higher as we increase the number of parameters in nested model classes, Theorem \ref{thm:largermodelssmoothnesssecond} will effectively explain why the double descent phenomena, and in particular, why the generalization performance of interpolators enhances as the number of parameters grows. However, to the best of our knowledge, there is no direct answer to this behavior without incurring in additional hypothesis on the matter.
    
    Our \textit{rationale} for explaining this phenomenon posits that, by increasing the number of parameters, larger neural networks are better at capturing invariances in the data, a notion that has been supported both theoretically \citep{bronstein2021geometric} and empirically \citep{goodfellow2009measuring} by the machine learning community. Additionally, as will be shown in Section \ref{sec:invariances}, this capability of capturing invariances leads to \textit{smoother} models with higher rate functions. Combining these two factors, Theorem \ref{thm:largermodelssmoothnesssecond} can help to elucidate why the generalization performance of interpolators improves with an increase in parameters. The experimental setup presented in Figures \ref{fig:double_descent} and \ref{fig:over-parameterization} support this understanding.
    
    Finally, it is also quite interesting to look at the models on the \textit{left part} of the \emph{interpolation threshold} of Figure \ref{fig:double_descent}. Surprisingly, the sequence of rate functions, displayed in Figure \ref{fig:over-parameterization} (left), get increasingly lower as the size of the mode class increases, just the opposite of what happens in Figure \ref{fig:over-parameterization} (right). In the classical regime, increasing the number of parameters leads to less smooth models with higher generalization error. However, explaining this behavior in the  non-interpolation regime is out of the scope of this work, as the presented PAC-Chernoff bound is \emph{not tight} in that regime and no theoretical guarantees can be derived in such framework. However, it is worth noticing that, even with the lack of tightness guarantees, the proposed bound is \emph{close enough} to showcase the double descent phenomena itself.

    In the following section, we will connect the proposed oracle bound with many existing regularization techniques, allowing to explain, in an unified framework, the effectiveness in terms of generalization.

\section{Explicit Regularization}\label{sec:explicit}

    Structural risk minimization \citep{shawe1998structural} is a learning principle based on \textit{regularizers} that penalize the complexity of a model when minimizing the training loss. Let $r(\bmtheta)$ denote a regularizing function, the learning objective is then given by,
    \begin{equation}\label{eq:structuralriskminimization}
        \min_{\bmtheta\in\bmTheta} \ \Lhat + r(\bmtheta)\,.
    \end{equation}

    High-probability bounds of the form given in Equation~\eqref{eq:generalbounds}, $\L\leq \Lhat + {\cal C}(\bmtheta,D,\delta)$, have been traditionally used to justify the use of regularizers and to derive novel ones. According to these bounds, an obvious choice for regularizes is the complexity term,  $r(\bmtheta) = {\cal C}(\bmtheta,D,\delta)$. Then, the structural risk minimization problem given in Equation~\eqref{eq:structuralriskminimization} would correspond to minimize a high-probability  upper bound over the expected loss $\L$. Many regularizing methods have been studied from this perspective. One of the most commons one is being the $\ell^2$-norm regularization and other variations based on spectral norms, which recurrently show up within the complexity term of many generalization bounds \citep{bartlett2017spectrally}. However, as discussed in the introduction, these bounds are known to be loose in the context of over-parameterized model classes interpolating the data. In consequence, the rationale of using their associated  complexity measures ${\cal C}(\bmtheta,D,\delta)$ as regularizes is no longer as strong as it used to be in learning setups where the model class is not over-parameterized. 
    
    \begin{OpenQuestionBox}[label=op:regularizer]{Open Question}
    Is there a regularizer \(r(\bmtheta)\) that ensures near-optimal performance for over-parameterized model classes interpolating the training data?
    \end{OpenQuestionBox}
    
    The following result provides an answer to Open Question~\ref{op:regularizer}. It shows that the complexity measure of the PAC-Chernoff bound given in Theorem \ref{thm:LDTinv} can be used as a regularizer. That is to say, using the inverse rate function $\Iinv{\pn}$, we get an \textit{optimal regularizer for over-parameterized interpolators}. More precisely, for any $\epsilon>0$, let $\bmtheta^\star_\epsilon$ and $\bmtheta^{\cross}_\epsilon$ denote the interpolator with the best generalization performance and the interpolator with the smallest inverse rate function, respectively,  
    \begin{equation*}\label{eq:optimalinterpolators}
        \bmtheta^\star_\epsilon = \argmin_{\bmtheta\,:\,\Lhat\, \leq \, \epsilon}\,  \L\,, \quad\quad \bmtheta^{\cross}_\epsilon = \argmin_{\bmtheta\,:\,\Lhat\, \leq \, \epsilon} \, \Lhat + \Iinv{\pn}\,.
    \end{equation*}

    This result shows that, with h.p., the expected loss of the interpolator with the smallest inverse rate function $\bmtheta^{\cross}_\epsilon$ is very close to the expected loss of the optimal interpolator, $\bmtheta^\star_\epsilon$. 
    
    \begin{restatable}{thm}{optimialinterpolator}\label{thm:optimialinterpolator}
    For any $\epsilon>0$, with h.p. $1-\delta$ over $D\sim\nu^n$, \(|L(\bmtheta^\star_\epsilon) - L(\bmtheta^{\cross}_\epsilon)|\leq \epsilon\).
    \end{restatable}
    
In simpler terms, favoring models with a small inverse rate function is the same as favoring models with a larger rate function, which essentially means choosing \textit{smoother models}. This finding highlights that the \textit{smoothest interpolator} not only fits the data but also performs nearly optimally when it comes to generalization.

   \begin{HighlightBox}
     The inverse rate function $\Iinv{\pn}$ is an \textit{optimal regularizer for over-parameterized interpolators}.
   \end{HighlightBox}

    \subsection{Connecting the Inverse Rate with Existing Regularization Techniques}
    
    We have shown that the inverse rate function, a distribution-dependent and oracle element, is an \textit{optimal regularizer} for interpolators. However, in this section we will show how a wide range of existing regularizers are tightly connected to the inverse rate and, in consequence, to our definition of smoothness. We do not aim to fully explore these connections, and we do not claim that the new bounds that we derive are better than existing ones. Our focus here is to show how the (inverse) rate function of a model can be used to unify many existing regularization techniques which were previously assumed to be unrelated. 
    
    \subsubsection{Norm \(\ell_2\) Regularization}
    This regularizer, also known as weight decay, is widely used by the ML community and it is known to improve generalization even in over-parameterized model classes interpolating the training data \citep{goodfellow2016deep}. The following result defines a connection between the inverse rate of a model and its norm, under the widely used assumption that the loss function \(\ell(\bmy, \bmx, \bmtheta)\) is Lipschitz with respect to the parameters of the model $\bmtheta$ \citep{li2019convergence}.
    
    \begin{restatable}{prop}{explicitlipschitz}\label{prop:explicitlipschitz}
        If the loss function \(\ell(\bmy, \bmx, \bmtheta)\) is Lipschitz w.r.t. \(\bmtheta\) with constant \(M > 0\), then, for any \(\bmtheta_0 \in \bmTheta_0 = \{\bmtheta \in \bmTheta \ | \ \V_{\nu}(\ell(\bmy, \bmx, \bmtheta)) = 0\}\), it verifies that
        \begin{equation*}
            \Iinv{\pn} \leq \sqrt{2Ma}\ \|\bmtheta - \bmtheta_0\|_2\,,
        \end{equation*}
        where \(a = min\big(1 , \pn \big)\).
    \end{restatable}

    In many common ML models, the null vector is a model with null variance, that is, $\mathbf{0} \in\bmTheta_0$, because the loss of this model is constant for any data sample. For example, in supervised classification problems with $K$ labels, a neural network with null weights has a constant loss equal to $\ln K$. In these cases, the above result shows that by promoting models with small parameter norm we achieve models with small inverse rate.  In consequence, the \(\ell_2\)-norm is a proxy to minimize the inverse rate and works as a regularizer. Figure \ref{fig:1} illustrates for our running example how the use of \(\ell_2\)-norm regularization leads to models with small parameter norm and larger rate function. This aligns with the existing evidence in the literature about the use of \(\ell_2\)-norm for successfully regularizing interpolators. 
    
    However, in the context of over-parameterized models interpolating the training data, it is well known that the \(\ell_2\)-norm of the parameters of the model does not correlate well with the generalization error \citep{jiang2019fantastic}. That is easy to explain if we consider that the \(\ell_2\)-norm just defines an upper-bound over the inverse rate, which, as shown in Section \ref{sec:bounds} is an optimal proxy for the generalization of interpolators. The gap of this upper-bound is the ultimate responsible of the failure of \(\ell_2\)-norm as a proxy for the generalization error of an interpolator. In fact, an open question in ML is:
    
    \begin{OpenQuestionBox}{Open Question}\label{op:regulizer2}
    When does the norm of an interpolator correlate with its generalization error?
    \end{OpenQuestionBox}
    
    The following result provides a characterization of a learning setup where these two elements are correlated.
    
     \begin{restatable}{cor}{explicitexponential}\label{cor:explicitexponential}
        Considering the log-loss, if it belongs to the exponential family with a constant base measure, that is, there exist \(s:\mathcal{X}\times \mathcal{Y} \to \mathbb{R}^p\), \(a:\bmTheta \to \mathbb{R}\) and \(k \in \mathbb{R}\) such that
        \begin{equation*}
            \ell(\bmy,\bmx,\bmtheta) := -\ln p(\bmy|\bmx, \bmtheta) = \bmtheta^T s(\bmx, \bmy) - a(\bmtheta) + k\quad \forall (\bmx, \bmy) \sim \nu\,;
        \end{equation*}
        Then, \(\forall \epsilon > 0\), exists \(n_0 > 0\) such that \(\forall n > n_0\):
        \begin{equation*}
            \left|\Iinv{\pn} - \sqrt{2\pn} \sqrt{\bmtheta^T \text{Cov}_\nu(s(\bmy, \bmx)) \bmtheta}\right|\leq \epsilon\,,
        \end{equation*}
      \noindent where $\text{Cov}_{\nu}(\cdot)$ is the covariance w.r.t. $\nu$ of the sufficient statistics of each $(\bmx,\bmy)$ sample.
    \end{restatable}
    
    This result shows that under the exponential family and a large enough number of data samples, the inverse rate is close to the scaled Riemannian norm of the model parameters under the positive definite matrix $\text{Cov}_\nu(s(\bmy, \bmx))$ that defines the Riemannian metric. Note that, in this case, the Riemannian norm will be an optimal regularizer as Theorem \ref{thm:optimialinterpolator} applies and the inverse rate is a perfectly tight proxy for the generalization error of interpolators as shown in Section \ref{sec:bounds}. The \(\ell_2\)-norm is obtained when the covariance matrix is a diagonal matrix with constant entries, a condition that requires the statistical independence of the components of the sufficient statistics \(s(\bmy, \bmx)\) with respect to the data-generating distribution. 
    
    In short, Proposition \ref{prop:explicitlipschitz} shows how the \(\ell_2\)-norm of the parameters of the model upper bounds its inverse rate and, then, it shows why biasing the optimizer towards models with smaller \(\ell_2\)-norm tends to reduce the generalization error of the interpolators. However, Corollary \ref{cor:explicitexponential} exemplifies a general setting to understand the gap in the inequality of Proposition \ref{prop:explicitlipschitz} and how this gap can manifest if $\text{Cov}_\nu(s(\bmy, \bmx))$, a distribution-dependent quantity, is very different from a diagonal matrix.
    
    \subsubsection{Distance From Initialization}
    
    The distance from initialization is known to be related to the generalization error of an interpolator \citep{nagarajan2019generalization}. In fact, it has been successfully used as a regularizer in \cite{hu2019simple}. In these works, an implicit assumption was that the initial models are produced through one of the well-established initialization schemes used in neural networks, such as Xavier or He initialization \citep{goodfellow2016deep}. These parameter initialization schemes provide an initial set of weights which guarantee that the variance of the activations across the different layers remain constant. In consequence,  the randomly initialized model makes almost constant predictions. If $\bmtheta_0$ denotes the initial parameter, we should see that $\V_{\nu}(\ell(\bmy, \bmx, \bmtheta_0))$ is very small. Figure \ref{fig:histograms} (right) shows how this is the case for a randomly intialized Inception model using He initialization. 
    
    Under the assumption that the initial model obtained by common initializers belongs to \(\bmTheta_0\), Proposition~\ref{prop:explicitlipschitz}  also shows a link between distance from initilizaton and the inverse rate function and, in consequence, with the generalization error of a model. The same reasoning used with  $\ell_2$-norm applies here: by promoting models with small distance from initialization we promote models with small generalization error. 
    
    Furthermore, as happens with the $\ell_2$-norm,  in the context of over-parameterized interpolators, distance from intialization is not a reliable proxy for the generalization error \citep{jiang2019fantastic}. Again, this is easy to explain if we consider that distance from intialization is just an upper-bound over of the inverse rate. In consequence, an open question in ML is:
    \begin{OpenQuestionBox}{Open Question}\label{op:regulizer3}
    When the distance from intialization of an interpolator perfectly correlates with its generalization error?
    \end{OpenQuestionBox}
    
    We provide again a characterization for the case where the cumulant generating function of the model $\J$ is a quadratic function or, equivalently, its second-order Taylor approximation around $\bmtheta_0\in\bmTheta_0$ is exact. In this case, as we show in the proof of Proposition~\ref{prop:Iinvnorm}, the cumulant generating function can be expressed as
    \begin{equation}\label{eq:secondorder}
        \J=\frac{1}{2}\lambda^2\big(\bmtheta - \bmtheta_0\big)^T \text{Cov}_{\nu} \big(\nabla_{\bmtheta} \ln p(\bmy|\bmx,\bmtheta_0)\big)\big(\bmtheta - \bmtheta_0\big)\,,
    \end{equation}
     where $\text{Cov}_{\nu}(\cdot)$ is the covariance w.r.t. $\nu$ of the gradient of the log-likelihood of each $(\bmx,\bmy)$ sample. Under this hypothesis, the inverse rate can be expressed as stated next.
     
    \begin{restatable}{prop}{Iinvnorm}\label{prop:Iinvnorm}
         For any \(\bmtheta \in \bmTheta\), if the equality of Equation~\eqref{eq:secondorder} holds, then 
        \begin{equation*}\label{eq:Iinv:Taylor}
           \textstyle \Iinv{\pn} = \sqrt{2\pn}\sqrt{\big(\bmtheta - \bmtheta_0\big)^T \text{Cov}_{\nu} \big(\nabla_{\bmtheta} \ln p(\bmy|\bmx,\bmtheta_0)\big)\big(\bmtheta - \bmtheta_0\big)}\,.
        \end{equation*}
    \end{restatable}
    
    When training  highly over-parameterized models using gradient descent, it is well known that parameters remain close to their initial values. This phenomenon is called lazy training and is a standard assumption in many theoretical analysis of deep neural networks \citep{jacot2018neural}. However, when lazy training takes place, the second-order Taylor approximation of the cumulant could become highly accurate. The above proposition then suggests how the Riemannian distance, induced by $\text{Cov}_{\nu} \big(\nabla_{\bmtheta} \ln p(\bmy|\bmx,\bmtheta_0)\big)$, would be an optimal proxy for the generalization error of an interpolator in the lazy training regime. This reasoning also helps to understand the limitations in this regard of the Euclidean-distance.

    \subsubsection{Input-Gradient and Lipschitz  Regularization}
    
    Input-gradient regularization is a technique used in deep learning to improve the robustness of neural network models \citep{gouk2021regularisation}. This approach focuses on the gradients of the model's output with respect to its input, which are indicative of how sensitive the model's predictions are to changes in the input data. 

    The following result shows how the inverse rate function and the expected norm of the input-gradient of a model are also related. This result is based on the \textit{log-sobolev inequalities} \citep{chafai2004entropies} and relies on the assumption that \(\nu(\bmy|\bmx)\) is deterministic, to simplify gradients with respect to the target variable, and that $\nu(\bmx, \bmy)$ is a uniformly strictly log-concave density, an assumption satisfied by a wide range of distributions. 
    
    \begin{restatable}{prop}{inputgradientIinv}\label{prop:inputgradientIinv}
    If $\nu(\bmx, \bmy)$ is an uniformly strictly log-concave density and \(\nu(\bmy|\bmx)\) is deterministic, then $\exists M > 0$, such that, for any \(\bmtheta \in \bmTheta\),
    \begin{equation*}
        \textstyle \Iinv{\pn} \leq \sqrt{\pn}\sqrt{M\E_\nu\Big[\big\Vert\nabla_{\bmx}\ell(\bmy,\bmx,\bmtheta)\big\Vert_{2}^{2}\Big]}\,.
    \end{equation*}
    \end{restatable}
    
    The above result explains why models with small input-gradient norm tend to have a small generalization error. And why a small generalization error may coexist with a large input-gradient norm, if the gap in the above inequality is large.
    
    From the above result, it is straightforward to derive a connection between the generalization of a model and its Lipschitz constant. If a model $\bmtheta$ is Lipschitz  with respect to the input data $\bmx$ with Lipschitz constant denoted $\text{Lip}(\bmtheta)$, then, by definition, the following inequality is verified:
    \begin{equation}\label{eq:Lip}
        \forall (\bmx,\bmy)\in\supp(\nu)\quad \|\nabla_{\bmx} \ell(\bmy,\bmx,\bmtheta)\|_2^2\leq \text{Lip}(\bmtheta)\,.
    \end{equation}
    
    The above inequality can be combined with Proposition~\ref{prop:inputgradientIinv} to derive a new upper bound on the inverse of the rate function and on the generalization error of a model:

    \begin{equation*}
        \textstyle \Iinv{\pn} \leq \sqrt{\pn}\sqrt{M\text{Lip}(\bmtheta)}\,.
    \end{equation*}
    
    In consequence, models with a small Lipschitz constant, will tend to have a small generalization error. This aligns fully with the existing literature \citep{neyshabur2017pac} and, also, explains the rationale behind regularization techniques that prevent overfitting by controlling the Lipschitz constant \citep{gouk2021regularisation}. 
        
    \subsubsection{Summary}
    
    In this section, we have theoretically characterized an optimal (distribution-dependent) regularizer for interpolators, even for over-parameterized model classes. This theoretical characterization can be used to understand why a wide range of commonly used regularization techniques promotes interpolators with a smaller generalization error. We are not aware of previous works able to characterize optimal regularizers for interpolators and establish such a wide range of connections with existing regularization methods in the machine learning community.
    
    We have also analyzed, in some specific but not less relevant situations, when could some of these common regularization methods be optimal for interpolators; even in over-parameterized model classes. In this regard, our work complements other related theoretical results, like the ones proposed by \cite{bartlett2020benign}, where the Euclidean norm was shown to be an optimal regularizer for linear regression interpolators under some specific assumptions.

    This section allows to draw connections between our definition of smoothness, as outlined in Section \ref{sec:smoothness}, and other existing definitions in the literature. We have shown how the \textit{smoothest interpolator} has near-optimal performance in Proposition~\ref{thm:optimialinterpolator}. The clearest connection with existing measures of smoothness arises from the characterization of the Lipschitz constant, as demonstrated by combining Equation \eqref{eq:Lip} and Proposition \ref{prop:inputgradientIinv}. The Lipschitz constant is frequently cited in the literature as a measure of a model's \textit{smoothness} \citep{gouk2021regularisation}. In this work, we have shown how models with a small Lipschitz constant have a small inverse rate and, in consequence, are smoother according to our own definition.

     
\section{Invariances}\label{sec:invariances}
    In many machine learning problems, we know that the input data $\bmx$ we observed have been \textit{transformed}, usually as a result of the \textit{measuring process}. For example, in the case of sensor readings, by random noise due to non-perfect sensors or, in the case of images, these may be rotated, blurred, etc, by the image capturing process \citep{bronstein2021geometric}. In this section, we discuss how widely used techniques in machine learning, like data augmentation \citep{shorten2019survey} and invariant architectures \citep{bronstein2021geometric}, can be jointly explained, using PAC-Chernoff bounds and the rate function, as techniques to improve learning under these \textit{transformed} inputs. 
     
    Let \(G\) denote a set of input-data transformations such that any $g \in G$ denotes a function, $g:{\cal X}\rightarrow{\cal X}$, defining a specific transformation. Assume there exists an \emph{unknown distribution} $h$ over $G$. In image classification, $G$ could be the set of all possible rotations, translations and/or reflections.

    The central hypothesis in this section is that the generative process of the data follows the framework defined in the previous paragraphs.
    
    \begin{assumption}\label{assump:transformedinput}The data-generating distribution has the following structure:
    \begin{equation*}
            \begin{aligned}
                (i) \,\, \bmx_0 \sim \nu_0(\bmx_0), \quad (ii) \,\,y \sim \nu(\bmy|\bmx_0), \quad (iii) \,\,g \sim h(g), \quad (iv)\,\, \bmx = g(\bmx_0)\,,
            \end{aligned}
    \end{equation*}
    \noindent where $\bmx_0$ denotes the un-transformed and un-observed input, which is distributed according to $\nu_0$, $\bmx$ denotes the observed (transformed) input. See Figure~\ref{fig:datatransformations} (left) for a graph representation.
    \end{assumption}

    \begin{figure}
        \begin{subfigure}[b]{.4\linewidth}
            \centering
            \begin{tikzpicture}[scale=0.9, transform shape]
                \begin{scope}[every node/.style={circle,thick,draw,minimum size=0.9cm}]
                    \node (A) at (0,0) {$\bmx_0$};
                    \node (D) at (0,2) {$g$};
    
                    \node[fill={rgb:blue,1;white,4}] (B) at (2,1) {$\bmx$};
                    \node[fill={rgb:blue,1;white,4}] (C) at (2,-1) {$\bmy$};
                \end{scope}
                
                \begin{scope}[>={Stealth[black]},
                              every edge/.style={draw=black,very thick}]
                    \path [->] (A) edge (B);
                    \path [->] (D) edge (B);
                    \path [->] (A) edge (C);
                \end{scope}
            \end{tikzpicture}
        \end{subfigure}\quad
        \begin{subfigure}[b]{.45\linewidth}
            \centering
            \begin{tikzpicture}[scale=0.8, transform shape]
                \begin{scope}[
                every node/.style={font=\sffamily\small},
                main node/.style={circle,thick,draw, minimum size=1cm},
                ]
                    \node[main node] (A) at (0,0) {$\bmx_0$};
                    \node[main node] (G1) at (0,2) {$g_1$};
                    \node[main node] (B) at (2,1) {$\bmx_1$};
                    \node[main node] (G2) at (2,2.5) {$g_2$};
                    \node[main node] (C) at (4,1) {$\bmx_2$};
                    \node[main node] (G3) at (4,2.5) {$g_{T-1}$};
                    \node[main node] (D) at (6,1) {$\bmx_{T-1}$};
                    \node[main node] (G4) at (6,2.5) {$g_{T}$};
    
                    \node[main node, fill={rgb:blue,1;white,4}] (F) at (8,1) {$\bmx_T$};
                    \node[main node, fill={rgb:blue,1;white,4}] (E) at (2,-1) {$\bmy$};
                    \node at ($(C)!.5!(D)$) {\ldots};
                    \node at ($(G2)!.5!(G3)$) {\ldots};
                \end{scope}
                
                \begin{scope}[>={Stealth[black]},
                              every edge/.style={draw=black,very thick}]
                    \path [->] (A) edge (B);
                    \path [->] (G1) edge (B);
                    \path [->] (G2) edge (C);
                    \path [->] (G3) edge (D);
                    \path [->] (G4) edge (F);
    
                    \path [->] (B) edge (C);
                    \path [->] (D) edge (F);
                    \path [->] (A) edge (E);
                \end{scope}
            \end{tikzpicture}
        \end{subfigure}
        \caption{Graph representation of Assumption~\ref{assump:transformedinput}. The left side represents the original assumption with a single transformation and the right side represents the \emph{generalized setting} with \(T\) transformations. Observed variables are marked with a blue-ish background.} 
        \label{fig:datatransformations}
    \end{figure}
    
    Under this assumption, the target value of an input is sampled \emph{before} its transformation, that is, \emph{transforming the inputs does not affect its label}. In statistical terms, under Assumption \ref{assump:transformedinput}, we have that the target random variable $\bmy$ is conditionally independent of $\bmx$ given $\bmx_0$. This fact perfectly aligns with many common settings, where, for example, rotating, reflecting or cropping an image does not alter the label of the object been represented in it \citep{bronstein2021geometric}. This assumption is weaker than the usual ones used in many related works \citep{chen2020group,bronstein2021geometric}, where the conditional generating distribution is assumed to be invariant under any transformation. These works assume that $\forall g\in G$, $\nu(\bmy|\bmx)=\nu(\bmy|g(\bmx))$. This assumption does not hold, for example, when considering rotation over digits, as the \textit{true label} of the image of a \emph{six} rotated 180 degrees \textit{changes to a nine}. 
    
    Consider the generalized scenario where several transformations can be applied, that is, \(g \in G\) is the composition of other transformations \(g(\bmx_0) = g_T \circ \cdots \circ g_1(\bmx_0)=\bmx\). We may denote \(G_1, \dots, G_T\) the sets of all possible individual transformations, with $G=\cup_t G_t$. Notice that the value of \(T\) is arbitrary and can differ from one transformation \(g \in G\) to another; however, to simplify the notation we will use the same \(T\) where one may consider that some \(g_i \in G_i\) are the identity. Then, we may consider the non-fully-transformed inputs \(\bmx_t = g_t\circ\cdots \circ g_1(\bmx_0)\), with $0<t<T$. The following result shows that the mutual information between the targets and the inputs, denoted $MI(\bmy\, ; \bmx)$, decreases the further the input is transformed.

    \begin{restatable}{prop}{mi}\label{prop:mi}
    Under Assumption \ref{assump:transformedinput}, let \(\bmx_t\) denote the random variable denoting a transformation of an input \(\bmx_0\) using a composition of \(t\) transformations, then
    \begin{equation*}
    MI(\bmy\, ; \bmx_{t})\leq MI(\bmy\, ; \bmx_{t-1}) \quad \forall t \in \mathbb{N}^+\,.
    \end{equation*}
    As a consequence, \(MI(\bmy; \bmx) = MI(\bmy; \bmx_T) \leq MI(\bmy; \bmx_0)\).
    \end{restatable}

    Going back to the main objective of this work, which is to study the generalization of interpolators, the rate function and PAC-Chernoff bounds can be directly applied to explain why learning on transformed data \(\bmx\) makes interpolators have worse generalization error compared to ``ideally'' learning on the untransformed data \(\bmx_0\), which is never observed. In the following sections, we will use this argument to explain the reason why invariant architectures and data augmentation promote models with better generalization errors by exploiting the fact that the inputs are transformed. 
    
    \begin{OpenQuestionBox}{Open Question}\label{op:transformedinputs}
    Why models interpolating transformed data, as described in Assumption \ref{assump:transformedinput}, have a higher generalization error? 
    \end{OpenQuestionBox}
    
\begin{figure}
        \centering
      \begin{subfigure}[b]{.37\linewidth}\centering \hspace{1.1cm}Loss distribution\\[2.5mm]
        \includegraphics[width=\linewidth]{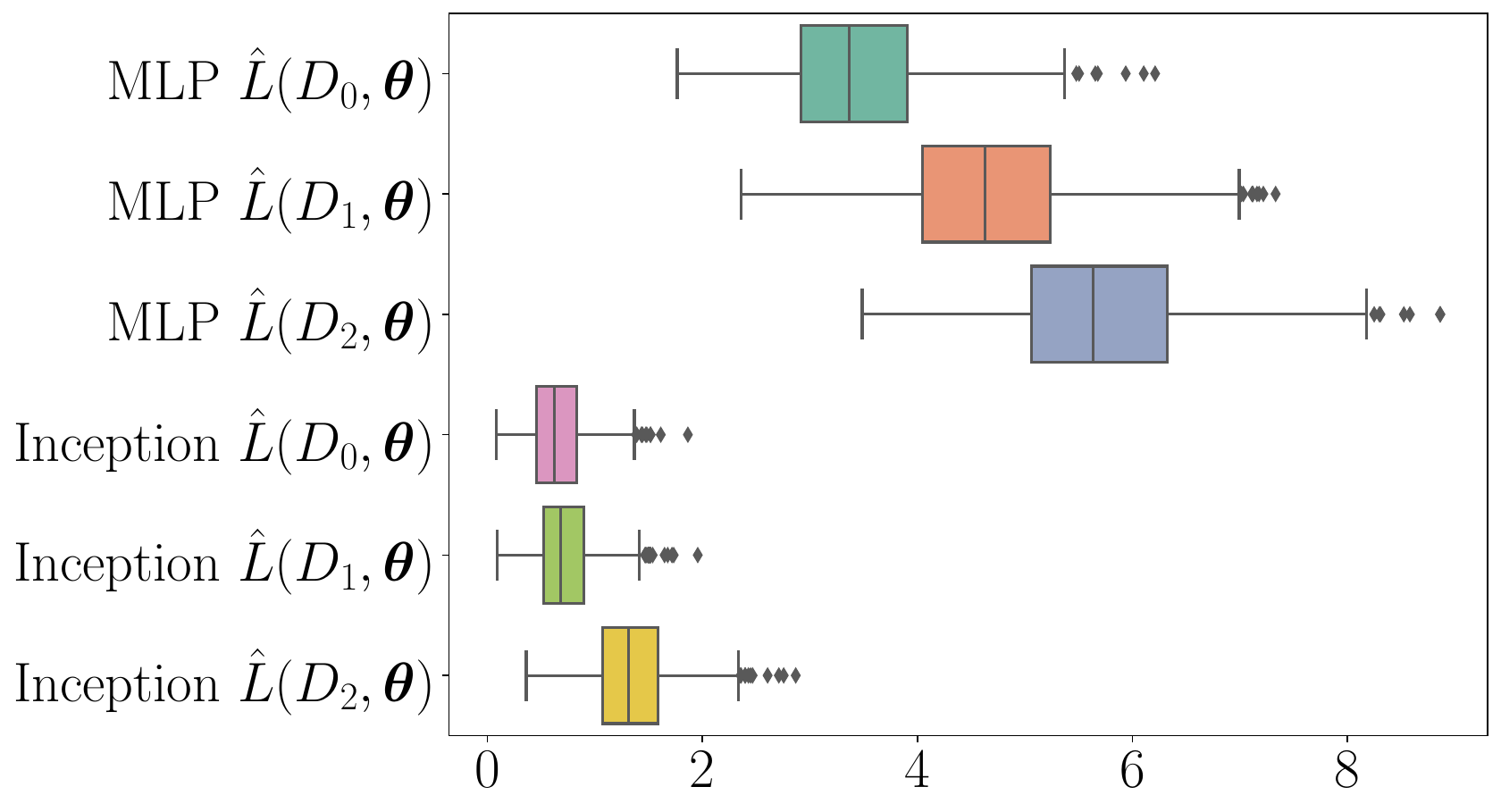}
      \end{subfigure}
      \begin{subfigure}[b]{.30\linewidth}\centering \hspace{0.3cm}Rate functions MLP\\[2mm]
        \includegraphics[width=\linewidth]{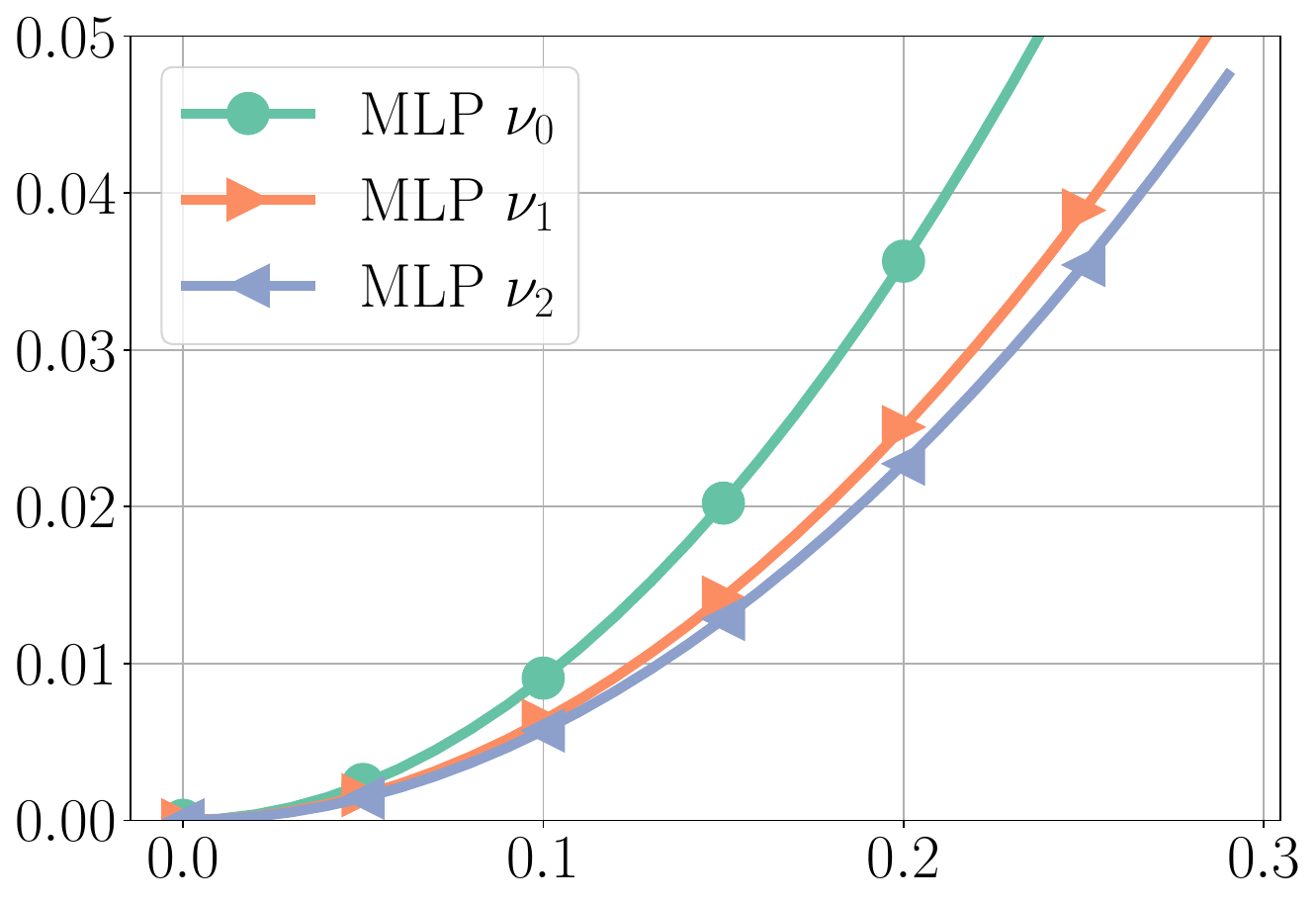}
      \end{subfigure}
      \begin{subfigure}[b]{.30\linewidth}\centering \hspace{0.1cm} Rate functions Inception\\[1.5mm]
        \includegraphics[width=\linewidth]{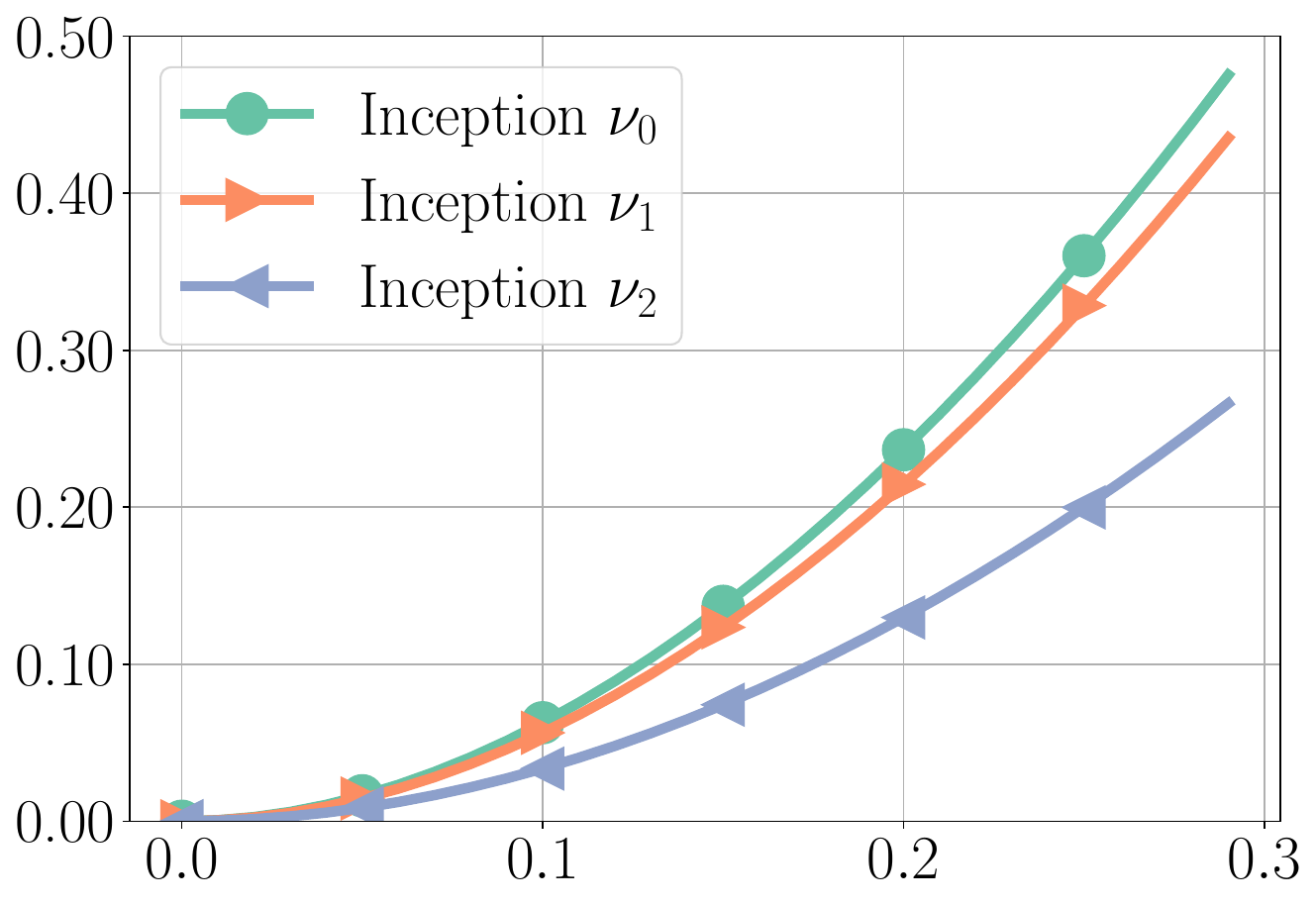}
      \end{subfigure}
    \caption{Illustration on the effect of transformed inputs. The distribution of \(\Lhat\) is shown on a fixed Multilayer perceptron and a fixed Inception model with similar amount of parameters for three different learning setups. For \emph{MLP} and \emph{Inception}, \(D_0 \sim \nu_0^{50}\) using Cifar10's test set; transformed inputs using random translations of \(3\) pixels, (\(D_1\sim \nu_1^{50}\)); and, transformed inputs with additional rotations up-to \(20 ^{\circ}\) are done (\(D_2\sim \nu_2^{50}\)).}
    \label{fig:invariance}
    \end{figure}

    Given a data-generating distribution, it is clear that the probability distribution of \(\Lhat\) rules the generalization error of the model \(\bmtheta \in \bmTheta\). Proposition~\ref{prop:mi} shows that learning on transformed variables reduces the information that the inputs \(\bmx\) have about their target \(\bmy\). As a result, one may expect that under  transformed inputs, \(\Lhat\) becomes less concentrated and with a higher expectation. Figure~\ref{fig:invariance} illustrates how this is the case for a Multi-Layer Perceptron and an Inception model on Cifar-10. In Figure~\ref{fig:invariance}  (left) we can see the distribution of \(\hat{L}(D_0, \bmtheta)\), \(\hat{L}(D_1, \bmtheta)\) and \(\hat{L}(D_2, \bmtheta)\) for datasets of size \(50\); where \(D_0\) is the unstransformed data, \(D_1\) corresponds to random translations applied to \(D_0\), and \(D_2\) corresponds to random rotations applied to \(D_1\). In short, these datasets are drawn from \(\nu_0, \nu_1 \) and \(\nu_2\) respectively. The figure shows how transforming the inputs once (translations) and twice (translations + rotations) increases the expectation and the scattering of the loss. This \emph{scattering} can be recognized by the size of the boxplots. As shown in Figure~\ref{fig:invariance} (center) and (right), the lower level of concentration of the distribution of $\hat{L}(\cdot, \bmtheta)$ under transformed inputs is also reflected by lower rate functions. Note that the same model can have different rate functions when applied over different data-generating distributions because the rate function depends on both the model and the data-generating distribution. Figure~\ref{fig:invariance} also shows that the effect of using transformed inputs can also be \emph{partly nullified} by using invariant architectures, as the case for Inception network which uses convolutional layers which are invariant to translations, $\nu_1$, but not to rotations, $\nu_2$. We will go deeper into this in the next subsection. 
    
    The observed changes in the distribution of the empirical loss $\hat{L}(\cdot, \bmtheta)$ under transformed inputs can be formalized in our framework by considering some specific assumptions. Although these assumptions can not be easily verified in general, the following results intend to provide \textit{theoretical justifications} for why transformed inputs make the distribution of the empirical loss $\hat{L}(\cdot, \bmtheta)$ less concentrated and with higher expectation. Formally, let \(\nu_t\) denote the marginal generating distribution of \(t\)-times transformed inputs \(\bmx_t = g_t\circ \cdots \circ g_1(\bmx_0)\), $L^{\nu_t}(\bmtheta)$ and $L^{\nu_{t+1}}(\bmtheta)$ denotes the expected loss of the model $\bmtheta$ under the data-generating distributions $\nu_t$ and $\nu_{t+1}$, respectively. Similarly, ${\cal I}_\bmtheta^{\nu_t}(a)$ and ${\cal I}_\bmtheta^{\nu_{t+1}}(a)$ denote the rate function of a model $\bmtheta$ under these two data-generating distributions. According to the following result, if the loss function is convex in $\bmx$ and the set of transformations satisfies that $\E_{g\sim h}[g(\bmx)]=\bmx$, then, transformed inputs, as described by Assumption \ref{assump:transformedinput}, make models have a higher expected loss.
    
    \begin{restatable}{prop}{invarianceconvex}\label{prop:invarianceconvex} Under Assumption \ref{assump:transformedinput}, if $\ell(\bmy,\bmx,\bmtheta)$ is convex under $\bmx$ and $\E_{g_t}[g_t(\bmx)]=\bmx$, then, \(\forall\bmtheta\in \bmTheta,\  L^{\nu_{t+1}}(\bmtheta)\geq L^{\nu_t}(\bmtheta)\).
    \end{restatable}

    The following result helps to explain why the empirical loss is less concentrated when the input-data is transformed as described in Assumption \ref{assump:transformedinput}. For any given input-data point $\bmx$, a given model incurs in some loss $\ell(\bmy,\bmx,\bmtheta)$, and when the same input-data point $\bmx$ is transformed by a randomly chosen transformation $g\sim h$, the model incurs in a different loss, $\ell(\bmy,g(\bmx),\bmtheta)$. This loss will be higher or lower than the loss over the untransformed input. The following result shows that the empirical loss under the transformed data is less concentrated if the relative change between the two losses, denoted by $\Delta(\bmy,\bmx,g,\bmtheta) = \ell(\bmy,g(\bmx),\bmtheta) - \ell(\bmy,\bmx,\bmtheta)$, is statistically independent of the loss itself, $\ell(\bmy,\bmx,\bmtheta)$. That is to say, how much the loss of the transformed-input changes does not depend of the loss of the unstransformed input itself. 
    
    \begin{restatable}{prop}{invarianceindependence}\label{prop:invarianceindependence} Under Assumption \ref{assump:transformedinput}, if the change observed in the loss after a transformation $\Delta(\bmy,\bmx,g,\bmtheta) = \ell(\bmy,g(\bmx),\bmtheta) - \ell(\bmy,\bmx,\bmtheta)$ is statistically independent of the loss itself, $\Delta(\bmy,\bmx,g,\bmtheta)\perp \ell(\bmy,\bmx,\bmtheta)$, then 
    \begin{equation*}
       {\cal I}^{\nu_{t+1}}_{\bmtheta}(a)\leq {\cal I}^{\nu_{t}}_{\bmtheta}(a)\quad \forall a>0\,.
    \end{equation*}
    \end{restatable}
    Summarizing, when input-data is transformed (Assumption \ref{assump:transformedinput}), the input-data provides less information about the target variable (Proposition \ref{prop:mi}). This causes the distribution of the empirical loss $\hat{L}(\cdot, \bmtheta)$ to be less concentrated and with higher expected loss $\L$ (as empirically shown in Figure \ref{fig:invariance} and theoretically argued in Propositions \ref{prop:invarianceconvex} and \ref{prop:invarianceindependence}). 

    \begin{HighlightBox}
        Transformed input-data makes the expected loss of the model $\L$ higher and the distribution of $\hat{L}(\cdot, \bmtheta)$ less concentrated.
    \end{HighlightBox}
    
    Finally, we re-state here the distribution-dependent bound given in Theorem \ref{thm:LDTinv} using the extended notation used in this section. With high probability over random draws $D_t\sim\nu_t^n$, simultaneously for all $\bmtheta\in\bmTheta$, then:
    \begin{equation}\label{eq:chernoffbound:transformeddata}
    \text{if $\hat{L}(D_t,\bmtheta)\leq \epsilon$ \ then \ \ }    \big({\cal I}^{\nu_t}_{\bmtheta}\big)^{-1}(\pn)\leq L^{\nu_t}(\bmtheta) \leq \big({\cal I}^{\nu_t}_{\bmtheta}\big)^{-1}(\pn) + \epsilon\,.
    \end{equation}

    The above bound easily explains why the presence of transformed input-data makes interpolators have larger generalization error. The presence of transformed input-data reduces the level of concentration of ${\hat L}(D_t,\bmtheta)$ and, in consequence, makes the rate function smaller (as shown in Figure \ref{fig:invariance} and Proposition \ref{prop:invarianceindependence}). Thus, the inverse rate function increases and so does \(({\cal I}^{\nu_t}_{\bmtheta})^{-1}(\pn)\); as a result, given that \(\epsilon\) is supposedly very small, the generalization error of the interpolator increases too. 
    \begin{HighlightBox}
        The PAC-Chernoff bound explains why interpolators of transformed inputs have higher generalization error.
    \end{HighlightBox}
    

    \subsection{Invariant Architectures}
    
    Invariant architectures refers to a kind of neural networks that remain unaffected by certain transformations in the input data \citep{bronstein2021geometric}. In case of images, these transformations might include scaling, rotation, translation, or other geometrical changes.
    
    \begin{definition}\label{def:invariantmodel}
        Given a set of transformations \(G\), a model \(\bmtheta \in \bmTheta\) is called \(G\)-invariant if 
        \begin{equation*}
            p(\bmy|\bmx, \bmtheta) = p(\bmy|g(\bmx), \bmtheta) \quad \forall g \in G\ \ \forall(\bmx, \bmy) \in \mathcal{X} \times \mathcal{Y}\,.
        \end{equation*}
    \end{definition}
    
    These invariant models, such as convolutional neural networks (CNNs) \citep{lecun1998gradient}, are crucial because they can recognize and process patterns irrespective of their position, scale, or orientation. This invariance is particularly important in tasks like image and speech recognition, where the essence of the data needs to be captured regardless of such variations. Invariant architectures have consistently shown superior performance compared to non-invariant models \citep{goodfellow2016deep}.
    
    The generalization error of invariant neural networks has been widely studied before. Different generalization bounds have been shown to have complexity terms which tend to shrink when the architecture of the model is invariant \citep{sokolic2017robust,elesedy2022group,behboodi2022pac}. Although these bounds provide theoretical grounds on why generalization error is reduced with invariant architectures, the problem lies in that they only show that the complexity measures of the respective bounds are reduced, not the generalization error. However, based on the arguments given in Section \ref{sec:intro}, there is a growing empirical \citep{zhang2017understanding} and theoretical evidence \citep{nagarajan2019uniform,gastpar2023fantastic,wang2024near}  that many of these bounds are non-tight, or, probably, even vacuous, and do not satisfy basic criteria such as the bound decreasing with larger dataset sizes, as specifically happens, for example, with the bound provided by \cite{behboodi2022pac}. In consequence, to the best of our knowledge, there are not widely accepted, specific generalization bounds that comprehensively captures the effect of the invariance of an architecture on a interpolator's generalization error, specially in over-parameterized model classes. An open question is then the following, 
    \begin{OpenQuestionBox}{Open Question}\label{op:invariantarchitectures}
    Why do (over-parameterized) invariant interpolators generalize better? 
    \end{OpenQuestionBox}
    
    Although the final generalization error of a model is also influenced by other factors such as the optimization process, we can provide a partial answer to this relevant open question by analyzing the rate function of invariant models. The key advantage of invariant architectures under transformed inputs (Assumption \ref{assump:transformedinput}) is that they are \textit{unaffected} by input-data transformations and, in consequence, they avoid all the problems described in the previous subsection. This can be formally stated in the following result,
    \begin{restatable}{prop}{invariance}\label{prop:invariance} Under Assumption \ref{assump:transformedinput}, if a model \(\bmtheta \in \bmTheta\) is \(G_t\)-invariant then, 
    \begin{equation*}
        L^{\nu_{t+1}}(\bmtheta) = L^{\nu_t}(\bmtheta) \quad \text{and} \quad {\cal I}^{\nu_{t+1}}_{\bmtheta}(a) = {\cal I}^{\nu_{t}}_{\bmtheta}(a)\quad \forall a>0\,.
    \end{equation*}
    \end{restatable}

    As argued in the above section, transformed-input data makes the empirical loss $\hat{L}(\cdot, \bmtheta)$ be less concentrated (with a lower rate function) and with a higher expected loss; however, the previous result shows that invariant architectures allow to bypass this problem, as the distribution of the empirical loss will be unaffected by transformed-input data. 

    Figure \ref{fig:invariance} clearly illustrates this point. Recall that $D_0 \sim \nu_0^{50}$ defines the distribution of the input data, and $D_1 \sim \nu_1^{50}$ the distribution after applying random translations over $D_0$. As expected, for a fixed MLP model, whose  architecture is non invariant to translations, the expected loss significantly increases and the rate function significantly decreases. While for a fixed \textit{Inception} model, the changes are minimal. Here, the \textit{Inception} model does not suffer a severe downgrade from the presence of translations as the MLP. The reason why \emph{Inception} does not report exactly the same results on \(D_0\) and \(D_1\), as predicted by Proposition \ref{prop:invariance} is because, even though convolutions and max-pooling are invariant to translations, in this experiment, translations were implemented by padding the image (for example with zeros), leading to slightly different results in the model output.

    The PAC-Chernoff bound of Equation~\eqref{eq:chernoffbound:transformeddata} neatly captures this phenomenon. As argued in the above section, the presence of transformed input-data implies that $\left({\cal I}^{\nu_1}_\bmtheta\right)^{-1}(\pn)\geq \left({\cal I}^{\nu_0}_\bmtheta\right)^{-1}(\pn)$. However, if the model $\bmtheta$ is invariant to the input-data transformations present in the data-generating distribution $\nu_1$, then, due to Proposition \ref{prop:invariance}, the complexity term of our bound does not increase. This, in combination with the fact that our distribution-dependent PAC-Chernoff bound is perfectly tight for interpolators, provides, to the best of our knowledge, the best theoretical explanation of why invariant models interpolating the training data have a smaller generalization error. 
    
    \begin{HighlightBox}
    The PAC-Chernoff bound explains why invariant interpolators have smaller generalization errors than non-invariant interpolators under transformed inputs.
    \end{HighlightBox}

    \begin{figure}
      \begin{minipage}[c]{0.4\textwidth}
        \centering
        \scalebox{0.74}{
        \begin{tabular}{ccccc}\toprule
          Model & Train Acc. & Test Acc. & Test NLL \\ \midrule
          Inception & \(100.0\%\) & \(74.08\%\) & \(1.00\) \\
          Inception-Shuffle& \(100.0\%\) & \(42.46\%\) & \(2.45\) \\
          MLP& \(99.99\%\) & \(51.69\%\) & \(3.29\) \\
          MLP-Shuffle& \(99.99\%\) & \(51.12\%\) & \(3.29\) \\
          \midrule
          Initial Inception& \(10.00\%\) & \(10.00\%\) & \(2.30\) \\
          Initial MLP& \(10.00\%\) & \(9.96\%\) & \(2.30\) \\
          \bottomrule
          \end{tabular}
          }
        \end{minipage}
      \hfill
      \begin{minipage}[c]{0.5\textwidth}
        \centering
            \includegraphics[width = 0.99\textwidth]{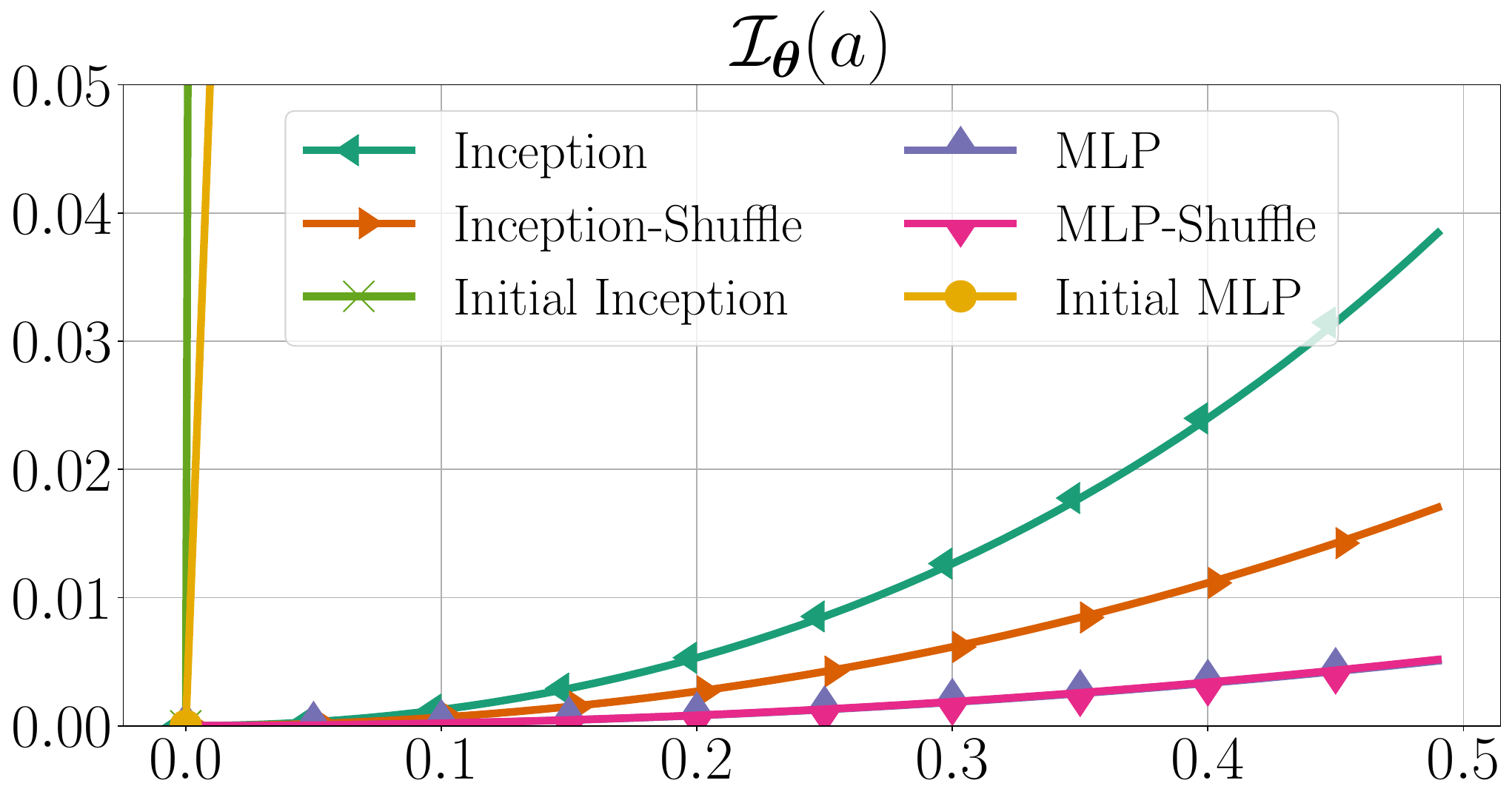}
        \end{minipage}
        \caption{Results of the random shuffling experiment. Metrics of trained Inception and MLP models with and without a fixed random shuffling of the inputs in both train and test datasets. The rate function of all these models is shown in the right, each of them w.r.t. their corresponding shuffled or un-shuffled data distribution they where trained and evaluated on.}
        \label{fig:rates}
    \end{figure}

    The results in Figure~\ref{fig:rates} attempt to illustrate this behavior in a different way. The specific amount of invariances of an Inception network are unknown; however, we know at least this model is invariant to translations by the usage of convolutional layers. In this experiment, we attempt to remove this invariance by performing a pre-fixed random shuffling of the pixels of the images. One can understand this as an extra input-data transformation \(g_{T+1}\) that performs a permutation of the pixels on top of the data-generating distribution \(\nu_{T}\) (we are implicitly assuming that Cifar10's data is generated by \(T\) unknown transformations of an unknown distribution \(\nu_0\)). As a result of this, a convolutional network like \textit{Inception} would no longer be invariant in the context of \(\nu_{T+1}\), at least, to image translations. According to the analysis at the beginning of the section, the empirical loss will be less concentrated under \(\nu_{T+1}\) and, in consequence, the rate function will be lower with respect to \(\nu_{T}\). In the figure, we can see this effect in the rate functions of the learned Inception models, under \(\nu_{T}\) (\textit{Inception} label) and under \(\nu_{T+1}\) (\textit{Inception-Shuffle} label)\footnote{Note that the rate functions are computed under the log-loss or negative log-likelhood (NLL).}.

    The same experiment is performed to a MLP, which presents no local invariances; as a result and as expected, the rate function of this model is not affected by the presence of the random permutation. In fact, although the model is not invariant to shuffling, the architecture is; in the sense that for any MLP \(\bmtheta\) and permutation \(g_{T + 1}\), there exists a MLP \(\bmtheta'\) which makes the same predictions on \(\nu_{T+1}\) that \(\bmtheta\) does on \(\nu_T\) (by permuting the weights in the first layer). Thus, a learning algorithm can find the ``same'' MLP under \(\nu_T\) and \(\nu_{T + 1}\). As a result, both MLP models present (nearly) the same rate function in the experiment. The table in this figure also shows how the generalization error is much higher for non-invariant models, as predicted by our theoretical analysis.

\subsection{Data Augmentation}\label{sec:dataaugmentation}

    Data augmentation is a widely-used technique in machine learning that enhances the robustness and generalization ability of models, particularly in the field of computer vision \citep{shorten2019survey}. By applying various transformations like rotations, reflections, scaling, and translations to existing data samples, data augmentation artificially expands the dataset. This process helps the learned models to become invariant or less sensitive to these transformations, leading to improved performance on unseen data. 

    Recent research has delved deeply into the theoretical foundations of data augmentation (DA), with significant contributions coming from studies such as \cite{chen2020group} and \cite{Lyle2020}. These works build on the assumption than the set of transformations $G$ forms a \textit{group} and that these transformations do not alter the data-generating distribution, also assuming \textit{an equality in distribution assumption}. Within this context, these studies illustrate how DA can decrease variance and improve bounds on generalization error. \cite{Lyle2020} further presents a PAC-Bayes bound based on these premises, showing that data augmentation can lower the bound's complexity measure. However, these studies encounter a critical challenge: the \textit{equality in distribution} assumption does not hold in reality, as it implies that any model $\bmtheta$ would yield identical expected losses on both transformed and non-transformed input data, a premise contradicted by evidence shown in Figure \ref{fig:invariance}, where two models exhibit deteriorating performance as the data is increasingly transformed. Additionally, the PAC-Bayes bound introduced by \cite{Lyle2020} is not proven to be tight for interpolators. Thus, as highlighted in the introduction, merely demonstrating that data augmentation reduces the complexity term of a bound, which might be vacuous, does not assure a lower generalization error for an interpolator.

    In this section, we demonstrate that our theoretical framework offers a more nuanced explanation. First, we will illustrate that data augmentation defines a new, more concentrated empirical loss. This finding is consistent with prior studies indicating that data augmentation decreases the variance of the empirical loss \citep{chen2020group,Lyle2020}. Second, we will explore how the complexity term of the PAC-Chernoff bound, introduced in Theorem \ref{thm:LDTinv}, becomes smaller with the application of data augmentation. As this bound is also perfectly tight for interpolators trained on augmented datasets, we can deduce that the generalization error of interpolators under data augmentation is reduced, in opposite to the non-tight bounds used in earlier research \citep{Lyle2020}.

    As argued, for example, by \cite{chen2020group}, data augmentation is equivalent to optimize, using Monte-Carlo estimates, the so-called \emph{data-augmented loss}, denoted as \(\ell_{G}\), which is the result of averaging the standard loss over all transformations obtained from a given input $\bmx$,
    \begin{equation}\label{eq:DA:loss}
        \ell_{G}(\bmy, \bmx, \bmtheta) = \E_{g\sim h}\left[  \ell(\bmy,  g(\bmx), \bmtheta)\right]\,.
    \end{equation}
    
    A pertinent question arises regarding whether the use of the augmented loss $\ell_{G}$ results in a more concentrated empirical loss (characterized by a larger rate function), given that more concentrated empirical losses lead to interpolators with a reduced generalization error. 

    The response to this inquiry is somewhat complex. Thus, in order to make the rest of the section easier to follow, we are summarizing now the key points of the discussion:
    \begin{enumerate}
        \item It is crucial to recognize that each transformed dataset, denoted $D_T$, corresponds to an original, unaltered dataset $D_0$, from which $D_T$ is generated through random transformations $g_t\sim h_t$ for $t=1,\ldots,T-1$. Two main objects are discussed: 
        \begin{enumerate}
            \item The conventional empirical loss for transformed inputs $\Lhat$, here referred to as $\hat{L}^{\ell}(D_T, \bmtheta)$ to highlight that $D_T\sim\nu^n_T$, and that we are using the standard non-augmented loss $\ell$.
            \item The augmented empirical loss for original, non-transformed inputs, referred to as $\hat{L}^{\ell_G}(D_0, \bmtheta)$, with $D_0$ being the non-transformed dataset, $D_0\sim\nu_0^n$, where $\hat{L}^{\ell_G}(D_0, \bmtheta)=\tfrac{1}{n}\sum_{\bmy_i,\bmx_i\in D_0} \ell_G (\bmy_i,\bmx_i,\bmtheta)$ and $\ell_{G}$ as defined in Equation \eqref{eq:DA:loss}.
        \end{enumerate}
        
        It is important to keep in mind that $\hat{L}^{\ell_G}(D_0, \bmtheta)$ is an unknown quantity as \(D_0\) is un-observed. However, $\hat{L}^{\ell_G}(D_0, \bmtheta)$ will play a fundamental role in explaining the underlying mechanics of data-augmentation.
        
        \item Theorem \ref{thm:da} will showcase, using the rate function, that the augmented empirical loss for non-transformed inputs $\hat{L}^{\ell_G}(D_0, \bmtheta)$ exhibits a higher concentration than the empirical loss for transformed inputs $\hat{L}^{\ell}(D_T, \bmtheta)$. As a result of this, and the fact that their expected value (population errors) are equal, minimizing  $\hat{L}^{\ell_G}(D_0, \bmtheta)$ is a \emph{better} optimization objective as observing lower generalization errors is more likely due to the Chernoff's bound in Theorem~\ref{thm:LDT}.
        
        However, in practice \(D_0\) is unobserved, and DA involves minimizing $\hat{L}^{\ell_G}(D_T, \bmtheta)$ rather than $\hat{L}^{\ell_G}(D_0, \bmtheta)$. Therefore, merely establishing that $\hat{L}^{\ell_G}(D_0, \bmtheta)$ is a better minimization objective than $\hat{L}^{\ell}(D_T, \bmtheta)$ is insufficient for a nuance understanding of DA. The following points discuss the link between $\hat{L}^{\ell_G}(D_T, \bmtheta)$ and $\hat{L}^{\ell_G}(D_0, \bmtheta)$.

        \item Proposition \ref{prop:group} will show that  $\hat{L}^{\ell_G}(D_T, \bmtheta)$ and $\hat{L}^{\ell_G}(D_0, \bmtheta)$ are equal when \(G\) is a group and \(h\) is an uniform distribution. As a result of this, an interpolator of $\hat{L}^{\ell_G}(D_T, \bmtheta)$ will interpolate for $\hat{L}^{\ell_G}(D_0, \bmtheta)$, which exhibits a higher concentration than $\hat{L}^{\ell}(D_T, \bmtheta)$.
        
        However, many DA techniques fall outside of this group hypothesis, such as rotations \emph{within a finite range} or random-cropping.

        \item Proposition \ref{prop:interpolatorsda} will reveal that both $\hat{L}^{\ell_G}(D_T, \bmtheta)$ and $\hat{L}^{\ell_G}(D_0, \bmtheta)$ share the same ``optimal interpolators'' (minimal error). In essence, if $\bmtheta$ serves as a perfect interpolator  for $\hat{L}^{\ell_G}(D_T, \bmtheta)$, then, it will similarly function as a perfect interpolator for $\hat{L}^{\ell_G}(D_0, \bmtheta)$.

         As a conclusion on the discussion, such an interpolator is likely to exhibit a lower generalization error compared to those minimizing $\hat{L}^{\ell}(D_T, \bmtheta)$, given that $\hat{L}^{\ell_G}(D_0, \bmtheta)$ is more concentrated than $\hat{L}^{\ell}(D_T, \bmtheta)$. As a result of this and Chernoff's bound, an interpolator minimizing $\hat{L}^{\ell}(D_T, \bmtheta)$ is more likely to exhibit a lower generalization error.
    \end{enumerate}

    In the following result, we use \(L^{\nu_0,\ell_G}(\bmtheta)\) and \(\mathcal{I}^{\nu_0,\ell_G}_\bmtheta(a)\) to denote the expected loss and rate function, respectively, associated to the augmented loss $\ell_G$ under the un-transformed data-generating distribution $\nu_0$. Recall that, under this notation, \(\L\) would be equal to \(L^{\nu_T,\ell}(\bmtheta)\) and, similarly, \(\I{a}\) would be equal to \(\mathcal{I}^{\nu_T,\ell}_\bmtheta(a)\), as they are implicitly defined over the standard loss \(\ell\) and the (observed) transformed data-generating distribution $\nu_T$, which is the data-generating distribution from which we observe the training data sets.

    \begin{restatable}{thm}{da}\label{thm:da}
        Under Assumption~\ref{assump:transformedinput}, it verifies that
        \begin{equation*}
            \forall \bmtheta\in\bmTheta,\quad L^{\nu_T,\ell}(\bmtheta) =L^{\nu_0,\ell_G}(\bmtheta)\quad and \quad \mathcal{I}^{\nu_T,\ell}_\bmtheta(a)\leq \mathcal{I}^{\nu_0,\ell_G}_\bmtheta(a)  \,\,\, \forall a>0\,.
        \end{equation*}
    \end{restatable}
    
\begin{figure}
        \centering
      \begin{subfigure}[b]{.39\linewidth}\centering \hspace{1.4cm}Loss distribution\\[2.5mm]
        \includegraphics[width=\linewidth]{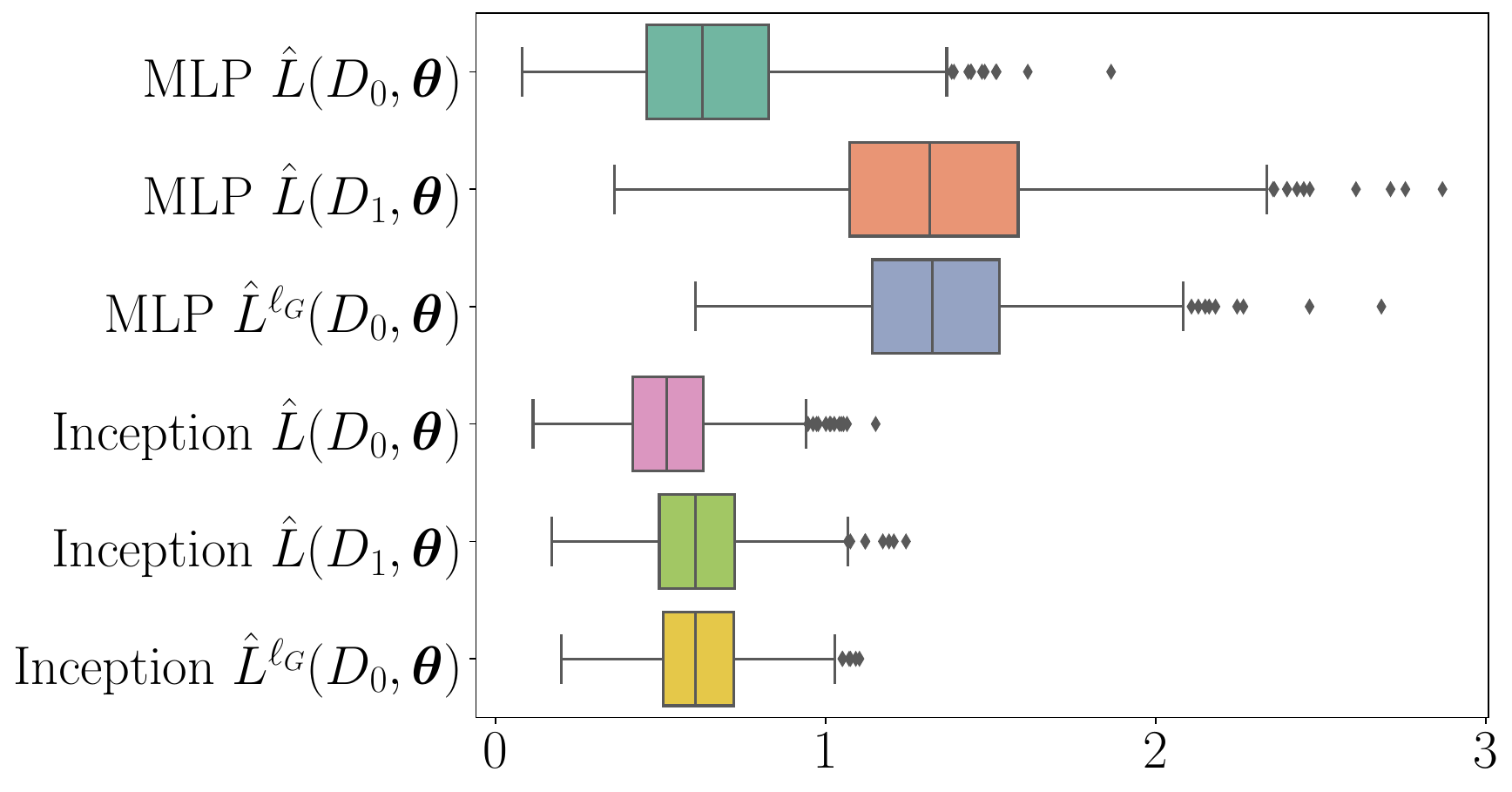}
      \end{subfigure}
      \begin{subfigure}[b]{.29\linewidth}\centering \hspace{0.3cm}Inception\\[1.5mm]
        \includegraphics[width=\linewidth]{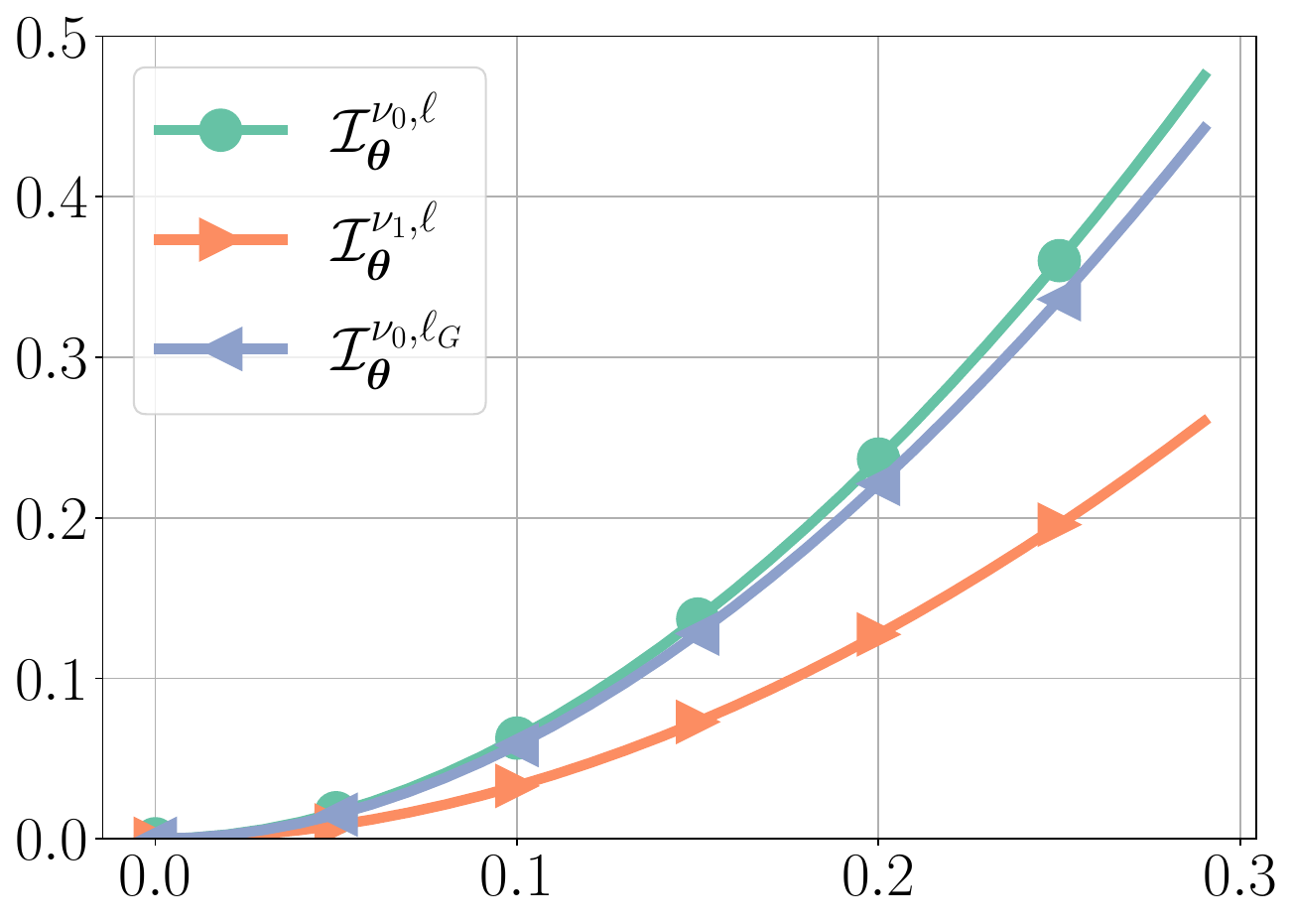}
      \end{subfigure}
      \begin{subfigure}[b]{.29\linewidth}\centering \hspace{0.1cm}Inception DA\\[1.5mm]
        \includegraphics[width=\linewidth]{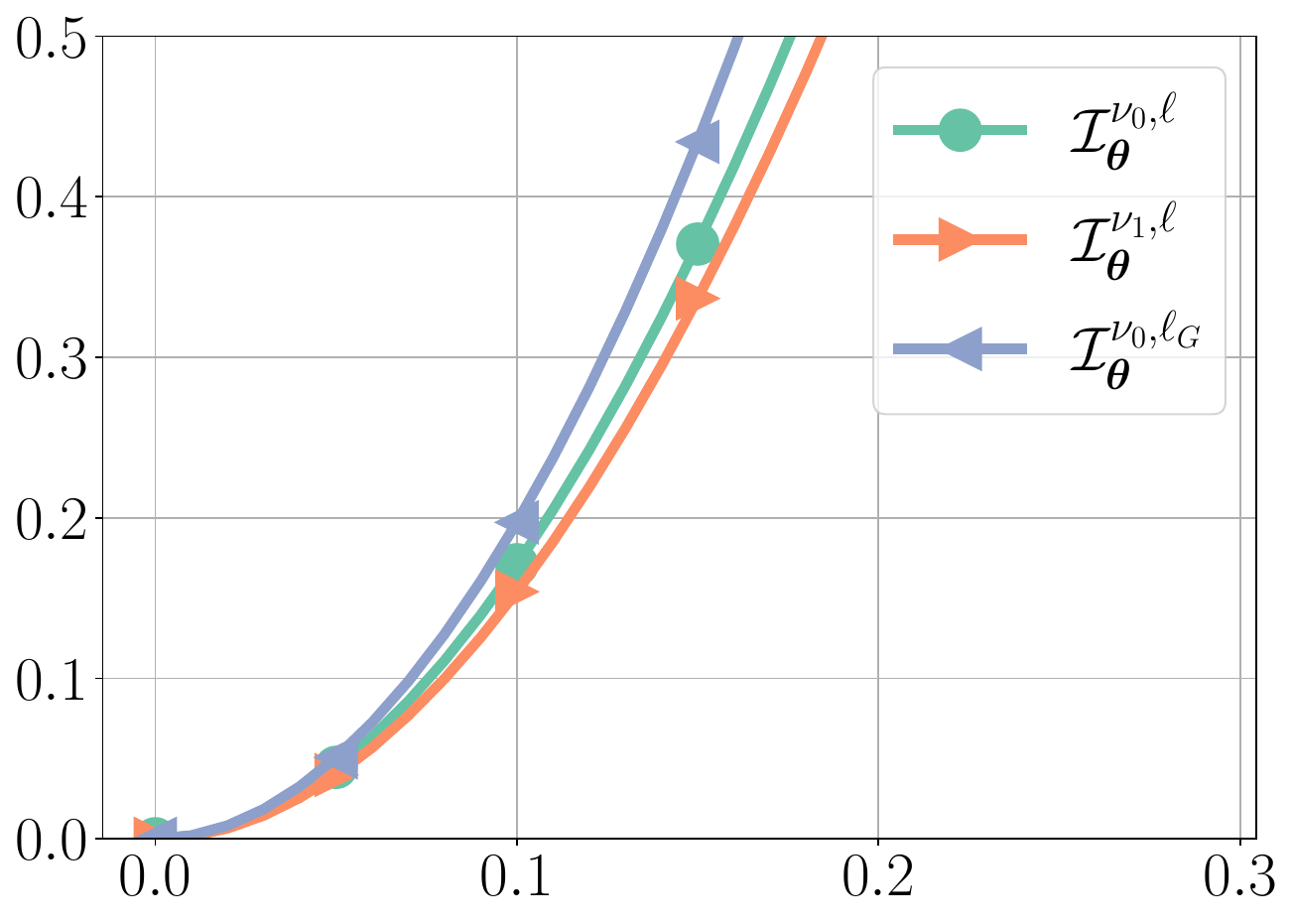}
      \end{subfigure}
    \caption{Illustration on the effect of data augmentation and transformed inputs. The distribution of \(\hat{L}(D_0, \bmtheta)\), \(\hat{L}(D_1, \bmtheta)\) and \(\hat{L}^{\ell_G}(D_0, \bmtheta)\) are shown on a fixed Multilayer perceptron and a fixed Inception model with similar amount of parameters for three different learning setups. For \emph{MLP} and \emph{Inception}, we use Cifar10's test set to simulate \(D_0 \sim \nu_0^{50}\); transformed inputs, (\(D_1\sim \nu_1^{50}\)), are simulated by applying random translations of \(3\) pixels, and rotations up-to \(20 ^{\circ}\) to  Cifar10's test set. The augmented loss \(\ell_G\) is computed using the same \textit{augmentations} used to generate \(D_1\). The obtained rate functions for the Inception model are shown in the middle figure. The last figure resembles the obtained rate functions with an Inception model trained using the augmented loss \(\ell_G\).}
    \label{fig:da}
    \end{figure}

    Figure \ref{fig:da} illustrates the above result using a fixed MLP and a fixed Inception model over CIFAR-10. Figure \ref{fig:da} (left) shows how the distribution of the empirical loss over transformed data $\hat{L}^{\ell}(D_1, \bmtheta)$ with $D_1\sim\nu^{50}_1$ is more dispersed than the distribution of the empirical loss over the un-transformed data $\hat{L}^{\ell}(D_0,\bmtheta)$ with $D_0\sim\nu^{50}_0$ for both models. And, also, how the distribution of $\hat{L}^{\ell_G}(D_0, \bmtheta)$ becomes more concentrated and with the same expectation than the distribution of $\hat{L}^{\ell}(D_1, \bmtheta)$. Figure \ref{fig:da} (center) also shows how the rate functions of the Inception  models on the three different setups changes as expected. We want to remark again that the models are fixed and the changes in the distribution of their empirical losses are due to changes in the data-generating distribution and/or the loss. 
    
    The following result is an adaptation of the PAC-Chernoff bound given in Theorem \ref{thm:LDTinv} for this setup, which describes the effect of using the data-augmented loss on the generalization error of interpolators. 
    
    \begin{restatable}{cor}{corda}\label{cor:corda}
        With h.p. \(1 - \delta\) over \(D_0 \sim \nu^n_0\), for all \(\bmtheta\in \bmTheta\), simultaneously, 
        \begin{equation*}
        \text{if } \hat{L}^{\ell_G}(D_0, \bmtheta) \leq \epsilon \quad\text{then}\quad  \big(\mathcal{I}^{\nu_0,\ell_G}_\bmtheta\big)^{-1}(\textstyle \pn) \leq  L^{\nu_T,\ell}(\bmtheta) \leq \big(\mathcal{I}^{\nu_0,\ell_G}_\bmtheta\big)^{-1}(\textstyle \pn) + \epsilon\,.
        \end{equation*}
    \end{restatable}

    \noindent If we reformulate the above bound for interpolators under the standard loss,
    \begin{equation}
    \text{if } \hat{L}^{\ell}(D_T, \bmtheta) \leq \epsilon, \text{ then } \big(\mathcal{I}^{\nu_T,\ell}_\bmtheta\big)^{-1}(\textstyle \pn)\leq L^{\nu_T,\ell}(\bmtheta) \leq \big(\mathcal{I}^{\nu_T,\ell}_\bmtheta\big)^{-1}(\textstyle \pn) + \epsilon,
    \end{equation}
    and acknowledge that Theorem \ref{thm:da} also states that  \((\mathcal{I}^{\nu_T,\ell}_\bmtheta)^{-1}(s)\geq (\mathcal{I}^{\nu_0,\ell_G}_\bmtheta)^{-1}(s) \ \forall s>0\), it becomes evident why an interpolator leveraging the augmented loss exhibits lower generalization error and thus superior generalization capability. According to this result, the process of learning by minimizing $\hat{L}^{\ell_G}(D_0, \bmtheta)$ is advantageous over minimizing $\hat{L}^{\ell}(D_T, \bmtheta)$. As Theorem \ref{thm:da} states, the augmented loss results in a more concentrated empirical loss and a reduced inverse rate function. Consequently, as posited by Corollary \ref{cor:corda}, the generalization error associated with an interpolator is also diminished.
    
    However, as previously highlighted, in the practice of data augmentation (DA), we do not minimize $\hat{L}^{\ell_G}(D_0, \bmtheta)$ due to the lack of access to the unaltered dataset $D_0$; instead, we focus on minimizing $\hat{L}^{\ell_G}(D_T, \bmtheta)$, which refers to the augmented loss calculated over the observed transformed dataset. If the set of transformations $G$ forms a group and its associated probability distribution \(h\) is uniform ---two prevalent assumptions in DA--- it follows from the next result that minimizing $\hat{L}^{\ell_G}(D_T, \bmtheta)$ effectively amounts to minimizing $\hat{L}^{\ell_G}(D_0, \bmtheta)$.

    \begin{restatable}{prop}{group}\label{prop:group}
    Under Assumption \ref{assump:transformedinput}, if the transformations $G$ define a group and its probability distribution \(h\) is uniform, then 
    \begin{equation}\label{eq:group}
    \hat{L}^{\ell_{G}}(D_T, \bmtheta)=\hat{L}^{\ell_{G}}(D_0, \bmtheta)\,,
    \end{equation}
    \noindent where $D_0$ is any un-transformed dataset and $D_T$ is the corresponding transformed dataset obtained by applying a transformation $g\sim h$ to each of the samples in $D_0$. 
    \end{restatable}

    When multiple transformations \(g_1, \dots, g_T\) are considered, the analysis can be seamlessly expanded into a sequence of equalities, provided that each corresponding set of transformations \(G_1, \dots, G_T\) forms a group and is governed by uniform distributions \(h_1, \dots, h_T\). From this angle, it becomes feasible to combine this approach with the insights from the previous section on invariant architectures, as these two methods could address different sets of chained transformations. For instance, convolutional neural networks naturally handle translations, whereas data augmentation can accommodate other types of transformations.
    
    In summary, Theorem \ref{thm:da} shows that $\hat{L}^{\ell_G}(D_T, \bmtheta)$ for a dataset $D_T\sim\nu^n_T$ is more concentrated than $\hat{L}(D_T, \bmtheta)$ for a dataset $D\sim\nu^n_T$. Furthermore, the Chernoff bound presented in Corollary \ref{cor:corda} indicates that interpolators operating with the augmented loss exhibit a reduced generalization error. To the best of our knowledge, this is the strongest result of this type explaining why interpolators under data augmentation (DA) have better generalization error, a result achieved through the application of distribution-dependent bounds.
 
    \begin{HighlightBox}
    Theorem \ref{thm:da} shows that  the augmented loss makes the empirical loss  more concentrated. Then, the PAC-Chernoff bound of Corollary \ref{cor:corda}, explains why interpolators under the augmented loss have a smaller generalization error.
    \end{HighlightBox}
        
    However, some transformations in data augmentation (DA) fail to form a group. For instance, rotations within the range of -20 to 20 degrees do not constitute a group because combining a rotation of 20 degrees with another of 10 degrees results in a 30-degree rotation, which exceeds the range. To establish a group of rotations, it is necessary to consider the full spectrum of rotations between -360 and 360 degrees, a range not typically employed in practice. Similarly, random cropping operations also do not create a group. In such scenarios, the analysis can still proceed under the assumption that \(\hat{L}^{\ell_{G}}(D_T, \bmtheta)\approx\hat{L}^{\ell_{G}}(D_0, \bmtheta)\). Therefore, by minimizing $\hat{L}^{\ell_{G}}(D_T, \bmtheta)$, we are, in effect, also indirectly minimizing the loss $\hat{L}^{\ell_{G}}(D_0, \bmtheta)$, which is more concentrated and has interpolators with a reduced generalization error.

    The subsequent result presents an alternative approach for handling transformations that do not inherently form a group. In such scenarios, it is necessary that for every transformation in the set \(G\), an inverse transformation exists, denoted as \(\forall g \in G\), \(\exists g^{-1} \in G\). An example fulfilling this criterion is the set of rotations within the range of -20 to 20 degrees, where for any given rotation \(a\) in \([-20,20]\), its inverse rotation \(-a\) is also within the \([-20,20]\) interval. Under these conditions, if \(\bmtheta\) acts as a \textit{perfect interpolator} with respect to \(\hat{L}^{\ell_G}(D_T, \bmtheta)\) ---the actual empirical loss targeted by the learning algorithm--- then, \(\bmtheta\) similarly qualifies as a \textit{perfect interpolator} for \(\hat{L}^{\ell}(D_0, \bmtheta)\). Here, the term \textit{perfect interpolator} is used to describe models that achieve the minimal possible empirical loss.
    
    \begin{restatable}{prop}{interpolatorsda}\label{prop:interpolatorsda}
    Given an un-transformed dataset $D_0$, for any transformed dataset $D_T$ derived from $D_0$ under Assumption \ref{assump:transformedinput}, if  \( m_\bmtheta:=\essinf_{(\bmx, \bmy) \sim \nu_T} \ell(\bmy, g(\bmx),\bmtheta) \quad \forall g \in G\), and \(\forall g \in G\), \(\exists g^{-1} \in G\). Then, it verifies that
        \begin{equation*}
        \forall \bmtheta\in\bmTheta \quad \hat{L}^{\ell_G}(D_T, \bmtheta)= m_{\bmtheta} \iff \hat{L}^\ell(D_0, \bmtheta)= m_{\bmtheta}\,. 
        \end{equation*}    
    \end{restatable}

    This result implies that if a model achieves perfect interpolation on the transformed dataset \(D_T\) using the augmented loss, it concurrently achieves perfect interpolation on the original, untransformed dataset \(D_0\) under the standard loss. This means that when the learning algorithm identifies an interpolator utilizing the augmented loss (as typically occurs in practice), it is essentially also identifying an interpolator for the untransformed dataset \(D_0\). This scenario parallels the one encountered with invariant architectures, where the algorithm retrieves interpolators for the untransformed dataset. As highlighted at the outset of Section \ref{sec:invariances}, interpolators for the untransformed dataset \(D_0\) are associated with a lower generalization error compared to those for the transformed dataset \(D_T\).
    
    Figure \ref{fig:da} (right) illustrates the rate function of an inception model trained with data augmentation achieving almost perfect interpolation, where \(\hat{L}^{\ell_G}(D_T, \bmtheta)\approx m_{\bmtheta}\). It's particularly noteworthy to observe how this model's various rate functions are nearly indistinguishable, indicating that the model has nearly become invariant to rotations too. According to Proposition \ref{prop:invariance}, if a model is invariant to rotations, then \({\cal I}^{\nu_0,\ell}(a) = {\cal I}^{\nu_1,\ell}(a)\), which aligns with the outcome of this experiment.
   
    Within this theoretical framework, distinct from the methodologies referenced earlier, we can establish a connection between invariant architectures, data augmentation, and the discussions on explicit regularization presented in Section \ref{sec:explicit}. Specifically, we can explain why these regularization techniques, in conjunction with invariant architectures or data augmentation, help to find interpolators with smaller generalization errors, a relationship already empirically showcased in Figure \ref{fig:1}. This is achieved by extending the analysis from Section \ref{sec:explicit} to the new inverse rate functions $\left({\cal I}^{\nu_0}_\bmtheta\right)^{-1}(\pn)$ and $\big(\mathcal{I}^{\nu_0,\ell_G}_\bmtheta\big)^{-1}(\textstyle \pn)$, associated respectively with invariant architectures and data augmentation. For instance, implementing Proposition \ref{prop:explicitlipschitz} on these inverse rate functions elucidates the benefit of $\ell_2$-norm or distance-from-initialization regularization in the context of invariant architectures and/or data augmentation. Similarly, Proposition \ref{prop:inputgradientIinv} clarifies the advantage of input-gradient normalization in these settings.

\section{Over-Parameterization and Smooth Interpolation}\label{sec:Over-parameterization}  
    
     Many theoretical contributions have established that over-parameterization is directly related to the strong performance of interpolators \citep{neyshabur2018towards,bubeck2021law}. Specifically, the celebrated study by \citep{bubeck2023universal}\footnote{Outstanding Paper Award at NeurIPS'21.}, building on an isoperimetry assumption, demonstrates that any model capable of interpolating training data beneath the noise threshold requires a (Euclidean) Lipschitz constant of the order at least $\sqrt{nd/p}$, where $d$ represents the data's ambient dimension. We sketch this result using our notation: We say a model $\bmtheta$ is interpolating below the noise threshold if $\Lhat\leq L^\star := \min_{\bmtheta\in \bmTheta} \L$. Similarly, we denote $Lip(\bmtheta)$ the Lipschitz constant of the model $\bmtheta$, as defined in Equation~\eqref{eq:Lip}. \citetalias{bubeck2023universal} (Theorem 1) can be informally stated as: under an isoperimetry assumption, for any $\epsilon\in (0,L^\star)$ and any $\delta \in (0,1)$, with high probability $1-\delta$ over $D\sim\nu^n$, for all $\bmtheta\in\bmTheta$, 
    \begin{equation}\label{eq:overparam:bubeck}
     \text{if} \quad \Lhat\leq \epsilon \quad \text{then}\quad  
            Lip(\bmtheta)\geq \Omega\Big((L^\star - \epsilon) \sqrt{\tfrac{nd}{p}} \Big)\,.
    \end{equation}

    This implies that to keep models with $O(1)$-Lipschitz constants that continue to interpolate the training data as the number of samples $n$ grows, the number of parameters $p$ must also increase. \citetalias{bubeck2023universal} employs the Lipschitz constant as a measure of a model's \textit{smoothness}. According to their findings, over-parameterization emerges as a crucial prerequisite for interpolators to exhibit small Lipschitz constants. Given that a lower Lipschitz constant enhances generalization, over-parameterization becomes a necessary condition for achieving interpolators with reduced generalization error.
    
    In this section, we show how using PAC-Chernoff bounds we can arrive to a more nuanced understanding of the role that over-parameterization plays in generalization error of interpolators. In fact, the following result shows that over-parameterization is a  requirement that naturally emerges from the PAC-Chernoff bound of Theorem \ref{thm:LDTinv}. The following result derives a lower bound over the number of parameters of a model class based on the rate function of a model interpolating the data below the noise-threshold.
    
    \begin{restatable}{thm}{overinvariance}\label{thm:overinvariance}
        For any $\epsilon\in(0, L^\star)$ and any $\delta \in (0,1)$,  with high probability $1-\delta$ over $D\sim\nu^n$, for all $\bmtheta\in\bmTheta$, simultaneously, 
        \begin{equation*}\label{eq:overparam}
         \text{if} \quad \Lhat\leq \epsilon \quad \text{then}\quad  
            p \geq \frac{n\I{L^\star-\epsilon} + \ln\delta}{\ln k}\,.
        \end{equation*}
    \end{restatable}
    
    The above bound connects the smoothness of an interpolator, measured by the rate function as discussed in Section \ref{sec:smoothness}, and the minimum number of parameters in the model class. As the size of the training dataset increases, the number of parameters must also increase linearly if we wish to maintain the same degree of \textit{smoothness} in the model. This result generalizes the result of \citetalias{bubeck2023universal} and \textit{it does not require a isoperimetry assumption}. 
    \begin{HighlightBoxFit}
      Over-parameterization is a necessary condition for smooth interpolators.
    \end{HighlightBoxFit}
    
    The primary advantage of the aforementioned findings are their ability to establish a link between over-parameterization, interpolation, and our prior analyses about model's smoothness and its rate function. This enables us to show how \citetalias{bubeck2023universal}'s analysis of \textit{smooth interpolation}, using isoperimetry and Lipschitz continuity, represents merely one of several approximations to the characterization presented in Theorem \ref{thm:overinvariance}. As described Section \ref{sec:explicit}, the norm of parameters, the distance from initialization, the input-gradient norm, and the Lipschitz constant of a model, serve as proxies of the (inverse) rate function of a model and, consequently, of its smoothness. We first proceed to show how \citetalias{bubeck2023universal}'s characterization of \textit{smooth interpolation} can be directly deduced from Theorem \ref{thm:overinvariance} under the same isoperimetry assumption. This can also be derived under log-concave assumptions using the same techniques used in Proposition \ref{prop:inputgradientIinv} and Equation~\eqref{eq:Lip}.
    
    \begin{restatable}{cor}{overinvariancelipchitz}\label{cor:overinvariancelipchitz}
        If $\ell(\bmy,\bmx,\bmtheta)$ is Lipschitz w.r.t. $\bmx$ with constant $Lip(\bmtheta)$ and satisifies a $c$-isoperimetry assumption, then for any $\epsilon\in(0, L^\star)$ and any $\delta \in (0,1)$,  with high probability $1-\delta$ over $D\sim\nu^n$, for all $\bmtheta\in\bmTheta$, simultaneously, 
        \begin{equation*}\label{eq:overparamlip}
         \text{if} \quad \Lhat\leq \epsilon \quad \text{then}\quad  
            Lip(\bmtheta)\geq \sqrt{\tfrac{nd} {2c(p\ln k - \ln \delta)}}(L^\star - \epsilon) \,.
        \end{equation*}
    \end{restatable}
    \noindent We can similarly derive a bound with a giving norm or distance-from-initialization. 
    \begin{restatable}{cor}{overinvariancenorm}\label{cor:overinvariancenorm}
       If the loss function \(\ell(\bmy, \bmx, \bmtheta)\) is Lipschitz w.r.t. \(\bmtheta\) with constant \(M > 0\), then, for any \(\bmtheta_0 \in \bmTheta_0 = \{\bmtheta \in \bmTheta \ | \ \V_{\nu}(\ell(\bmy, \bmx, \bmtheta)) = 0\} \subset \bmTheta\), any $\epsilon\in(0, L^\star)$ and any $\delta \in (0,1)$,  with high probability $1-\delta$ over $D\sim\nu^n$, for all $\bmtheta\in\bmTheta$, simultaneously, 
        \begin{equation*}\label{eq:overparamnorm}
         \text{if} \quad \Lhat\leq \epsilon \quad \text{then}\quad  
            \|\bmtheta-\bmtheta_0\|_2\geq \sqrt{\tfrac{n}{8M(p\ln k - \ln \delta)}} (L^\star - \epsilon)\,.
        \end{equation*}
    \end{restatable}

    Corollary \ref{cor:overinvariancelipchitz} and \ref{cor:overinvariancenorm} imply that, as the number of samples $n$ increases, to have \textit{smooth} interpolators with a small Lipschitz constant, or a small parameter norm, or a small distance from initialization, we need to increase the number of parameters of the model class too. In any case, all these results are approximations of the general result given in Theorem \ref{thm:overinvariance}.  Note that a similar result could be derived in the context of input-gradient norms using Proposition \ref{prop:inputgradientIinv}.

   \begin{HighlightBox}
      Interpolating with a small parameter norm, or a small distance from initialization, or a small input-gradient norm requires over-parameterization.
    \end{HighlightBox}

    Remarkably, we can also link Theorem \ref{thm:overinvariance} with invariant architectures and data-augmentation given in Section \ref{sec:invariances}. For example, Theorem \ref{thm:da}, which says that the use of the data-augmented loss induced a higher rate function, in combination with Theorem \ref{thm:overinvariance}, would explain why interpolating under data-augmentation will require, at some point, a higher number of parameters. Moreover, the same could be said about interpolating using more invariant architectures, as they define models with higher rate functions. 
    \begin{HighlightBox}
      Interpolating using data-augmentation and/or a model-invariant architecture requires over-parameterization.
    \end{HighlightBox}
To the best of our knowledge, we are not aware of any other results showing this kind of relationships between over-parameterization and smooth interpolators.

\section{Related Work}\label{sec:relatedwork}

   In the context of uniform convergence bounds, many efforts have been made towards obtaining tighter generalization bounds  \citep{kawaguchi2017generalization, bartlett2017spectrally, neyshabur2017exploring, golowich2018size, liang2019fisher}, by, for example, only considering the  models effectively visited by the optimization algorithm. However, it is clear that these bounds are not actually meaningful (see \cite{nagarajan2019uniform} for further references and discussions). The PAC-Bayes framework has also adapted to this new paradigm by exploiting properties of the models induced by the training data set  (such as low spectral norm \citep{neyshabur2017pac}, noise stability \citep{arora2018stronger}, de-randomization \citep{negrea2020defense}, and compression \citep{arora2018stronger, zhou2018non}), but none of these bounds are shown to be empirically tight in the over-parameterized model classes and, for example, are unable to describe the interplay between invariant architectures and over-parameterization. In fact, very recently, \cite{gastpar2023fantastic} proved that, for over-parameterized model classes, none of these bounds are tight for all data-generating distributions. 
    
   Our work mainly differs from these related works in the sense that we use \textit{oracle complexity measures}, that is, PAC-Chernoff bounds and the rate function assumes access to the data-generating distribution. Consequently, we are able to tightly bound the generalization error of an interpolator, as elaborated in Section \ref{sec:smoothness}. This capability stems from our ability to derive robust results such as Theorem \ref{thm:smallerL} even under unbounded losses. Furthermore, Theorem \ref{thm:smallerL} and the PAC-Chernoff bound of Theorem \ref{thm:LDTinv} enables to leverage architectural invariances with respect to transformations inherent in the data-generating distribution.
    
   Many other recent theoretical studies try to understand specific learning techniques in deep learning in an isolated and independent way to simplify the problem (e.g, \citep{vidal2017mathematics, arora2015deep, gilbert2017towards, patel2016probabilistic, schoenholz2016deep, poole2016exponential, tishby2015deep, arora2018optimization, gunasekar2018implicit, hardt2016identity, soudry2018implicit, chen2020group, bubeck2023universal}). As a result of such isolated approach, they are not fully able, most of the times, to establish links among different learning techniques, as done in this work. Furthermore, relying on the insights provided by the field of geometric deep learning \citep{bronstein2021geometric}, our work provides a theoretical explanation, from a statistical inference point of view, on why such invariances induce smoother model classes and better generalization. 
    
\section{Conclusions and Limitations}\label{sec:conclusions}

    In this work, we discussed how a growing literature \citep{zhang2017understanding,nagarajan2019uniform,wang2024near,gastpar2023fantastic} is suggesting that bounds that solely depend on the training data are provably vacuous for over-parameterized model classes, and points to the need of exploring novel approaches to understand generalization in deep learning. Then, we argued in favor of distribution-dependent bounds, as bounds which directly depend on the data generating distribution, instead of the training data sample. A distribution-dependent PAC-Chernoff bound, outlined in Theorem \ref{thm:LDTinv}, is introduced, which shows how this bound is perfectly tight for any interpolator. Based on the complexity measure of this bound, a novel definition of \emph{smoothness} is proposed in Definition \ref{def:smoothness}, which is able to provide a clear answer to the previously unanswered question of which interpolators generalize more effectively.
    
    In Section~\ref{sec:doubledescent}, we showed how PAC-Chernoff bounds are able to capture the complex phenomenon of double-descent. Our bound is able to explain why is possible to have interpolators with a smaller generalization error and, at the same time, defined by a larger number of parameters.
    
    In Section \ref{sec:explicit}, we showed how the complexity measure of our PAC-Chernoff bound defines an regularizer that provides near-optimal performance for over-parameterized model classes interpolating the training data, as detailed in Theorem \ref{thm:optimialinterpolator}. During the rest of Section~\ref{sec:explicit}, we illustrated how a broad spectrum of existing regularization techniques can be shown to be proxies of this optimal regularizer. In certain cases, we even identified how some variations of existing regularizers can act as optimal surrogates for an interpolator's generalization error. This analysis serves to better understand the limitations of existing regularization techniques such as parameter norm, distance from initialization or input-gradient norm. 

    In Section \ref{sec:invariances}, we described how having transformed inputs, as depicted in Assumption \ref{assump:transformedinput}, make learning harder because interpolators have a higher generalization error, as shown by the PAC-Chernoff bound of Equation \ref{eq:chernoffbound:transformeddata}. The same PAC-Chernoff bound, in combination with other results, was also able to show why invariant architectures and data augmentation induce interpolator with smaller generalization error. We finished this section by discussing how this theoretical framework can establish a connection between invariant architectures, data augmentation, and the regularizers analyzed in  Section \ref{sec:explicit}.
    
    Finally, in Section \ref{sec:Over-parameterization}, we established how over-parameterization is an essential requisite for achieving smooth interpolators, with smoothness being definable across various dimensions, including parameter-norm, Lipschitz constant, or the employment of invariant architectures or data augmentation. Highlighting how this theoretical framework is able to establish connections among different, previously thought, unrelated approaches. We concluded the discussion by addressing the compatibility of larger model classes with smooth interpolators that exhibit superior generalization. 
    
    In summary, the primary insight from this study is that, distribution-dependent PAC-Chernoff bounds are a powerful tool to better understand the generalization capabilities of (over-parameterized) interpolators.

    \subsection{Discussion of Limitations}

    A limitation of this research is its assumption of a finite model class. We believe this limitation can be overcome by employing recently introduced PAC-Bayes-Chernoff bounds \citep{casado2024pac}. This PAC-Bayesian bound is also a distribution-dependent bound, relying on a similar rate function. As a result, these findings could potentially be extended to infinite model classes. However, exploring such an extension is beyond the scope of this paper. Finally, this study has not delved into the significant role that stochastic gradient descent (SGD) plays in identifying interpolators with minimal generalization error.

    Another limitation of our work is the lack of explicit connections between the smoothness definition introduced in this paper and other existing definitions in the literature. For instance, smoothness is often characterized using the largest eigenvalue of the Hessian matrix \citep{nesterov2003introductory}, the change in loss value within a perturbation radius \citep{keskar2017large}, or Lipschitz continuity of the loss function \citep{shalev2014understanding}. While our notion of smoothness is tied directly to the PAC-Chernoff framework and its distribution-dependent complexity term, future work could explore how this concept aligns or diverges from these established definitions. Establishing such connections could provide additional insights into how our framework complements or generalizes existing theoretical constructs. We have chosen to focus here on the implications of our specific definition for generalization, leaving these broader comparisons for future research. This limitation underscores the potential for follow-up studies to bridge our smoothness definition with the broader landscape of smoothness concepts in machine learning theory.

\section*{Acknowledgements} 

AM acknowledges funding for cloud computing from Google Cloud for Researchers program, from Grant PID2022-139293NB-C31 funded by MCIN/AEI/10.13039/501100011033 and by ERDF, a way of making Europe. LO acknowledges financial support from PID2022-139856NB-I00 funded by MCIN/ AEI / 10.13039/501100011033 / FEDER, UE and from the Autonomous Community of Madrid (ELLIS Unit Madrid).

\appendix

\section{Experimental Settings}\label{app:discussion:assumption}
In this appendix we expand in the details of the conducted experiments. More precisely, the architectures used and the hyper-parameters and learning methods of each figure.

\subsection{Learning Settings for Figures}

     A GitHub Repository with the conducted experiments can be found in \url{https://github.com/Ludvins/2024_PAC-Chernoff-Bound}. Regarding the experimental setting of this work, we have mainly used a small InceptionV3~\citep{szegedy2016rethinking} used in \cite{zhang2017understanding}, and Cifar10 dataset~\citep{krizhevsky2009learning}. Before showing the specific specific architecture of our small Inception, taken from \cite{zhang2017understanding}, we need to detail some of the modules that compose it:
    \begin{enumerate}
        \item Convolutional module: Convolutional layer, batch-normalization and ReLU activation.
        \item Inception module with output channels \(o_{1\times 1}\) and \(o_{3\times 3}\): Consists on 2 different convolutional layers, one with kernel \(1 \times 1\) and \(o_{1\times 1}\) output channels and another with kernel \(3 \times 3\) and \(o_{3\times 3}\) output channels. The output of this layers is then concatenated, so the total number of output channels is \(o_{1\times 1} + o_{3\times 3}\).
        \item Downsample module: Convolutional module with kernel size \(3\), stride \(2\) and padding \(0\) and MaxPooling with kernel size of \(3\) and stride \(2\). The outputs of these two layers is concatenated.
        \end{enumerate}
        With these elements, the architecture of our small InceptionV3 network is
        \begin{enumerate}
            \item Convolutional module with \(96\) output channels, kernel size \(3\), stride \(1\) and padding \(0\).
            \item Inception Module with \(o_{1 \times 1} = 32\) and \(o_{3 \times 3} = 32\).
            \item Inception Module with \(o_{1 \times 1} = 32\) and \(o_{3 \times 3} = 48\).
            \item DownSample Module with \(o_{3 \times 3} = 80\).
            \item Inception Module with \(o_{1 \times 1} = 112\) and \(o_{3 \times 3} = 48\).
            \item Inception Module with \(o_{1 \times 1} = 96\) and \(o_{3 \times 3} = 64\).
            \item Inception Module with \(o_{1 \times 1} = 80\) and \(o_{3 \times 3} = 80\).
            \item Inception Module with \(o_{1 \times 1} = 48\) and \(o_{3 \times 3} = 96\).
            \item DownSample Module with \(o_{3 \times 3} = 96\).
            \item Inception Module with \(o_{1 \times 1} = 176\) and \(o_{3 \times 3} = 160\).
            \item Inception Module with \(o_{1 \times 1} = 176\) and \(o_{3 \times 3} = 160\).
            \item Adaptative Average Pooling layer with kernel \(7 \times 7\).
            \item Fully connected layer from \(16464\) to the number of classes (i.e, \(10\)).
        \end{enumerate}
        
        Where the total number of parameters of this model is \(1.814.106\).    
    
    \paragraph{Figure~\ref{fig:1}.} For this experiments, all Inception models where trained using SGD with momentum \(0.9\) and learning rate \(0.01\) with exponential decay of \(0.95\). All models are trained for \(30.000\) iterations of batches of size \(200\) or until the train loss is under \(0.005\). These settings are selected to ensure that the random label model converges to an interpolator. Random cropping is employed using \texttt{RandomResizeCrop} function of \texttt{torchvision} with scale \((0.8, 1.0)\) and ratio \((0.9, 1.1)\). For \(\ell_2\) regularization, the multiplicative factor is \(0.01\).
    
    \paragraph{Figure~\ref{fig:histograms}.} For this figure, Standard, L2-Crop and Initial model from Figure~\ref{fig:1} are used. Subsets of size \(n=50\) of CIFAR10's test split are used to approximate samples of the data generating distribution and build the histograms.

    \paragraph{Figure~\ref{fig:double_descent}.} For this figure, we used a generalization of LeNet5 (three convolutional layers a two fully connected with ReLu activation and average pooling), where the number of channels of the convolutional layers was parameterized by \(k\). Precisely, the first layer had \(3\) input and \(\lfloor 6k \rceil\) output channels; the second layer \(\lfloor 6k \rceil\) input and \(\lfloor 16k \rceil\) output channels; and the last layer \(\lfloor 16k \rceil\) input and \(\lfloor 120k \rceil\) output channels. The set of models are created ranging \(k\) from \(0.2\) to \(4.9\) every \(0.1\); raising models from \(7k\) parameters to models with \(1.2M\) parameters. Each of this models is then trained until the train loss is lower than \(0.01\) or until the train loss has not lowered in two epochs (this only happens in the smallest models).
    
    \paragraph{Figure~\ref{fig:over-parameterization}. } The rate function of a subset of all the models in Figure~\ref{fig:double_descent} is computed here.

    \paragraph{Figure~\ref{fig:invariance}.} The batch size is fixed to \(250\) and images are standardize (this was necessary to improve learning in the MLP model). The precise MLP has 3 hidden layers with \(512\) units, with a total of \(1.735.178\) parameters. All models are trained until the interpolation regime, that is, until the train loss is under \(0.015\), which, in the worst case where \(20.000\) iterations for the MLP. Inception models are trained using a learning rate of \(0.001\) whereas MLP models use \(0.1\), both with \(0.9\) momentum and \(0.95\) exponential decay. Regarding the data, \(D_0\) is CIFAR10's test set, \(D_1\) is the result of performing random translations of \(5\%\) and \(D_2\) considers random translations of \(5\%\) and rotations of up-to \(20\%\). Both transformations are computed using \texttt{RandomAffine} function of \texttt{torchvision}.

    \paragraph{Figure~\ref{fig:rates}.} The model's specification and training setup is the same as in Figure~\ref{fig:invariance}. Regarding the data, the random shuffling of the pixels was performed using a random permutation using \texttt{Numpy}; the dataset was fully permuted and stores as a new dataset.
    
    \paragraph{Figure~\ref{fig:da}.}  The model's specification and training setup is the same as in Figure~\ref{fig:invariance}. \(D_1\) considers random translations of \(5\%\) and rotations of up-to \(20\%\) (the same as \(D_2\) in Figure~\ref{fig:invariance}). Data-augmentation was produced using the same transformations as those that define \(D_1\).

    \subsection{Estimating the Cumulant Function \texorpdfstring{$\J$}{J} and the Rate Function \texorpdfstring{$\I{\cdot}$}{I(·)}}
    
    From the definition of the cummulant function \(\J\),
    \begin{equation*}
        \J = \ln \E_{\nu}\left[e^{\lambda (L(\bmtheta)-\ell(\bmy,\bmx,\bmtheta))}\right] = \ln \E_{\nu}\left[ p(\bmy|\bmx,\bmtheta)^\lambda\right]  - \E_{\nu}[\ln p(\bmy|\bmx, \bmtheta)^\lambda]\,,
    \end{equation*}
    it is clear that computing its true value requires access to the true data generation distribution \(\nu\). However, in real-world problems, this distribution is unknown and innacesible. 
    
    The Machine Learning community is used to approximate this kind of quantities (such as the expected loss $\L$) using separate validation datasets \(D^{val}\). In fact, due to the large amount of data available in nowadays problems, using this approach is perfectly doable, leading to
    \begin{equation*}\label{eq:cummulant_estimation}
        \J \approx \ln\left(\frac{1}{M}\sum_{(\bmx, \bmy) \in D^{val}} p(\bmy|\bmx, \bmtheta)^\lambda \right) - \frac{1}{M}\sum_{(\bmx, \bmy) \in D^{val}}\ln p(\bmy|\bmx, \bmtheta)^\lambda\,.
    \end{equation*}
    It is important to notice that the above estimator is biased  due to the first term and Jensen's Inequality. In fact,
    \begin{equation*}
    \begin{aligned}
        \E_{D^{val}}\left[\ln\left(\frac{1}{M}\sum_{(\bmx, \bmy) \in D^{val}} p(\bmy|\bmx, \bmtheta)^\lambda \right)\right] &\leq \ln\left( \E_{D^{val}}  \left[\frac{1}{M}\sum_{(\bmx, \bmy) \in D^{val}} p(\bmy|\bmx, \bmtheta)^\lambda \right]\right)\\
        &= \ln\E_{\nu}\left[ p(\bmy|\bmx,\bmtheta)^\lambda\right]\,.
    \end{aligned}
    \end{equation*}
    As a result, if the size of \(D^{val}\) is not large enough, we might end understimating the cummulant function. 
    
    In regard of computational stability, computing the estimation in Equation~\eqref{eq:cummulant_estimation} can be computationally unstable due to the use of probabilities. We encourage the use of log-probabilities and log-sum-exp operations as,
    \begin{equation*}
        \J \approx \ln\left(\sum_{(\bmx, \bmy) \in D^{val}} \exp (\lambda \ln p(\bmy|\bmx, \bmtheta)) \right) - \ln M - \frac{1}{M}\sum_{(\bmx, \bmy) \in D^{val}}\lambda \ln p(\bmy|\bmx, \bmtheta)\,.
    \end{equation*}
    From this, it is straight-forward to compute the log-probabilities of the model (for example skipping the softmax layer of a NN), multiply them by \(\lambda\) and compute the mean and log-sum-exp of these quantities.

    Once the cummulant function has being approximated, computing the rate function relies on computing the optimal value of \(\lambda\),
    \begin{equation*}
        \I{a} = \sup_{\lambda > 0} \lambda a - \J\,.
    \end{equation*}
    In this regard, trying to optimize the value of \(\lambda\) doing automatic optimization resulted in a very unstable method in our experiments. Thus, the method we are recommending and the one we have employed is using a \emph{binary search} algorithm. Fixed a range in which to optimize lambda \([\lambda_{min}, \lambda_{max}]\), a binary search algorithm has complexity \(\mathcal{O}(\log_{2}(\lambda_{max} - \lambda_{min}))\). In fact, if (due to the nature of the problem) the needed value of \(\lambda_{max}\) is too large, one might perform the binary search in \([\ln(\lambda_{min}), \ln(\lambda_{max})]\), which has the same complexity but makes it easier to consider larger values of \(\lambda\).

    It is clear that computing the rate function is more complex than computing only the cummulant function (as the former requires the latter). In fact, the next result shows that it might not be necessary to compute the rate function, as the cummulant might be enough to characterize smoothness.
    
    \begin{proposition*}\label{propJvsIextended}
        If $\forall \lambda \geq 0$, $\J\leq \Jprime$, then $\forall a\geq 0$, we have $\I{a}\geq \Iprime{a}$. 
    \end{proposition*}
    \begin{proof}
        Direct consequence of the \(\I{a}\) being the Legendre transform of \(\J\).
    \end{proof}
    
\begin{figure}
      \begin{subfigure}{.43\linewidth}
        \includegraphics[width=\linewidth]{./imgs/rates/inception_l2_crop_random.pdf}
      \end{subfigure}
      \begin{subfigure}{.55\linewidth}
          \includegraphics[width=\linewidth]{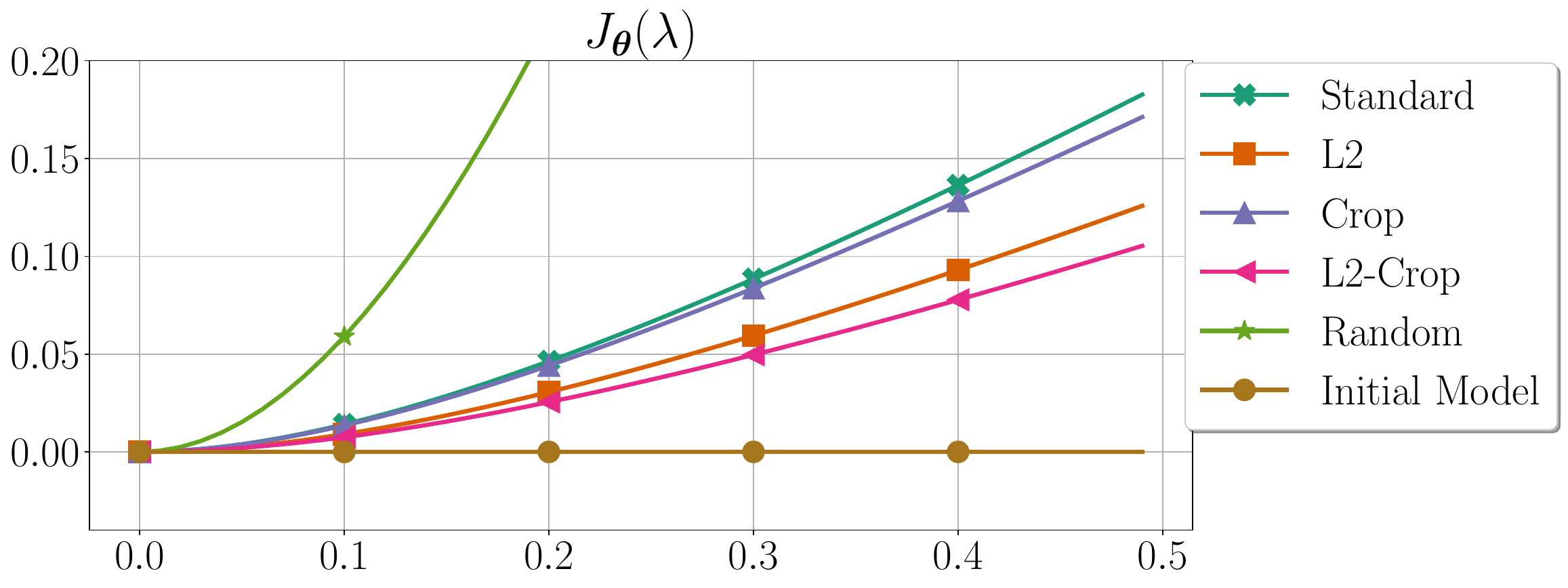}
      \end{subfigure}
        \caption{Illustration on the relationship between the rate function \(\I{a}\) (left) and the cummulant function \(\J\) (right). They display the rate and the cummulant functions for the same set of models (same experiment than Figure~\ref{fig:1}).}
        \label{fig:illustration:rateVSjensen}
    \end{figure}

    From this result, if a model \(\bmtheta\) has a higher cummulant function than another model \(\bmtheta'\), then \(\bmtheta\) is smoother than \(\bmtheta'\) and many results apply. In fact, this is the method we recommend to analyze whether a model is smoother than another. Figure \ref{fig:illustration:rateVSjensen} clearly illustrates this case. \textit{Just plotting the cummulants is enough to understand which models are smoother}. 
    
\section{Theorems and Proofs}\label{app:proofs}

\subsection{The Rate Function (go back to Section~\ref{sec:rate})}\label{app:proofs:smoothness}

    \Iwelldefined*\label{proof:Iwelldefined}
    \begin{proof}
        First, from Assumption \ref{assump:lowerbound}, we have that $m_{\bmtheta}\geq 0$, then \(\ell(\bmy,\bmx,\bmtheta) \geq 0 \ \forall (\bmx, \bmy) \in \mathcal{X}\times\mathcal{Y}\). Then, it verifies that \(\lambda(\L-\ell(\bmy,\bmx,\bmtheta)))\leq \lambda \L\). Taking exponential and expectations,
        \begin{equation*}
            \E_{\nu}\left[e^{\lambda (L(\bmtheta)-\ell(\bmy,\bmx,\bmtheta))}\right] \leq  \E_{\nu}\left[e^{\lambda L(\bmtheta)}\right]\,.
        \end{equation*}
        Lastly, the expectation on the r.h.s. is constant, leading to
        \begin{equation*}
                \J = \ln \E_{\nu}\left[e^{\lambda (L(\bmtheta)-\ell(\bmy,\bmx,\bmtheta))}\right] \leq  \ln \E_{\nu}\left[e^{\lambda L(\bmtheta)}\right] = \lambda \L\,.
        \end{equation*}
        As, by Assumption \ref{assump:lowerbound}, $\forall\bmtheta\in\bmTheta$, $\L<\infty$ and the function $\J$ is well-defined for $\lambda>0$. Next we need to show that the supremum over $\lambda$ is reached in the definition of the rate function. For that, we are showing that it is actually a maximum. First, we have that, 
        \begin{equation*}
            \frac{\partial}{\partial \lambda} (\lambda a - \J) = a - \frac{\partial}{\partial \lambda} \J\,,
        \end{equation*}
        Where the second derivative is negative as \(\frac{\partial^2}{\partial \lambda ^2} \J \geq 0\) (the cumulant is convex). As a result, when the previous derivative is zero, the maximum is reached. In fact the optimum of \(\lambda\) is a $\lambda^\star$ such that $a=\frac{\partial}{\partial \lambda} J_\bmtheta(\lambda^\star)$. Then, $\forall a \in (0, \L  - m_{\bmtheta})$, we need to show that $\exists \lambda^\star \in \mathbb{R}^+$. We have that $\frac{\partial}{\partial \lambda} J_\bmtheta(\lambda)$ is a continuous function, as it is combination of continuous functions:
        \begin{equation*}
            \frac{\partial}{\partial \lambda} \J = \frac{\E_{\nu}[p(\bmy|\bmx, \bmtheta)^\lambda \ln p(\bmy | \bmx, \bmtheta)]}{\E_{\nu}[p(\bmy|\bmx, \bmtheta)^\lambda]} - \E_{\nu}[\ln p(\bmy|\bmx, \bmtheta)]\,.
        \end{equation*}
        By standard properties of the cumulant generating function, we have that $\frac{\partial}{\partial \lambda} J_\bmtheta(0) = 0$. On the other had, also by standard properties of the cummulant \cite[Lemma 1]{herdegen2008theorem}, we have that 
        \begin{equation}\label{eq:limitGradientJ}
        \lim_{\lambda\rightarrow \infty } \frac{\partial}{\partial \lambda}\J = \esssup_{(\bmx, \bmy)}\  \L - \ell(\bmy,\bmx,\bmtheta) = \L - m_\bmtheta\,.
        \end{equation}
        Thus, if  $\frac{\partial}{\partial \lambda} J_\bmtheta(\lambda)$ is continuous, $\frac{\partial}{\partial \lambda} J_\bmtheta(0) = 0$ and $\lim_{\lambda\rightarrow \infty } \frac{\partial}{\partial \lambda}\J=\L - m_\bmtheta$, then $\forall a \in [0, \L  - m_{\bmtheta})$ there always exist a \(\lambda^\star \in \mathbb{R}^+ \) such that $a=\frac{\partial}{\partial \lambda} J_\bmtheta(\lambda^\star)$. Thus, for such values of \(a\) the rate function is finite and well defined.
        
        For $a\geq \L- m_\bmtheta$, we have that the supremum is reached when $\lambda\to \infty$, because $\J$ is monotonically increasing. Every $a\geq \L- m_\bmtheta$, can be written as $a=\L-b$, where $b\leq m_\bmtheta$. Then the limit when $\lambda\to \infty$ for any $a\geq \L- m_\bmtheta$ can be written as follows,
        \begin{align*}
            \lim_{\lambda\rightarrow\infty} \lambda (\L- b) - \J &= \lim_{\lambda\rightarrow\infty} -\lambda b -\ln \E_\nu\left[p(\bmy|\bmx,\bmtheta)^\lambda\right] = \lim_{\lambda\rightarrow\infty} -\ln \E_\nu\left[\left(\frac{p(\bmy|\bmx,\bmtheta)}{e^{-b}}\right)^\lambda\right]\\
            &= -\ln \E_\nu\left[\lim_{\lambda\rightarrow\infty}\left(\frac{p(\bmy|\bmx,\bmtheta)}{e^{-b}}\right)^\lambda\right] = -\ln \E_\nu\left[\1(p(\bmy|\bmx,\bmtheta)=e^{-b}\right]\\
            &= -\ln \E_\nu[\1(\ell(\bmy,\bmx,\bmtheta)=b)] = -\ln \mathbb{P}_\nu(\ell(\bmy,\bmx,\bmtheta)=b)\,.
        \end{align*}
    
        If $b< m_\bmtheta$ or, equivalently, $a>\L-m_\bmtheta$, we then have $\I{a}=\infty$. And if $a=L-m_\bmtheta$, we then have $\I{a} = -\ln \mathbb{P}_\nu(\ell(\bmy,\bmx,\bmtheta)=m_\bmtheta)$, which, may be well defined or equal to $\infty$. 
        

        We now proceed to show that inverse rate is well defined.  By definition of the inverse rate function, 
        \[\Iinv{s}=\inf_{\lambda\geq 0}\frac{s+\J}{\lambda}\leq \lim_{\lambda\rightarrow\infty}\frac{s+\J}{\lambda} =\lim_{\lambda\rightarrow\infty}\frac{\J}{\lambda} \]
        Then, by L'Hôpital's rule and Equation \eqref{eq:limitGradientJ}, we have

                \[\lim_{\lambda\rightarrow\infty}\frac{\J}{\lambda} =\lim_{\lambda\rightarrow\infty}\nabla_\lambda \J =\L - m_\bmtheta\]
                
        From Assumption \ref{assump:lowerbound} we have that $\L<\infty$ and $m_\bmtheta\geq 0$, then we deduce that $\forall s\geq 0$ we have that $\Iinv{s}<\infty$. 
    \end{proof}
    
\ifjmlr
\else
    \Rateproperties*\label{proof:Rateproperties}
    \begin{proof}
        Many of these properties are immediate from the fact that \(J\) is a cummulant function and \(\I{a}\) its Legendre transformation. However, we instantiate the proofs on our settings.
        \begin{enumerate}[label=(\roman*)]
            \item The Legendre transform of a convex function is also convex; thus, \(\I{a}\) is convex and \(\Iinv{s}\) is concave.
            \item By definition of the rate function, \(\I{a} = \sup_{\lambda > 0} \lambda a -\J = \lambda^\star_a a - J_{\bmtheta}(\lambda^\star_a)\); where \(\lambda^\star_a\) is the optimal value of \(\lambda\) for each given \(a\). It is clear that, evaluating on \(a'\) increases the value as \(a'\geq a\),
                \begin{equation*}
                    \lambda^\star_a a' - J_{\bmtheta}(\lambda^\star_a) \geq \lambda^\star_a a - J_{\bmtheta}(\lambda^\star_a)\,.
                \end{equation*}
                In fact, it is also clear that the supremum is higher or equal than the l.h.s., that is,
                \begin{equation*}
                    \I{a'} = \sup_{\lambda > 0} \lambda a' -\J  \geq \lambda^\star_a a' - J_{\bmtheta}(\lambda^\star_a)\,.
                \end{equation*}
                As a result, \(\I{a'} \geq  \I{a}\). By definition of the inverse rate function,
                \begin{equation*}
                    \Iinv{s'} = \inf_{\lambda > 0} \frac{s' + \J}{\lambda} = \frac{s' + J_{\bmtheta}(\lambda^\star_{s'})}{\lambda^\star_{s'}}\,,
                \end{equation*}
                where \(\lambda^\star_s\) is the optimal value of \(\lambda\) for each given \(s\). It is clear that, evaluating on \(s\) decreases the value as \(s'\geq s\),
                \begin{equation*}
                    \frac{s' + J_{\bmtheta}(\lambda^\star_{s'})}{\lambda^\star_{s'}} \geq \frac{s + J_{\bmtheta}(\lambda^\star_{s'})}{\lambda^\star_{s'}}\,.
                \end{equation*}
                In fact, it is also clear that the infimum is lower or equal than the r.h.s., that is,
                \begin{equation*}
                    \frac{s + J_{\bmtheta}(\lambda^\star_{s'})}{\lambda^\star_{s'}} \geq \inf_\lambda \frac{s + J_{\bmtheta}(\lambda)}{\lambda} = \Iinv{s}\,.
                \end{equation*}
                As a result, \(\Iinv{s'} \geq  \Iinv{s}\).
            \item For the rate function, the derivative is \(\frac{\partial}{\partial a} \I{a} = \lambda^\star_a\); 
            where \(\lambda^\star_a\) is the optimal value for each \(a\) of the rate function. From the fact that the cummulant function is positive, it is clear that the optimal \(\lambda\) tends to zero as \(a\) goes to zero. For the inverse rate function, this is direct consequence of being the inverse of the rate function.
            \item 
            \item This is clear because these functions depend on \(\bmtheta\) though the loss function.
        \end{enumerate}
    \end{proof}
    
    \LDT*\label{proof:LDT}
    \begin{proof}
        Cramér-Chernoff's bound states that for any random variable \(X\), it verifies that \(P(X \geq s) \leq \inf_{t > 0} \mathbb{E}[e^{t X}]e^{-t s}\). Applying this result to the random variable over possible datasets \( \hat{L}(D,\bmtheta) - L(\bmtheta)\), for a fixed $\bmtheta\in\bmTheta$, leads to 
        \begin{equation*}
             P( L(\bmtheta) - \hat{L}(D,\bmtheta) \geq s) \leq \inf_{t > 0} \mathbb{E}\left[e^{t ( L(\bmtheta) - \hat{L}(D,\bmtheta) ) }\right]e^{- t s}\,.
        \end{equation*}
        The expectation in the r.h.s. can be transformed as
        \begin{equation*}
            \begin{aligned}
             \mathbb{E}\left[e^{t (  L(\bmtheta) - \hat{L}(D,\bmtheta) )}\right] &= \mathbb{E}\left[e^{t (  \tfrac{1}{n}\ln p(D|\bmtheta)- \E_\nu[\ln p(\bmy| \bmx, \bmtheta)]}\right] = \mathbb{E}\left[e^{\tfrac{t}{n}\ln p(D|\bmtheta)}\right] e^{-t\mathbb{E}_\nu[\ln p(\bmy| \bmx, \bmtheta)]}\,.
             \end{aligned}
        \end{equation*}
        Moreover, the first expectation in this last term is
        \begin{equation*}
            \begin{aligned}
            \mathbb{E}\left[e^{\tfrac{t}{n}\ln p(D|\bmtheta)}\right] &= \E_{\nu^n}\left[p(D|\bmtheta)^{\tfrac{t}{n}}\right] = \E_{\nu}\left[P(\bmy|\bmx,\bmtheta)^{\tfrac{t}{n}}\right]^n =\mathbb{E}\left[e^{\tfrac{t}{n}\ln p(\bmy|\bmx,\bmtheta)}\right]^n\,.
            \end{aligned}
        \end{equation*}
        Using this, and parameterizing \(t\) as \(\lambda n\), with \(\lambda > 0\),
        \begin{equation*}
        P(L(\bmtheta) - \hat{L}(D,\bmtheta) \geq s) \leq \inf_{\lambda > 0} \mathbb{E}\left[e^{\lambda (  L(\bmtheta) - \ell(\bmy, \bmx, \bmtheta)) }\right]^n e^{- \lambda n s}\,.
        \end{equation*}
        Taking exponential and logarithm on the r.h.s, and using the definition of the smoothness function \(\J\) and the rate function \(\I{a}\):
        \begin{equation*}
            \begin{aligned}
                P(L(\bmtheta) - \hat{L}(D,\bmtheta) \geq s) &\leq \inf_{\lambda > 0} e^{ n \ln \mathbb{E}\left[e^{\lambda (L(\bmtheta)  - \ell(\bmy, \bmx, \bmtheta))}\right] - \lambda  ns} \leq \inf_{\lambda > 0} e^{n\J  - \lambda n s} = e^{-n \I{s}}\,.
            \end{aligned}
        \end{equation*}
    \end{proof}

    \crammer*\label{proof:crammer}
    \begin{proof}
    Direct application of Cramér's Theorem \citep{cramer1938nouveau,ellis2006entropy}, over the random variable $X=\L - \Lhat$, for a fixed $\bmtheta$, where the randomness comes from $D\sim\nu^n$. Similar to the application of Chernoff's bound in Theorem~\ref{thm:LDT}. 
    \end{proof}
\fi    
    
    \tightchernoff*\label{proof:tightchernoff}
    \begin{proof}
        It is clear from the definition of \(m_\bmtheta := \essinf_{(\bmy, \bmx) \sim \nu} \ell(\bmy, \bmx, \bmtheta)\) and the rate function \(\I{a}\), that the limit at both sides is \(0\).
    \end{proof}

\subsection{The Rate Function Characterizes the Generalization of Interpolators (go back to Section~\ref{sec:smoothness})}

    \LDTinv*\label{proof:LDTinv}
    \begin{proof}
        By Chernoff's Theorem \ref{thm:LDT}, for a given $\bmtheta$, we have, \(\P\Big(L(\bmtheta) - \hat{L}(D,\bmtheta) \geq a\Big)\leq e^{-n \I{a}}\). Naming \(\delta' = e^{-n \I{a}}\) and re-arranging terms, \(a = \Iinv{-\frac{1}{n}\ln \delta'}\). This allows us to rewrite the first equation as
        \begin{equation*}
            \P\Big(L(\bmtheta) - \hat{L}(D,\bmtheta) \geq \Iinv{\tfrac{1}{n}\ln \tfrac{1}{\delta'}} \Big)\leq \delta'\,.
        \end{equation*}
        
        Reversing and using an union bound over the set of models, 
        \begin{equation*}
            \P\Big( \bigcup_{\bmtheta\in\bmTheta} L(\bmtheta) - \hat{L}(D,\bmtheta) \geq \Iinv{\tfrac{1}{n}\ln \tfrac{1}{\delta'}} \Big)\leq \sum_{\bmtheta\in\bmTheta} \P\Big(  L(\bmtheta) - \hat{L}(D,\bmtheta) \geq \Iinv{\tfrac{1}{n}\ln \tfrac{1}{\delta'}} \Big)\,.
        \end{equation*}
        As we have $k^p$ different models, the r.h.s. can be rewritten as
        \begin{equation*}
            \P\Big( \bigcup_{\bmtheta\in\bmTheta} L(\bmtheta) - \hat{L}(D,\bmtheta) \geq \Iinv{\tfrac{1}{n}\ln \tfrac{1}{\delta'}} \Big)\leq k^p \delta' \,.
        \end{equation*}
        By reparameterizing the above inequality with $\delta'=\delta k^{-p}$ we have
        \begin{equation*}
            \P\Big( \bigcup_{\bmtheta\in\bmTheta} L(\bmtheta) - \hat{L}(D,\bmtheta) \geq \Iinv{\pn}  \Big)\leq  \delta \,.
        \end{equation*}
        Which verifies,
        \begin{equation*}
            1-\P\Big( \bigcup_{\bmtheta\in\bmTheta} L(\bmtheta) - \hat{L}(D,\bmtheta) \geq \Iinv{\pn} \Big)\geq 1-\delta\,.
        \end{equation*}
        Which is equivalent to, 
        \begin{equation*}
            \P\Big( \bigcap_{\bmtheta\in\bmTheta} L(\bmtheta) - \hat{L}(D,\bmtheta) \leq \Iinv{\pn} \Big)\geq 1-\delta \,.
        \end{equation*}
    \end{proof}

    \corLDTinv*\label{proof:corLDTinv}
    \begin{proof} Thet proof follows the same approach as Theorem~\ref{thm:LDTinv}, with the union bound applied to the models in \(\bar{\bmTheta}\) rather than directly to the models in \(\bmTheta\). This adjustment is valid because, when applying the union bound, we only need to consider the total number of distinct random variables; in this case, \(\Lhat\)  with \(D\sim\nu^n\). Models in \(\bmTheta\) that are not in \(\bar{\bmTheta}\) define random variables that are duplicated by definition and therefore do not need to be accounted for in the application of the union bound.
    \end{proof}
    
    \boundlimit*\label{proof:boundlimit}
    \begin{proof} Let \(\bar{Z}_n := \sqrt{n} (\L - \Lhat)\) and \(J_{\bar{Z}_n}, {\cal I}_{\bar{Z}_n}\) and \({\cal I}^{-1}_{\bar{Z}_n}\) denote its cumulant, rate and inverse rate function. Then, by properties of the cummulant, it verifies that \(J_{\bar{Z}_n}(\lambda) = nJ_{\bmtheta}\big(\tfrac{\lambda}{\sqrt{n}}\big)\). Using the definition of the rate function, we can deduce that 
        \[ {\cal I}_{\bar{Z}_n}(a) =\sup_{\lambda} \ a\lambda+n J_{\bmtheta}\big(\tfrac{\lambda}{\sqrt{n}}\big)=n\sup_{\lambda} \ \tfrac{\lambda}{\sqrt{n}}\tfrac{a}{\sqrt{n}}+ J_{\bmtheta}\big(\tfrac{\lambda}{\sqrt{n}}\big)=n\I{\tfrac{a}{\sqrt{n}}}\,.
        \]
        Using a second order Taylor expansion over $\I{a}$ around $a=0$, where \(\I{0} = 0\) and \(\frac{\partial}{\partial a}\I{0} = 0\), we got that
        \begin{equation*}
           \I{\tfrac{a}{\sqrt{n}}} = \tfrac{1}{2} \tfrac{\partial^2}{\partial a^2} \I{0} \big(\tfrac{a}{ \sqrt{n}}\big)^2 + \big(\tfrac{a}{ \sqrt{n}}\big)^2 h\big(\tfrac{a}{ \sqrt{n}}\big)\,,
        \end{equation*}
        where \(\lim_{t\to 0} h(t) = 0\). Furthermore, it verifies that \citep{ellis2006entropy}
        \[
        \tfrac{\partial^2}{\partial a^2} \I{0} = \left( \tfrac{\partial^2}{\partial \lambda^2}J_\bmtheta(0)\right)^{-1}\,.
        \]
        From there, we have that 
        \begin{align*}
    \frac{\partial^2}{\partial \lambda^2}\J&= - \frac{\partial}{\partial \lambda} \frac{\E_\nu[ e^{-\lambda \ell(\bmy,\bmx,\bmtheta)} \ell(\bmy,\bmx,\bmtheta)]}{\E_\nu[ e^{-\lambda \ell(\bmy,\bmx,\bmtheta)}]}\\
    &= \frac{\E_\nu[ e^{-\lambda \ell(\bmy,\bmx,\bmtheta)} (\ell(\bmy,\bmx,\bmtheta))^2]}{\E_\nu[ e^{-\lambda \ell(\bmy,\bmx,\bmtheta)}]} - \frac{\E_\nu[ e^{-\lambda \ell(\bmy,\bmx,\bmtheta)} \ell(\bmy,\bmx,\bmtheta)]^2}{\E_\nu[ e^{-\lambda \ell(\bmy,\bmx,\bmtheta)}]^2}.
\end{align*}
        Which evaluated at $\lambda=0$, gives 
        \[\tfrac{\partial^2}{\partial \lambda^2}J_\bmtheta(0)= \mathbb{V}_\nu( \ell(\bmy, \bmx, \bmtheta)) =: \sigma^{2}\]
        Then,
        \[\lim_{n\rightarrow\infty} {\cal I}_{\bar{Z}_n}(a) = \lim_{n\rightarrow\infty} n \I{\tfrac{a}{\sqrt{n}}} = \lim_{n\rightarrow\infty} n \frac{1}{2} \sigma^{-2} \big(\tfrac{a}{ \sqrt{n}}\big)^2 +  n \big(\tfrac{a}{ \sqrt{n}}\big)^2 h\big(\tfrac{a}{ \sqrt{n}}\big)=\frac{1}{2}  \sigma^{-2} a^2\,.\]
        Then, as both the rate and its inverse are continuous functions, it verifies that
        \[
        \lim_{n\rightarrow\infty} {\cal I}^{-1}_{\bar{Z}_n}(s) = \left(\lim_{n\rightarrow\infty} {\cal I}_{\bar{Z}_n}\right)^{-1}(s) = \sqrt{2\sigma^2s}
        \]
        By definition of the inverse rate, we got that
        \[ {\cal I}^{-1}_{\bar{Z}_n}(s) = \inf_{\lambda} \frac{s+J_{\bar{Z}_n}(\lambda)}{\lambda}=\inf_{\lambda} \frac{s+n J_{\bmtheta}(\tfrac{\lambda}{\sqrt{n}})}{\lambda}=\sqrt{n}\inf_{\lambda} \frac{\tfrac{s}{n}+n J_{\bmtheta}(\tfrac{\lambda}{\sqrt{n}})}{\tfrac{\lambda}{\sqrt{n}}}=\sqrt{n}\Iinv{\tfrac{s}{n}}\,.\]
        As a result,
        \[\lim_{n\rightarrow\infty} \sqrt{n}\Iinv{\tfrac{s}{n}} = \lim_{n\rightarrow\infty} {\cal I}^{-1}_{\bar{Z}_n}(s) = \sqrt{2\sigma^2s}\,.\]
         
    \end{proof}
    
    \boundtightness*\label{proof:boundtightness}
    \begin{proof}
        By Theorem~\ref{thm:LDTinv}, \(\L \leq \Lhat +  \Iinv{\textstyle \pn}\), and, by Proposition  \ref{prop:Iwelldefined}, $\I{a}$ is well defined $\forall a\in[0,\L-m_\bmtheta)$. Then, by Proposition  \ref{prop:Rateproperties}, $\forall b>0$, $\Iinv{b}\in [0,\L-m_\bmtheta)$. In consequence, we have, 
        \begin{equation*}
            \L \leq \Lhat +  \Iinv{\textstyle \pn}\leq \Lhat + \L -m_\bmtheta\,.
        \end{equation*}
        The result follows from $m_\bmtheta\geq0$, $\Lhat\leq \epsilon$ and rearranging terms..
    \end{proof}

    \smootherlimit*\label{proof:smootherlimit}
    \begin{proof} Let \(\bar{Z}_n := \sqrt{n} (\L - \Lhat)\) and \(J_{\bar{Z}_n}, {\cal I}_{\bar{Z}_n}\) and \({\cal I}^{-1}_{\bar{Z}_n}\) denote its cumulant, rate and inverse rate function. Then, by properties of the cummulant, it verifies that \(J_{\bar{Z}_n}(\lambda) = nJ_{\bmtheta}\big(\tfrac{\lambda}{\sqrt{n}}\big)\). Using the definition of the rate function, we can deduce that 
        \[ {\cal I}_{\bar{Z}_n}(a) =\sup_{\lambda} \ a\lambda+n J_{\bmtheta}\big(\tfrac{\lambda}{\sqrt{n}}\big)=n\sup_{\lambda} \ \tfrac{\lambda}{\sqrt{n}}\tfrac{a}{\sqrt{n}}+ J_{\bmtheta}\big(\tfrac{\lambda}{\sqrt{n}}\big)=n\I{\tfrac{a}{\sqrt{n}}}\,.
        \]
        Using a second order Taylor expansion over $\I{a}$ around $a=0$, where \(\I{0} = 0\) and \(\frac{\partial}{\partial a}\I{0} = 0\), we got that
        \begin{equation*}
           \I{\tfrac{a}{\sqrt{n}}} = \tfrac{1}{2} \tfrac{\partial^2}{\partial a^2} \I{0} \big(\tfrac{a}{ \sqrt{n}}\big)^2 + \big(\tfrac{a}{ \sqrt{n}}\big)^2 h\big(\tfrac{a}{ \sqrt{n}}\big)\,,
        \end{equation*}
        where \(\lim_{t\to 0} h(t) = 0\). Furthermore, it verifies that 
        \[
        \tfrac{\partial^2}{\partial a^2} \I{0} = \left( \tfrac{\partial^2}{\partial \lambda^2}J_\bmtheta(0)\right)^{-1}= \mathbb{V}_\nu( \ell(\bmy, \bmx, \bmtheta))^{-1} =: \sigma^{-2}\,.
        \]
        Then,
        \begin{equation}\label{eq:limitrate}
        \lim_{n\rightarrow\infty} {\cal I}_{\bar{Z}_n}(a) = \lim_{n\rightarrow\infty} n \I{\tfrac{a}{\sqrt{n}}} = \lim_{n\rightarrow\infty} n \frac{1}{2} \sigma^{-2} \big(\tfrac{a}{ \sqrt{n}}\big)^2 +  n \big(\tfrac{a}{ \sqrt{n}}\big)^2 h\big(\tfrac{a}{ \sqrt{n}}\big)=\frac{1}{2}  \sigma^{-2} a^2\,.\end{equation}

        Let us assume that \( \mathbb{V}_\nu( \ell(\bmy, \bmx, \bmtheta)) \leq \mathbb{V}_\nu( \ell(\bmy, \bmx, \bmtheta' ))\). Then, we aim to show that there exists \(\beta > 0\) such that \(\I{a} \geq \Iprime{a} \quad \forall a \leq \beta\). For a fixed value of \(a\), by the limit in Equation~\ref{eq:limitrate}, we got that for any \(\epsilon > 0\), there exists \(n_0(a), n_0'(a) > 0\) such that
        \[
        \big| n \I{\tfrac{a}{\sqrt{n}}} - \tfrac{1}{2}\mathbb{V}_\nu( \ell(\bmy, \bmx, \bmtheta))^{-2}a^2\big| < \epsilon \quad \forall n > n_0(a)\,,
        \]
        and
        \[
        \big| n \Iprime{\tfrac{a}{\sqrt{n}}} - \tfrac{1}{2}\mathbb{V}_\nu( \ell(\bmy, \bmx, \bmtheta'))^{-2}a^2\big| < \epsilon \quad \forall n > n_0'(a)\,.
        \]
        Let \(\epsilon = \mathbb{V}_\nu( \ell(\bmy, \bmx, \bmtheta))^{-2}a^2 - \mathbb{V}_\nu( \ell(\bmy, \bmx, \bmtheta'))^{-2}a^2 \geq 0 \) and \(n_0^\star(a) = \max\{n_0(a), n_0'(a)\}\). Then, 
        \[
            \I{\tfrac{a}{\sqrt{n}}}  \geq  \Iprime{\tfrac{a}{\sqrt{n}}} \quad \forall n > n_0^\star(a) \,.
        \]
        This implies that \( \I{c}  \geq  \Iprime{c}\), for any \(c \leq a/\sqrt{n_0^\star(a)}\), as there exists \(n = (a/c)^2\) verifying \(c = a/\sqrt{n}\). We may consider then \(\beta := \sup_{a> 0 } \{a/\sqrt{n_0^\star(a)}\} > 0\), which might be infinite. It verifies that
        \[
        \I{a}  \geq  \Iprime{a} \quad \forall a \leq \beta\,.
        \]
        Let us now assume that there exists \(\beta > 0\) such that \(\bmtheta\) is \(\beta\)-smoother than \(\bmtheta'\), or, equivalently \(\I{a}  \geq  \Iprime{a} \quad \forall a \leq \beta\). Then, 
        \[
        \lim_{n\rightarrow\infty} n \I{\tfrac{a}{\sqrt{n}}} \geq \lim_{n\rightarrow\infty} n \Iprime{\tfrac{a}{\sqrt{n}}} \,.
        \]
        Using the limit in Equation~\ref{eq:limitrate}, we got
        \[
        \mathbb{V}_\nu( \ell(\bmy, \bmx, \bmtheta))^{-2}a^2 \geq \mathbb{V}_\nu( \ell(\bmy, \bmx, \bmtheta'))^{-2}a^2\,,
        \]
        leading to 
        \[
        \mathbb{V}_\nu( \ell(\bmy, \bmx, \bmtheta)) \leq \mathbb{V}_\nu( \ell(\bmy, \bmx, \bmtheta'))\,.
        \]

    \end{proof}

    \smallerL*\label{proof:smallerL}
    \begin{proof}
    If $\bmtheta$ is  $\beta$-smoother than $\bmtheta'$, by Definition \ref{def:smoothness}, \(\forall a\in(0,\beta] \quad \I{a}\geq \Iprime{a}\), where $\beta = \Iinv{\pn}$. Then, we have that 
    \[
    \Iinv{\pn}\leq \Iinvprime{\pn}\,.
    \]
    As the rate function $\Iprime{a}$ is invertible and its image lies in $[0,\Lprime-m_{\bmtheta'})$, where $m_{\bmtheta'}\geq 0$, due to Assumption \ref{assump:lowerbound}, we have that $\Iinvprime{s}\in[0,\Lprime-m_{\bmtheta'})$. In consequence, we also have 
    \begin{equation*}
        \Iinvprime{\pn}\leq \Lprime\,.
    \end{equation*}
    As \(\Iinv{\pn}\leq \Iinvprime{\pn}\), it verifies \(\Iinv{\pn}\leq \Lprime\). By the PAC-Chernoff bound of Theorem \ref{thm:LDTinv} and because $\Lhat\leq \epsilon$, we have with h.p. $1-\delta$ over $D\sim\nu^n$, 
    \begin{equation*}
        \L\leq \Lhat + \Iinv{\pn}\leq \epsilon + \Iinv{\pn}\,.
    \end{equation*}
    Combining the last two inequalities, we have 
    \begin{equation*}
        \L\leq \epsilon + \Iinv{\pn} \leq \epsilon + \Lprime\,.
    \end{equation*}
    The statement of the theorem directly derives from the above inequality. 
    \end{proof}

\subsection{Explicit Regularization (go back to Section~\ref{sec:explicit})}\label{app:proofs:regularization}
    \optimialinterpolator*\label{proof:optimialinterpolator}
    \begin{proof}
    From the definitions of \(\bmtheta^{\cross}_\epsilon\) and \(\bmtheta^\star_\epsilon\), given by 
    \begin{equation*}
        \bmtheta^\star_\epsilon = \argmin_{\bmtheta\,:\,\Lhat\, \leq \, \epsilon}\,  \L\,, \quad\quad \bmtheta^{\cross}_\epsilon = \argmin_{\bmtheta\,:\,\Lhat\, \leq \, \epsilon} \, \Lhat + \Iinv{\pn}\,,
    \end{equation*}
    it is clear that \(L(\bmtheta^\star_\epsilon) \leq L(\bmtheta^{\cross}_\epsilon)\). On the other hand, because for any $s\geq 0$, $\Iinv{s}\in[0,\L-m_\bmtheta)$ (Proposition \ref{prop:Rateproperties}), we have that $\forall\bmtheta\in\bmTheta$: 
    \begin{equation*}
       \Iinv{\pn} + \Lhat \leq \L +\Lhat\,.
    \end{equation*}
    By definition of \(\bmtheta^{\cross}_\epsilon\) and \(\bmtheta^{\star}_\epsilon\), 
    \begin{equation*}
         \mathcal{I}_{\bmtheta^{\cross}_\epsilon}^{-1}(\pn) + \hat{L}(D, \bmtheta^{\cross}_\epsilon) \leq  \mathcal{I}_{\bmtheta^{\star}_\epsilon}^{-1}(\pn) + \hat{L}(D, \bmtheta^{\star}_\epsilon)\,,
    \end{equation*}
    which gives 
    \begin{equation*}
       \mathcal{I}_{\bmtheta^{\cross}_\epsilon}^{-1}(\pn) + \hat{L}(D, \bmtheta^{\cross}_\epsilon) \leq  L(\bmtheta^{\star}_\epsilon) + \hat{L}(D, \bmtheta^{\star}_\epsilon) \leq L(\bmtheta^{\star}_\epsilon) + \epsilon\,.
    \end{equation*}
    This, in combination with the PAC-Chernoff bound of Theorem \ref{thm:LDTinv} gives
    \begin{equation*}
       L(\bmtheta^{\cross}_\epsilon) \leq \mathcal{I}_{\bmtheta^{\cross}_\epsilon}^{-1}(\pn) + \hat{L}(D, \bmtheta^{\cross}_\epsilon) \leq L(\bmtheta^{\star}_\epsilon) + \epsilon\,.
    \end{equation*}
    From this, we deduce that, \(L(\bmtheta^{\cross}_\epsilon)\leq L(\bmtheta^\star_\epsilon)  +\epsilon\). Thus, we got that \(L(\bmtheta^\star_\epsilon) \leq L(\bmtheta^{\cross}_\epsilon)\) and \(
    L(\bmtheta^{\cross}_\epsilon)\leq L(\bmtheta^\star_\epsilon)  +\epsilon\), finishing the proof.
    \end{proof}
    
    \explicitlipschitz*\label{proof:explicitlipschitz}
    \begin{proof}
         If the loss is Lipschitz continuous with constant $M$, \(\forall y,x,\bmtheta\quad \|\nabla_{\bmtheta} \ell(\bmy,\bmx,\bmtheta)\|^2_2\leq M\). Then, $\J$ verifies
         \begin{equation*}
         \begin{aligned}
             \|\nabla_{\bmtheta}\J\|^2_2&=||-\lambda \E_{\nu p^\lambda}\left[\nabla_{\bmtheta}\ell(\bmy, \bmx, \bmtheta)\right] + \lambda \E_\nu\left[\nabla_{\bmtheta}\ell(\bmy, \bmx, \bmtheta) \right]||_2^2\\
            &\leq \lambda^2 \E_{\nu p^\lambda}\left[\|\nabla_{\bmtheta}\ell(\bmy, \bmx, \bmtheta)\|_2^2\right] +  \lambda^2\E_\nu\left[\|\nabla_{\bmtheta}\ell(\bmy, \bmx, \bmtheta) \|_2^2\right]\\
             &\leq 2M \lambda^2\,,
         \end{aligned}
         \end{equation*}
         where \(\E_{\nu p^\lambda}\left[\nabla_{\bmtheta}\ell(\bmy, \bmx, \bmtheta)\right] = \frac{\E_{\nu}[p(\bmy|\bmx, \bmtheta)^\lambda \ell(\bmy, \bmx, \bmtheta)]}{\E_{\nu}[p(\bmy|\bmx, \bmtheta)^\lambda]} \). With this, we have
         \begin{equation*}
             |\J - J_{\bmtheta_0}(\lambda)|\leq 2M \lambda^2\|\bmtheta-\bmtheta_0\|^2_2
             \implies \J\leq 2M \lambda^2\|\bmtheta-\bmtheta_0\|^2_2\,.
         \end{equation*}
         Then, for any \(a \geq 0\), we have \( \frac{a+\J}{\lambda}\leq \frac{a+2M \lambda^2\|\bmtheta-\bmtheta_0\|^2_2}{\lambda}\); where by definition of the inverse rate we got 
         \begin{equation*}
             \Iinv{a} \leq \frac{a+\J}{\lambda}\leq \frac{a+2M \lambda^2\|\bmtheta-\bmtheta_0\|^2_2}{\lambda}\,.
         \end{equation*}
         As the inequality holds for any \(\lambda \geq 0\), we take the one that minimizes the r.h.s, leading to \( \Iinv{a}\leq \sqrt{2M a}\|\bmtheta-\bmtheta_0\|_2\). On the other hand, $\Iinv{a}$ is Lipschitz with constant $M$ as
         \[ \| \nabla_{\bmtheta} \Iinv{a}\|^2_2 = \Big\|\frac{\nabla_{\bmtheta} J_\bmtheta(\lambda^\star_a)}{\lambda^\star_a}\Big\|^2_2\leq \frac{\|\nabla_{\bmtheta} J_\bmtheta(\lambda^\star_a)\|^2_2}{\lambda^{\star,2}_a}\leq 2 M\,.\]
        
         In consequence, \(\textstyle (\Iinv{\pn}-{\cal I}^{-1}_{\bmtheta_0}(\pn))^2 \leq 2M ||\bmtheta - \bmtheta_0||^2\). Which implies that \(\Iinv{\pn} \leq \sqrt{2M} ||\bmtheta - \bmtheta_0||_2\). Thus, it simultaneously holds that
         \[\Iinv{\pn} \leq \sqrt{2M} ||\bmtheta - \bmtheta_0||_2 \quad \text{and} \quad \Iinv{\pn}\leq \sqrt{2M \pn}\|\bmtheta-\bmtheta_0\|_2\,.\]
    \end{proof}
    
    \explicitexponential*\label{proof:explicitexponential}

    \begin{proof} Using Theorem~\ref{thm:boundlimit}, we got that \(\lim_{n\rightarrow\infty} \ \sqrt{n}\, \Iinv{\pn} = \sqrt{2\mathbb{V}_\nu( \ell(\bmy, \bmx, \bmtheta))\ln \tfrac{k^p}{\delta}}\). The proof concludes from the fact that using the exponential family,   
    \begin{equation*}
        \V_\nu\big(\ell(\bmy,\bmx,\bmtheta) \big) = \V_\nu\big(\bmtheta^T s(\bmy,\bmx) - a(\bmtheta) + k\big) = \bmtheta^T\text{Cov}_{\nu} \big(s(\bmy,\bmx)\big) \bmtheta\,.
    \end{equation*}
    \end{proof}
    
    \Iinvnorm*\label{proof:Iinvnorm}
    \begin{proof}
    A second-order Taylor expansion of $\J$ w.r.t. to $\bmtheta$ centered around $\bmtheta_0$ is
    \begin{equation*}
    J_{\bmtheta_0}(\lambda)  + \nabla_\bmtheta J_{\bmtheta_0}(\lambda)(\bmtheta-\bmtheta_0) + \frac{1}{2} (\bmtheta-\bmtheta_0)^T\nabla_{\bmtheta \bmtheta}J_{\bmtheta_0}(\lambda)(\bmtheta-\bmtheta_0)\,.
    \end{equation*}
    By standard properties of the cumulant generating function over centered random variables, we have $J_{\bmtheta_0}(\lambda)=0$. While the $\nabla_\bmtheta J_{\bmtheta}(\lambda)$ can be expressed as,
    \begin{equation*}
        \nabla_\bmtheta J_{\bmtheta}(\lambda) = \lambda \E_{\nu p^\lambda}\left[\nabla_{\bmtheta}\ln p(\bmy|\bmx,\bmtheta)\right] - \lambda \E_\nu\left[\nabla_{\bmtheta}\ln p(\bmy|\bmx,\bmtheta) \right]\,,
    \end{equation*}
    where \(\nu p^{\lambda}\) denotes \(\nu  p^{\lambda}(\bmy,\bmx) = \frac{\nu(\bmy,\bmx)p(\bmy|\bmx,\bmtheta)^{\lambda}}{\E_\nu[p(\bmy|\bmx,\bmtheta)^{\lambda}]}\). At \(\bmtheta_0\), we have that \(\nu p^{\lambda} = \nu\), because, by definition, $p(\bmy|\bmx,\bmtheta_0)$ is constant for any $(\bmy,\bmx)$. In consequence, the gradient at $\bmtheta_0$ simplifies as, 
    \begin{equation*}
        \nabla_\bmtheta J_{\bmtheta_0}(\lambda) = \lambda \E_{\nu}\left[\nabla_{\bmtheta}\ln p(\bmy|\bmx,\bmtheta)\right] - \lambda \E_\nu\left[\nabla_{\bmtheta}\ln p(\bmy|\bmx,\bmtheta) \right] =  0\,.
    \end{equation*}
    
    The Hessian of the cumulant $\J$ w.r.t. to $\bmtheta$ can be written as follows:
        \begin{equation*}
            \begin{aligned}
                \nabla_{\bmtheta \bmtheta}\J &= \lambda^2 Cov_{\nu p^\lambda} (\nabla_{\bmtheta} \ln p(\bmy|\bmx,\bmtheta) ) + \lambda \E_{\nu p^\lambda}\left[\nabla_{\bmtheta \bmtheta}\ln p(\bmy|\bmx,\bmtheta)\right] \\
                &\quad- \lambda \E_\nu\left[\nabla_{\bmtheta \bmtheta}\ln p(\bmy|\bmx,\bmtheta) \right]\,.
            \end{aligned}
        \end{equation*}
    Again, at \(\bmtheta_0\), we have that \(\nu p^{\lambda} = \nu\), so the Hessian of the cumulant $\J$ at $\bmtheta_0$ simplifies as, \(\nabla_{\bmtheta \bmtheta}J_{\bmtheta_0}(\lambda) = \lambda^2 Cov_{\nu} (\nabla_{\bmtheta} \ln p(\bmy|\bmx,\bmtheta_0) )\). With this, the second order Taylor expansion of \(\J\) evaluated on \(\bmtheta_0\) is \(          \frac{\lambda^2 }{2}(\bmtheta-\bmtheta_0)^T \text{Cov}_{\nu} (\nabla_{\bmtheta} \ln p(\bmy|\bmx,\bmtheta_0)) (\bmtheta-\bmtheta_0)\). 
    
        On the other hand, if, at the definition of $\Iinv{s}$ given in Equation \eqref{eq:inverseratefunction}, we replace $\J$ by the above Taylor expansion, we have that, 
        \begin{equation*}
        \Iinv{s} =  \inf_{\lambda>0} \ \frac{s}{\lambda} + \frac{\lambda}{2}(\bmtheta-\bmtheta_0)^T \text{Cov}_{\nu} (\nabla_{\bmtheta} \ln p(\bmy|\bmx,\bmtheta_0)) (\bmtheta-\bmtheta_0)\,.
        \end{equation*}
        The optimal value of \(\lambda\) is acquired derivating the above expression w.r.t. \(\lambda\), giving,
        \begin{equation*}
            \frac{-s}{\lambda^2} + \frac{1}{2}(\bmtheta-\bmtheta_0)^T \text{Cov}_{\nu} (\nabla_{\bmtheta} \ln p(\bmy|\bmx,\bmtheta_0)) (\bmtheta-\bmtheta_0) = 0\,.
        \end{equation*}
        One may show that it is a minimum by taking the second derivative. From this, the optimal value of \(\lambda\) is \(
            \lambda^\star = \left(\frac{2s}{(\bmtheta-\bmtheta_0)^T \text{Cov}_{\nu} (\nabla_{\bmtheta} \ln p(\bmy|\bmx,\bmtheta_0)) (\bmtheta-\bmtheta_0)}\right)^{\frac{1}{2}}\). Using this expression,
        \begin{equation*}
            \Iinv{s} = 2s \sqrt{\frac{1}{2s}(\bmtheta-\bmtheta_0)^T \text{Cov}_{\nu} (\nabla_{\bmtheta} \ln p(\bmy|\bmx,\bmtheta_0)) (\bmtheta-\bmtheta_0)}\,.
        \end{equation*}
        Evaluating the expression on \(s = \pn\) and \(\bmtheta = \mathbf{0}\) concludes the proof.
    \end{proof}
    
    \inputgradientIinv*\label{proof:inputgradientIinv}
    \begin{proof}
    Using  \cite{chafai2004entropies}'s Corollary 2.1 on $\phi(f)=- \ln f$ and $f_\bmtheta(y,\bmx)=e^{-\lambda \ell(\bmy, \bmx,\bmtheta)}$; we get that 
    \begin{equation*}
        \J \leq M\lambda^2\ \E_\nu\Big[\big\Vert\nabla_{x}\ell(\bmy, \bmx,\bmtheta)\big\Vert_{2}^{2}\Big]\,.
    \end{equation*}
    Then, 
    \begin{equation*}
        \Iinv{s} = \inf_{\lambda > 0} \frac{s + \J}{\lambda} \leq \inf_{\lambda > 0} \frac{s +  M\lambda^2\E_\nu[|\nabla_{x}\ell(\bmy, \bmx,\bmtheta)|^2]}{\lambda}\,.
    \end{equation*}
    Deriving w.r.t. to \(\lambda\) to compute the optimal value, gives \(\lambda_{inf} = \sqrt{\frac{s}{M \E_\nu[|\nabla_{x}\ell(\bmy, \bmx,\bmtheta)|^2]}}\). Using this,
    \begin{equation*}
        \Iinv{s} \leq \sqrt{s}\sqrt{M \E_\nu[|\nabla_{x}\ell(\bmy, \bmx,\bmtheta)|^2]}\,.
    \end{equation*}
    \end{proof}
    

\subsection{Invariances (go back to Section~\ref{sec:invariances})}\label{app:proofs:invariances}
    
    \mi*\label{proof:mi}
    \begin{proof}
    According to Assumption \ref{assump:transformedinput}, the targets \(\bmy\) are independent of the input \(\bmx_t\) given \(\bmx_{t-1}\), that is,  $\bmy \perp \bmx_t|\bmx_{t-1}$. This is a result of the fact that \(\bmx_{t-1}\) d-connects \(\bmx_t\) and \(\bmy\) in the graph representation of Assumption \ref{assump:transformedinput}. The result follows then by the data processing inequality. 
    \end{proof}

    \invarianceconvex*\label{proof:invarianceconvex}
    \begin{proof}
    By definition we have \(L^{\nu_{t+1}}(\bmtheta)= \E_{\nu_{t+1}}[\ell(\bmy,\bmx_{t+1}),\bmtheta)] = \E_{\nu_t}\E_{g_{t+1}}[\ell(\bmy,g_{t+1}(\bmx_t),\bmtheta)]\). As $\ell(\bmy,\bmx,\bmtheta)$ is convex in $\bmx$, the expectation under \(g_{t+1}\) can be moved inside \(\ell\): 
    \begin{equation*}
        L^{\nu_{t+1}}(\bmtheta) \geq \E_{\nu_t} [\ell(\bmy,\E_{g_{t+1}}[g_{t+1}(\bmx_t)],\bmtheta)]\,.
    \end{equation*}
    Given that $\E_{g_{t+1}}[g_{t+1}(\bmx)]=\bmx$, we got \(L^{\nu_{t+1}}(\bmtheta) \geq \E_{\nu_t}[\ell(\bmy, \bmx_t,\bmtheta)] = L^{\nu_{t}}(\bmtheta)\).
    \end{proof}
    
    \invarianceindependence*\label{proof:invarianceindependence}
    \begin{proof}
    First of all we will show that \( J^{\nu_{t+1}}_{\bmtheta}(\lambda) \geq J^{\nu_{t}}_{\bmtheta}(\lambda)\);  from there, the result is immediate due to definition of the inverse rate. By definition, we got \(J^{\nu_{t+1}}_{\bmtheta}(\lambda) = \lambda L^{\nu_{t+1}}(\bmtheta) + \ln \E_{\nu_{t+1}}[e^{-\ell(\bmy,\bmx_{t+1},\bmtheta)}]\), where \(\nu_{t+1}\) can be expanded leading to
    \begin{equation*}
        J^{\nu_{t+1}}_{\bmtheta}(\lambda) = \lambda L^{\nu_{t+1}}(\bmtheta) + \ln \E_{\nu_{t}}\E_{g_{t+1}}\left[e^{-\lambda \ell(\bmy,g_{t+1}(\bmx_{t}),\bmtheta)}\right]\,.
    \end{equation*}
    Using the decomposition of the loss, \(J^{\nu_{t+1}}_{\bmtheta}(\lambda) = \lambda L^{\nu_{t+1}}(\bmtheta) + \ln \E_{\nu_{t}}\E_{g_{t+1}}\left[e^{-\lambda (\ell(\bmy,\bmx_{t},\bmtheta)+\Delta(\bmy,\bmx_t,g_{t+1},\bmtheta))}\right]\), which can be written as
    \begin{equation*}
        J^{\nu_{t+1}}_{\bmtheta}(\lambda) = \lambda L^{\nu_{t+1}}(\bmtheta) + \ln \E_{\nu_{t}}\left[e^{-\lambda \ell(\bmy,\bmx_{t},\bmtheta)}\right]+\ln \E_{\nu_{t}}\E_{g_{t+1}}\left[e^{-\lambda\Delta(\bmy,\bmx_t,g_{t+1},\bmtheta)}\right].
    \end{equation*}
    Using Jensen's inequality in the last term,
    \begin{equation*}
        J^{\nu_{t+1}}_{\bmtheta}(\lambda) \geq \lambda L^{\nu_{t+1}}(\bmtheta) + \ln \E_{\nu_{t}}\left[e^{-\lambda \ell(\bmy,\bmx_{t},\bmtheta)}\right]-\lambda\E_{\nu_{t}}\E_{g_{t+1}} [\Delta(\bmy,\bmx_t,g_{t+1},\bmtheta)]\,,
    \end{equation*}
    and using the definition of \(\Delta(\bmy,\bmx_t,g_{t+1},\bmtheta)\),
    \begin{equation*}
        J^{\nu_{t+1}}_{\bmtheta}(\lambda) \geq \lambda L^{\nu_{t+1}}(\bmtheta) + \ln \E_{\nu_{t}}\left[e^{-\lambda \ell(\bmy,\bmx_{t},\bmtheta)}\right]-\lambda\E_{\nu_{t}}\E_{g_{t+1}} [\ell(\bmy,g_{t+1}(\bmx_t),\bmtheta)] + \lambda\E_{\nu_{t}}[\ell(\bmy,\bmx_t,\bmtheta)]]\,.
    \end{equation*}
    By definition of the expected loss,
    \begin{equation*}
        \begin{aligned}
        J^{\nu_{t+1}}_{\bmtheta}(\lambda) &\geq  \lambda L^{\nu_{t+1}}(\bmtheta) + \ln \E_{\nu_{t}}[e^{-\lambda \ell(\bmy,\bmx_{t},\bmtheta)}]-\lambda L^{\nu_{t+1}}(\bmtheta) + \lambda L^{\nu_{t}}(\bmtheta)\\
        &= \ln \E_{\nu_{t}}[e^{-\lambda \ell(\bmy,\bmx_{t},\bmtheta)}]+\lambda L^{\nu_{t}}(\bmtheta) = J^{\nu_{t}}_{\bmtheta}(\lambda)\,.
        \end{aligned}
    \end{equation*}
    From this inequality and by considering standard properties of the Legendre transform, we have \(J^{\nu_{t+1}}_{\bmtheta}(\lambda)
    \geq J^{\nu_{t}}_{\bmtheta}(\lambda) \implies {\cal I}^{\nu_{t+1}}_{\bmtheta}(a)\leq {\cal I}^{\nu_{t}}_{\bmtheta}(a)\). 
    \end{proof}
    
    \invariance*\label{proof:invariance}
    \begin{proof}
    It immediately follows from the definition of model invariant (Definition \ref{def:invariantmodel}) and the definitions of $L^{\nu_{t+1}}(\bmtheta)$, $L^{\nu_t}(\bmtheta)$, ${\cal I}^{\nu_{t+1}}_{\bmtheta}(a)$ and ${\cal I}^{\nu_{t}}_{\bmtheta}$. 
    \end{proof}
    
    \da*\label{proof:da}
    \begin{proof}
    Under Assumption~\ref{assump:transformedinput}, it is clear that
        \begin{equation*}
            L^{v_0, \ell_G}(\bmtheta) = \E_{\nu_0}[\ell_G(\bmy, \bmx, \bmtheta))] = \E_{\nu_0}\E_{g\sim h}[\ell(\bmy, g(\bmx), \bmtheta))]  = \E_{\nu}[\ell(\bmy, \bmx, \bmtheta))] = L^{\nu_T, \ell}(\bmtheta)\,.
        \end{equation*}
    
        On the other hand, from the definition of the cummulant function, \(
            \J = \ln \E_{\nu_T}[e^{-\lambda\ell(\bmy,\bmx, \bmtheta)}] +  \lambda\E_{\nu_T}[\ell (\bmy,\bmx, \bmtheta)]\). Where by definition
        \begin{equation*}
            \J = \ln \E_{\nu_0} \E_{g\sim h}[e^{-\lambda\ell(\bmy,g(\bmx), \bmtheta)}] +  \lambda\E_{\nu_0} \E_{g\sim h}[\ell (\bmy,g(\bmx), \bmtheta)]\,,
        \end{equation*}
        where expectations can be exchanged as
        \begin{equation*}
            \J = \ln \E_{\nu_0(\bmx)}\E_{\nu(\bmy|\bmx)}\E_{g}[e^{-\lambda\ell(\bmy,g(\bmx), \bmtheta)}] +  \lambda\E_{\nu_0(\bmx)}\E_{\nu(\bmy|\bmx)}\E_{g}[\ell (\bmy,g(\bmx), \bmtheta)]\,.
        \end{equation*}
        Applying Jensen's inequality to the exponential,
        \begin{equation*}
            \J \geq \ln \E_{\nu_0(\bmx)}\E_{\nu(\bmy|\bmx)}\left[e^{-\lambda\E_{g}\left[\ell(\bmy,g(\bmx), \bmtheta)\right]}\right] +  \lambda\E_{\nu_0(\bmx)}\E_{\nu(\bmy|\bmx)}\E_{g}[\ell (\bmy,g(\bmx), \bmtheta)]\,.
        \end{equation*}
        Where by definition of \(\ell_G\),
        \begin{equation*}
            \J \geq \ln \E_{\nu_0}[e^{-\lambda[\ell_{G}(\bmy,\bmx, \bmtheta)]}] +  \lambda\E_{\nu_0}[\ell_{G} (\bmy,\bmx, \bmtheta)] = J^{\nu_0, \ell_G}_\bmtheta(\lambda)\,.
        \end{equation*}
        From this point, the inequality regarding the inverse rate is clear from the definition.
    \end{proof}
    
    \corda*\label{proof:corda}
    \begin{proof}
        Under Assumption~\ref{assump:transformedinput}, it is clear that
        \begin{equation*}
            L^{v_0, \ell_G}(\bmtheta) = \E_{\nu_0}[\ell_G(\bmy, \bmx, \bmtheta))] = \E_{\nu_0}\E_{g\sim h}[\ell(\bmy, g(\bmx), \bmtheta))]  = \E_{\nu}[\ell(\bmy, \bmx, \bmtheta))] = L^{\nu_T, \ell}(\bmtheta)\,.
        \end{equation*}
        Where it verifies that \(\L :=  L^{\nu_T, \ell}(\bmtheta)\). Thus,  the result comes from the application of Theorem~\ref{thm:LDTinv}.
    
    \end{proof}
    
    \group*\label{proof:group}
    \begin{proof}
        On one hand, by definition of \(\ell_G\) it verifies that
        \begin{equation*}
            \begin{aligned}
            \hat{L}^{\ell_G}(D_T, \bmtheta) &= \frac{1}{n}\sum_i \E_{g' \sim h}[\ell(\bmy_i, g'(\bmx_i),\bmtheta)] = \frac{1}{n}\sum_i \int h(g') \ell(\bmy_i, g'(g_i(x_{0,i})),\bmtheta)\ dg'\,.
            \end{aligned}
        \end{equation*}
        As the set of transformations define a group, for each transformation \(g_i\) that generated each input, there exists \(g_i^{-1}\). Furthermore, there exists \(l := g' \circ g_i \in G_1\) such that \(g' = l\circ g_i^{-1}\). Thus,
        \begin{equation*}
            \hat{L}^{\ell_G}(D_T, \bmtheta) = \frac{1}{n}\sum_i \int_{g' = l \circ g_i^{-1}} h(l \circ g_i^{-1}) \ell(\bmy_i,l \circ g_i^{-1}\circ g_i(x_{0,i}),\bmtheta)\ dg'\,.
        \end{equation*}
        Then, using a change of variables 
        \[
        \hat{L}^{\ell_G}(D_T, \bmtheta) = \frac{1}{n}\sum_i \int_{l} h(l \circ g_i^{-1}) \ell(\bmy_i,l(x_{0,i}),\bmtheta)\ dl\,.
        \]
        As \(h\) is uniform \( h(l \circ g_i^{-1}) = h(l)\), 
        \begin{equation*}
            \hat{L}^{\ell_G}(D_T, \bmtheta) = \frac{1}{n}\sum_i \int_{l} h(l)\ell(\bmy_i,l(x_{0,i}),\bmtheta)dl =\hat{L}^{\ell_G}(D_0, \bmtheta)\,.
        \end{equation*}
    \end{proof}
    
    \interpolatorsda*\label{proof:interpolatorsda}
    \begin{proof}
        First of all, by definition we got that \(m_\bmtheta = \essinf_{(\bmx, \bmy) \sim \nu_T} \ell(\bmy, \bmx,\bmtheta)\). Then, it verifies that
        \begin{equation*}
            \hat{L}^{\ell_G}(D_T, \bmtheta)= m_\bmtheta  \iff  \ell_G(\bmy, \bmx, \bmtheta) = \essinf_{(\bmx, \bmy) \sim \nu_T} \ell(\bmy, \bmx,\bmtheta) \quad \forall (\bmx, \bmy) \in D_T\,.
        \end{equation*}
        Thus, by definition of \(\ell_G\):    
        \begin{equation*}
            \hat{L}^{\ell_G}(D_T, \bmtheta)= m_\bmtheta  \iff \E_{g}[\ell(\bmy, g(\bmx), \bmtheta)] = \essinf_{(\bmx, \bmy) \sim \nu_T} \ell(\bmy, \bmx,\bmtheta) \quad \forall (\bmx, \bmy) \in D_T\,.
        \end{equation*}
        Now, using that \(\essinf_{(\bmx, \bmy) \sim \nu_T} \ell(\bmy, g(\bmx),\bmtheta) = m_\bmtheta \ \forall g \in G\), reaching the essential infimum in expectation means all the losses inside the expectation reach such infimum:
        \begin{equation*}
            \hat{L}^{\ell_G}(D_T, \bmtheta)= m_\bmtheta  \iff  \ell(\bmy, g(\bmx), \bmtheta) = \essinf_{(\bmx, \bmy) \sim \nu_T} \ell(\bmy, \bmx,\bmtheta) \quad \forall g\in G, \forall (\bmx, \bmy) \in D_T\,.
        \end{equation*}
        Using that for every \((\bmx_0, \bmy) \in D_0\) exists \(g' \in G\) associated to that input such that \(\bmx = g'(\bmx_0)\) and that 
        \begin{equation*}
            \hat{L}^{\ell_G}(D_T, \bmtheta)= m_\bmtheta  \iff  \ell(\bmy, g\circ g'(\bmx_0), \bmtheta) = \essinf_{(\bmx, \bmy) \sim \nu_T} \ell(\bmy, \bmx,\bmtheta) \quad \forall g\in G, \forall (\bmx_0, \bmy) \in D_0\,.
        \end{equation*}
        Then, using that there exists \(g'^{-1} \in G\), we got that
        \begin{equation*}
            \hat{L}^{\ell_G}(D_T, \bmtheta)= m_\bmtheta  \iff  \ell(\bmy, \bmx_0, \bmtheta) = \essinf_{(\bmx, \bmy) \sim \nu_T} \ell(\bmy, \bmx,\bmtheta) \quad \forall (\bmx_0, \bmy) \in D_0\,.
        \end{equation*}
        As a result, the empirical loss \(\hat{L}^{\ell}(D_0, \bmtheta)\) is equal to \(m_\bmtheta\) too.
    \end{proof}

\subsection{Over-Parameterization (go back to Section~\ref{sec:Over-parameterization})}\label{app:proofs:over-parameterization}

    \overinvariance*\label{proof:overinvariance}
    \begin{proof}
        Due to fact that $\Lhat\leq \epsilon$ and to Theorem \ref{thm:LDTinv}, we have that with high probability $1-\delta$ over $D\sim\nu^n(y,x)$,
        \begin{equation*}
            \L\leq \epsilon + \Iinv{\pn}\,.
        \end{equation*}
        Then, just rearranging terms as follows, we get \(\L-\epsilon \leq \Iinv{\pn}\). Applying the rate function at both sides,
        \begin{equation*}
            \I{\L-\epsilon} \leq \pn = \frac{1}{n}\left(\ln k^p - \ln \delta \right)\,.
        \end{equation*}
        Then, we got \(\ln k^p \geq n\I{\L-\epsilon} + \ln\delta\) and
        \begin{equation*}
            p \geq \frac{n\I{\L-\epsilon} + \ln\delta}{\ln k}\,.
        \end{equation*}
        As $L^\star\leq \L$ and the rate function is monotonically increasing, then
        \begin{equation*}
            p \geq \frac{n\I{\L-\epsilon} + \ln\delta}{\ln k}\geq \frac{n\I{L^\star - \epsilon} + \ln\delta}{\ln k}\,.
        \end{equation*}
    \end{proof}

    \overinvariancelipchitz*\label{proof:overinvariancelipchitz}
    \begin{proof}
    Under the isoperimetry assumption, we have \(\I{a} \leq \frac{d a^2}{2 c Lip(\bmtheta)^2 } \). Due to fact that $\Lhat\leq \epsilon$ and to Theorem \ref{thm:LDTinv}, we have that with high probability $1-\delta$ over $D\sim\nu^n(y,x)$,
        \begin{equation*}
            \L\leq \epsilon + \Iinv{\pn}\,.
        \end{equation*}
        Then, just rearranging terms as follows, we get \(\L-\epsilon \leq \Iinv{\pn}\). Applying the rate function at both sides,
        \begin{equation*}
            \I{\L-\epsilon} \leq \pn = \frac{1}{n}\left(\ln k^p - \ln \delta \right)\,.
        \end{equation*}
        Then, we got \(\ln k^p \geq n\I{\L-\epsilon} + \ln\delta\) and
        \begin{equation*}
            p \geq \frac{n\I{\L-\epsilon} + \ln\delta}{\ln k}\,.
        \end{equation*}
        As $L^\star\leq \L$ and the rate function is monotonically increasing, then
        \begin{equation*}
            p \geq \frac{n\I{\L-\epsilon} + \ln\delta}{\ln k}\geq \frac{n\I{L^\star - \epsilon} + \ln\delta}{\ln k}\,.
        \end{equation*}
        Using the upper bound over the rate function we obtained before, we got that
        \begin{equation*}
            p \geq \frac{n\frac{d(L^\star - \epsilon)^2}{2cLip(\bmtheta)^2} + \ln\delta}{\ln k}\,.
        \end{equation*}
        Re-arranging terms,
        \begin{equation*}
            \sqrt{\frac{nd(L^\star - \epsilon)^2} {2c(p\ln k - \ln \delta)}} \leq Lip(\bmtheta)\,.
        \end{equation*}
    \end{proof}
    
    \overinvariancenorm*\label{proof:overinvariancenorm}
    \begin{proof} If the loss is Lipschitz continuous with constant $M$, \(\forall y,x,\bmtheta\quad \|\nabla_{\bmtheta} \ell(\bmy,\bmx,\bmtheta)\|^2_2\leq M\). Then, $\J$ verifies
        \begin{equation}
             \|\nabla_{\bmtheta}\J\|^2_2=||-\lambda \E_{\nu p^\lambda}\left[\nabla_{\bmtheta}\ell(\bmy, \bmx, \bmtheta)\right] + \lambda \E_\nu\left[\nabla_{\bmtheta}\ell(\bmy, \bmx, \bmtheta) \right]||_2^2 \leq 2M \lambda^2\,.
         \end{equation}
         where \(\E_{\nu p^\lambda}\left[\nabla_{\bmtheta}\ell(\bmy, \bmx, \bmtheta)\right] = \frac{\E_{\nu}[p(\bmy|\bmx, \bmtheta)^\lambda \ell(\bmy, \bmx, \bmtheta)]}{\E_{\nu}[p(\bmy|\bmx, \bmtheta)^\lambda]} \). With this, we have
         \begin{equation*}
             |\J - J_{\bmtheta_0}(\lambda)|\leq 2M \lambda^2\|\bmtheta-\bmtheta_0\|^2_2
             \implies \J\leq 2M \lambda^2\|\bmtheta-\bmtheta_0\|^2_2\,.
         \end{equation*}
         Then, for any \(a \geq 0\), we have  by definition of the rate function that
         \begin{equation*}
             \I{a} \geq a\lambda - \J \geq a\lambda-2M \lambda^2\|\bmtheta-\bmtheta_0\|^2_2\,.
         \end{equation*}
         As the inequality holds for any \(\lambda > 0\), maximizing it raises \(\lambda^\star = \frac{a}{4M\|\bmtheta-\bmtheta_0\|^2_2}\). Thus,
         \begin{equation*}
             \I{a} \geq a\lambda - \J \geq a\frac{a}{4M\|\bmtheta-\bmtheta_0\|^2_2}-2M \frac{a^2}{(4M\|\bmtheta-\bmtheta_0\|^2_2)^2}\|\bmtheta-\bmtheta_0\|^2_2 = \frac{a^2}{8M\|\bmtheta-\bmtheta_0\|^2_2}\,.
         \end{equation*}
        Due to fact that $\Lhat\leq \epsilon$ and to Theorem \ref{thm:LDTinv}, we have that with high probability $1-\delta$ over $D\sim\nu^n(\bmx, \bmy)$, \(\L\leq \epsilon + \Iinv{\pn}\). Then, just rearranging terms as follows, we get \(\L-\epsilon \leq \Iinv{\pn}\). Applying the rate function at both sides,
        \begin{equation*}
            \I{\L-\epsilon} \leq \pn = \frac{1}{n}\left(\ln k^p - \ln \delta \right)\,.
        \end{equation*}
        Then, we got \(\ln k^p \geq n\I{\L-\epsilon} + \ln\delta\) and \(p \geq \frac{n\I{\L-\epsilon} + \ln\delta}{\ln k}\). As $L^\star\leq \L$ and the rate function is monotonically increasing, then
        \begin{equation*}
            p \geq \frac{n\I{\L-\epsilon} + \ln\delta}{\ln k}\geq \frac{n\I{L^\star - \epsilon} + \ln\delta}{\ln k}\,.
        \end{equation*}
        Using the upper bound over the rate function we obtained before, we got that
        \begin{equation*}
            p \geq \frac{n\frac{(L^\star - \epsilon)^2}{8M\|\bmtheta-\bmtheta_0\|^2_2} + \ln\delta}{\ln k} \implies (L^\star - \epsilon)\sqrt{\frac{n}{8M(p\ln k - \ln \delta)}} \leq \|\bmtheta-\bmtheta_0\|_2\,.
        \end{equation*}
    \end{proof}

    \largermodelssmoothness*\label{proof:largermodelssmoothness}
    \begin{proof}
    We can apply Theorem \ref{thm:LDTinv} on $\bmtheta$ assuming that $\bmtheta\in\bmTheta'$, because we have that $\bmTheta\subset \bmTheta'$. In consequence, with h.p., we have 
    \begin{equation*}
    \L \leq \Lhat + \Iinv{\tfrac{1}{n}\ln\tfrac{k^{p'}}{\delta}}    
    \end{equation*}
    Using the fact that $\Lhat\leq\epsilon$, and that the inverse rate is bounded by the expected loss, we arrive to the following inequality
    \begin{equation*}
    \L \leq \epsilon + \Iinv{\tfrac{1}{n}\ln\tfrac{k^{p'}}{\delta}}\leq \epsilon + \L    
    \end{equation*}
    The same reasoning applies for $\bmtheta'$:
    \begin{equation*}
    \Lprime \leq \epsilon + \Iinvprime{\tfrac{1}{n}\ln\tfrac{k^{p'}}{\delta}}\leq \epsilon + \Lprime
    \end{equation*}
    Using the theorem's premise, $\Lprime + \epsilon< \L$, we can chain the last two h.p. upper bounds. And note that we can do that \textit{with no extra cost}, as both apply on the same bigger model class $\bmTheta'$. That is, it is the same upper bound, which holds simultaneously for all models within $\bmTheta'$, used on two different models $\bmtheta$ and $\bmtheta'$. Then, we have, 
    \begin{equation*}
    \Iinvprime{\tfrac{1}{n}\ln\tfrac{k^{p'}}{\delta}}< \Iinv{\tfrac{1}{n}\ln\tfrac{k^{p'}}{\delta}}\,.   
    \end{equation*}
    Naming \(a=\Iinv{\frac{1}{n}\ln\frac{k^{p'}}{\delta}}\), such that $\frac{1}{n}\ln\frac{k^{p'}}{\delta}=\I{a}$, we got that
    \begin{equation*}
    \Iinvprime{\I{a}}< \Iinv{\I{a}} = a\,.   
    \end{equation*}
    Applying \(\Iprime{\cdot}\) at both sides:
    \begin{equation*}
    \Iprime{\Iinvprime{\I{a}}} = \I{a} < \Iprime{a}\,.   
    \end{equation*}
    Then, we can negate the inverse statement using \(\neg\) notation as \(\neg\Big[ \I{a}\geq \Iprime{a}\Big]\). Then we have that $\bmtheta$ is not $\beta$-smoother than $\bmtheta'$ for any $\beta\geq a$. Which is equivalent to say that, $\bmtheta$ is not $\beta$-smoother than $\bmtheta'$ for any $\beta$ such that $\I{\beta}\geq \frac{1}{n}\ln\frac{k^{p'}}{\delta}$.
    \end{proof}

    \largermodelssmoothnesssecond*\label{proof:largermodelssmoothnesssecond}
    \begin{proof}
    If $\bmtheta'$ is  $\beta$-smoother than $\bmtheta$, by Definition \ref{def:smoothness}, \(\forall a\in(0,\beta] \quad \Iprime{a}\geq \I{a}\) where $\beta = \Iinvprime{\frac{1}{n}\ln\frac{k^{p'}}{\delta}}$. Then, we have that \(\Iinvprime{s}\leq \Iinv{s}\) for $s=\Iprime{\beta} = \mathcal{I}_{\bmtheta'}\Big(\Iinvprime{\tfrac{1}{n}\ln\tfrac{k^{p'}}{\delta}}\Big) = \tfrac{1}{n}\ln\tfrac{k^{p'}}{\delta}$. Thus, 
    \begin{equation*}
        \Iinvprime{\tfrac{1}{n}\ln\tfrac{k^{p'}}{\delta}}\leq \Iinv{\tfrac{1}{n}\ln\tfrac{k^{p'}}{\delta}}\leq \L\,.
    \end{equation*}
    Where we used that the inverse rate is upper bounded by \(\L\). By Theorem \ref{thm:LDTinv}, because $\Lhat\leq \epsilon$, we have with h.p. $1-\delta$ over $D\sim\nu^n$, 
    \begin{equation*}
        \Lprime\leq \Lhatprime + \Iinvprime{\tfrac{1}{n}\ln\tfrac{k^{p'}}{\delta}}\leq \epsilon + \Iinv{\tfrac{1}{n}\ln\tfrac{k^{p'}}{\delta}}\,.
    \end{equation*}
    By combining the last two inequalities, we have 
    \begin{equation*}
        \Lprime\leq \epsilon + \Iinvprime{\tfrac{1}{n}\ln\tfrac{k^{p'}}{\delta}} \leq \epsilon + \L\,.
    \end{equation*}
    \end{proof}

\bibliography{bibliography}
\bibliographystyle{apalike.bst}

\end{document}